\documentclass[10pt]{article}
\makeatletter
\ifnum\@ptsize=0\fi
\ifnum\@ptsize=2\fi

\usepackage[pdftex]{graphicx}

\usepackage{amssymb}

\usepackage[linesnumbered,ruled]{algorithm2e}
\usepackage{color}
    
\begin{document}




\title{Recognition of feature curves on 3D shapes using an algebraic approach to Hough transforms
}


\author{Maria-Laura Torrente, Silvia Biasotti, Bianca Falcidieno}
\maketitle

\begin{abstract}
Feature curves are largely adopted to highlight shape features, such as sharp lines, or to divide surfaces into meaningful segments, like convex or concave regions. 
Extracting these curves is not sufficient to convey prominent and meaningful information about a shape. We have first to separate the curves belonging to features from those caused by noise and then to select the lines, which describe non-trivial portions of a surface.
The automatic detection of such features is crucial for the identification and/or annotation of relevant parts of a given shape.  
To do this, the Hough transform (HT) is a feature extraction technique widely used in image analysis, computer vision and 
digital image processing, while, for 3D shapes, the extraction of salient feature curves is still an open problem.

Thanks to algebraic geometry concepts, the HT technique has been recently extended to include a vast class of algebraic curves, thus proving to be a competitive tool for yielding an explicit representation of the diverse feature lines equations. 
In the paper, for the first time we apply this novel extension of the HT technique to the realm of 3D shapes in
order to identify and localize semantic features like patterns, decorations or anatomical details on 3D objects
(both complete and fragments), even in the case of  features partially damaged or incomplete. 
The method recognizes various features, possibly compound, and it selects the most suitable feature profiles among families 
of algebraic curves.
\end{abstract}



\section{Introduction}
Due to the intuitiveness and meaningful information conveyed in human line drawings, feature curves have been largely
investigated in shape modelling and analysis to support several processes, ranging from non-photorealistic rendering to simplification, segmentation and sketching of graphical information \cite{Cole:2008,KST08, deGoes2011, Gehre16}. 

These curves can be represented as curve segments identified by a set of vertices, splines interpolating the feature points \cite{APM15}, $L^1$ medial skeletons \cite{Zhang2016} or approximated with known curves, like spirals \cite{Harary2011,Harary2012}.

Traditional methods proposed for identifying feature curves on 3D models can be divided into view dependent and view independent methods. The first ones extract feature curves from a projection of the 3D model onto a plane perpendicular to the view direction, while view independent techniques extract feature points by computing curvatures or other differential properties of the model surface.  


View independent feature curves assume various names according to the criteria used for their characterization (ridge/valleys, crest lines, sharp lines, demarcating curves, etc.). Furthermore, they are possibly organized into curve networks \cite{deGoes2011,Gehre16} and filtered to omit short and non-salient curves \cite{Cao2015}.

However, extracting feature curves is not sufficient to convey prominent and meaningful information about a shape. We have first to separate the curves belonging to features from those caused by noise and then, among the remaining curves, to select the lines which describe non-trivial portions of a surface.
These salient curves usually correspond to high level features that characterize a portion of a shape. They can be represented by a multitude of similar occurrences of similar simple curves (like the set of suction cups of an octopus tentacle), or by the composition of different simple curves (like the eye and pupil contours).
In addition, for some applications, it is necessary to develop methods applicable also in the case of curves partially damaged or incomplete, as in the case of archaeological artefacts.

To the best of our knowledge, the literature has not yet fully addressed the problem of identifying on surfaces similar occurrences of high level feature curves, of different size and orientation, and in the case they are degraded.
This is not true for 2D feature curves, where the  Hough Transform (HT) is commonly used for detecting lines as well as parametrized curves or 2D shapes in images. Our attempt is then to study an HT-based approach suitable for extracting feature curves on 3D shapes.   
We have found the novel generalization of the HT introduced in \cite{beltrametti2012algebraic} convenient for the detection of curves on 3D shapes. 
Using this technique we can identify suitable algebraic curves exploiting computations in the parameters space, thus providing an explicit representation of the equation of the feature curves. Moreover, we recognize which curves of the family are on the shape and how many occurrences of the same curve are there (see Section \ref{examples} for various illustrative examples).
Applying this framework to curves in the 3D space is not a trivial task: spatial algebraic curves can be represented as the intersection
of two surfaces and theoretical foundations for their detection via HT has already been laid using Gr\"obner bases theory 
(see \cite{beltrametti2012algebraic}). Nevertheless, the atlas of known algebraic surfaces is not as wide as that of algebraic plane curves, therefore working directly with curves could end with some limitations. Further,  similarly to \cite{KST08, Harary2011, Harary2012, Itskovich2011} 
our point of view is local since we are interested to the problem of the extraction of features contours that can be locally flattened onto a plane without any overlap.

\paragraph*{Contribution}
In this paper we describe a new method to identify and localize feature curves, which characterize semantic features like
patterns, decorations, reliefs or anatomical features on the digital models of 3D objects, even if the features are partially 
damaged or incomplete. The focus is on the extraction of feature curves from a set of potentially significant points using the 
cited generalization of the HT \cite{beltrametti2012algebraic}. 
This  technique takes advantage of a rich family of primitive curves that are flexible to meet the user needs. The method recognizes various features, possibly compound, and selects the most suitable profile among families of algebraic curves. Deriving from the HT, our method inherits the robustness to noise and the capability of dealing with data incompleteness
as for the degraded and broken 3D artefacts on which we realized our first experiments \cite{gch.2016}.
Our main contributions can be summarized as follows:
\begin{itemize}
\item To the best of our knowledge this is the first attempt to systematically apply the HT to the recognition of curves on 3D shapes. 
\item The method is independent of how the feature points are detected, e.g. variations of curvature, colour, or both; in general, we admit a multi-modal characterization of the feature lines to be identified, see Section \ref{FeaturePointChar}.
\item The method is independent of the model representation, we tested it on point clouds and triangle meshes but the same framework applies to other representations like quad meshes. Details on the algorithms are given in Section \ref{Sec:overview} and in the Appendix.
\item A vast catalogue of functions is adopted, which is richer than previous ones, and it is shown how to modify the parameters to include families of curves instead of a single curve, details are in Section \ref{Sec:curves}.
\item The set of curves is open and it can be enriched with new ones provided that they have an algebraic representation.
\item We introduce the use of curves represented also in polar coordinates, like the Archimedean spiral.
\item Our framework includes also compound curves, see Section \ref{sec:compound}.
\item As a proof of concept, we apply this method to real 3D scans, see Section \ref{examples}.
\end{itemize}
If compared to the previous methods, we think that our approach, conceived under the framework of the Hough Transform 
technique, can be used and tuned for a larger collection of curves. Indeed, we will show how spirals, geometric petals and 
other algebraic curves can be gathered using our recognition technique.
\section{Related work}
The literature on the extraction of feature curves and the Hough transform is vast and we cannot do justice to it here.
In this section we limit our references only to the methods relevant to our approach, focusing on HT-based curve detection and feature curve characterization. 

\paragraph*{Hough transform} 
We devote this section to a brief introduction to the HT technique, while we refer to recent surveys 
(for instance \cite{Mukhopadhyay2015,Kassim1999}) for a detailed overview.

HT is a standard pattern recognition technique originally used to detect straight lines in images, \cite{c1962method,DH72}. Since its original conception, HT has been extensively used and many generalizations have been proposed for identifying instances of arbitrary shapes over images \cite{ballard1981generalizing}, more commonly circles or ellipses. 
The first very popular extension concerning the detection of any parametric analytic curve is usually referred to as the Standard Hough Transform (SHT). In spite of the robustness of SHT to discontinuity or missing data on the curve, its use has been limited by a number of drawbacks, like the need of a parametric expression, or the dependence of the computation time and the memory requirements on the number of curve parameters, or even the need of finer parameters quantization for a higher accuracy of results.  

To overcome some of these limitations, other variants have been proposed, one of the most popular being
the Generalized Hough Transform (GHT) by Ballard \cite{ballard1981generalizing}. Since its conception, it proved to be very useful for detecting
and locating translated two-dimensional objects, without requiring a parametric analytic expression. 
Thus, GHT is more general than~SHT, as it is able to detect a larger class of rigid objects, still retaining the robustness
of~SHT. Nevertheless, GHT often requires brute force to enumerate all the possible orientations and scales of the 
input shape, thus the number of parameters needs to be increased in its process. Further, GHT cannot adequately handle shapes that are more flexible, as in the case of different instances of the same shape, which are similar but not identical, e.g. petals and leaves.

Recently, thanks to algebraic geometry concepts, theoretical foundations have been laid to extend the HT technique 
to the detection of algebraic objects of codimension greater than one (for instance algebraic space curves) 
taking advantage of various families of algebraic plane curves (see \cite{beltrametti2012algebraic} and
 \cite{beltrametti2013hough}). Being so general, such a method allows to deal with different shapes, possibly compound,  and to get the most suitable approximating profile among a large vocabulary of curves. 

In 3D, other variants of HT have been introduced and used but as far as we know none of them exploits the huge variety 
of algebraic plane curves (for this we refer again to the surveys \cite{Mukhopadhyay2015} and \cite{Kassim1999}). For instance, in \cite{alliez} the HT has been employed to identify recurring straight line elements on the walls of buildings. In that application, the HT is applied only to planar point sets and line elements are clustered according to their angle with respect to a main wall direction; in this sense, the Hough aggregator is used to select the feature line directions (horizontal, vertical, slanting) one at a time.

\paragraph*{Feature characterization}
Overviews on methods for extracting feature points are provided in \cite{Cole:2008,Lai2007}. 
In the realm of feature characterization it is possible to distinguish between features that are view-dependent, like silhouettes, suggestive contours and principal highlights \cite{DeCarlo}, mainly useful for rendering purposes \cite{Lawonn2017}, or those that are independent of the spatial embedding and, therefore, more suitable for feature recognition and classification processes.
In the following we sketch some of the methods that are relevant for our approach, i.e., characterizations that do not depend on the spatial embedding of the surface.

A popular choice to locate features is to estimate the curvature, either on meshes \cite{KST08,YBY*08} or point clouds \cite{Gumhold01,DOH08}). 
When dealing with curvature estimation, parameters have to be tuned according to the target feature scale and the underlying noise.
The Moving Least Square method \cite{PKG03} and its variations are robust to scale variations \cite{DOH08, kalogerakis07} and does not incorporate smoothness effects in the estimation. 
Alternative approaches to locate curvature extrema use discrete differential operators \cite{HPW05} or probabilistic methods, such as random walks \cite{Luo2010}. For details on the comparison of methods for curvature estimation, we refer to a recent benchmark \cite{vasa}.

Partial similarity and, in particular, self-similarity, is the keyword used to detect repeated features over a surface.
For instance, the method \cite{gal2006salient} is able to recognize repeated surface features (circles or stars) over a surface. However, being based on geometry hashing, the method is scale dependent and does not provide the exact parameters that characterize such features.

Despite the consolidate literature for images, feature extraction based on colour information is less explored for 3D shapes, and generally used as a support to the geometric one \cite{Biasotti2016:CGF,BCF*15}. Examples of descriptions for textured objects adopt a 3D feature-vector description, where  the colour is treated as a general property without considering its distribution over the shape, see for instance \cite{Suzuki01,Ruiz09,Starck07}.
Another strategy is to consider local image patches that describe the behaviour of the texture around a group of pixels. Examples of these descriptions are the Local Binary Patterns (LPB) \cite{ojala}, the Scale Invariant Feature Transform (SIFT) \cite{Lowe2004}, the Histogram of Oriented Gradients (HOG) \cite{DaTr05} and the Spin Images \cite{Johnson:1999}. The generalization of these descriptors to 3D textured models has been explored in several works, such as the VIP description~\cite{WuCLFP08}, the meshHOG \cite{meshHOG} and the Textured Spin-Images \cite{Pasqualotto2013}.
Further examples are the colour-CHLAC features computed on 3D voxel data proposed in \cite{Kanezaki:2010}; the sampling method \cite{Liu:2012} used to select points in regions of either geometry-high variation or colour-high variation, and to define a signature based on feature vectors computed at these points; the CSHOT descriptor \cite{TombariSS11}, meant to solve point-to-point correspondences coding geometry- and colour-based local invariant descriptors of feature points. However, these descriptors are local, sensitive to noise and, similarly to \cite{gal2006salient}, scale, furthermore they do not provide the parameters that characterize the features detected.

\paragraph*{Feature curves identification}
The extraction of salient features from surfaces or point clouds has been addressed either in terms of curves \cite{Gumhold01}, segments (i.e. regions) \cite{Shamir:2006:STAR} and shape descriptions \cite{CSUR2008}. Thanks to their illustrative power, feature curves are a popular tool for visual shape illustration \cite{KST08} 
and perception studies support feature curves as a flexible choice for representing the salient parts of a 3D model \cite{Cole:2008,Harary2011}.

Feature curves are often identified as ridges and valleys, thus representing the extrema of principal curvatures \cite{OBS04,YBY*08,Cao2015} or sharp features \cite{Lai2007}. 
Other types of lines used for feature curve representation are parabolic ones. They partition the surface into hyperbolic and elliptic regions, and zero-mean curvature curves, which classify sub-surfaces into concave and convex shapes \cite{koenderink1984}. Parabolic lines correspond to the zeros of the Gaussian and mean curvature, respectively. 
Finally, demarcating curves are the zero-crossings of the curvature in its gradient direction \cite{KST08, KST11}.
In general, all these curves, defined as the zero set of a scalar function, do not consider curve with knots, fact that an algebraic curve like the \emph{Cartesian Folium} (see \cite{shikin1995handbook}) could arrange.

In general, given a set of (feature) points, the curve fitting problem is largely addressed in the literature, \cite{Farin:1993,shikin,Piegl:1997,DOH08}. Among the others, we mention \cite{APM15} that recently grouped the salient points into a curve skeleton that is fitted with a quadratic spline approximation. Being based on a local curve interpolation, such a class of methods is not able to recognize entire curves, to complete missing parts and it is difficult to assess if a feature is repeated at different scales.

Besides interpolating approaches such as splines, it is possible to fit the feature curve set with some specific family of curves, for instance the \emph{natural 3D spiral} \cite{Harary2011} and the \emph{3D Euler spiral} \cite{Harary2012} have been proposed as a natural way to describe line drawings and silhouettes showing their suitability for shape completion and repair. However, using one family of curves at a time implies the need of defining specific solutions and algorithms for settings the curve parameters during the reconstruction phase.

Recently, feature curve identification has been addressed with co-occurrence analysis approaches \cite{SunkelEG2011,Li2015}. In this case, the curve identification is done in two steps: first, a local feature characterization is performed, for instance computing the Histogram of Oriented Curvature (HOC) \cite{Kerber2013} or a depth image of the 3D model \cite{SunkelEG2011}; second, a learning phase is applied to the feature characterization. 
The learning phase is interactive and requires 2-3 training examples for every type of curve to be identified and sketched; in case of multiple curves it is necessary also to specify salient nodes for each curve. Feature lines are poly-lines (i.e. connected sequences of segments) and do not have any global equation.
These methods are adopted mainly for recognizing parts of buildings (such as windows, doors, etc.) and features in architectural models that are similar to strokes.
Main limitations of these methods are the partial tolerance to scale variance, the need of a number of training curves for each class of curves, the non robustness to missing data and the fact that compound features can be addressed only one curve at a time~\cite{SunkelEG2011}.

In conclusion, we observe that the existing feature curve identification methods on surfaces do not satisfy all the good properties typical of the Hough transforms, such as the robustness to noise, the ability to deal with partial information and curve completion, the accurate evaluation of the curve parameters and the possibility of identifying repeated or compound curves. Moreover, the vocabulary of possible curves is generally limited to straight lines, circles, spirals, while the HT-based framework we are considering encompasses all algebraic curves.


\section{Overview of the method}
\label{Sec:overview}
Our approach to the extraction of peculiar curves from feature points of a given 3D model is general, it can be applied to identify anatomical features, extract patterns, localize decorations, etc. Our point of view is local since we are interested to the extraction of features contours that locally can be projected on a plane without any overlap. From the mathematical point of view, every surface can be locally projected onto a plane using an injective map and, if locally regular, it can be expressed in local coordinates as $(x,y,z(x,y))$ \cite{rudin}. 

We assume that the geometric model of an object is available as a triangulated mesh or a point cloud, possibly equipped with colour. Nevertheless, our methodology can be applied also to other model representations like quad meshes.
Moreover, the photometric information, which, if present, contains rich information about the real appearance of objects (see \cite{TWW01}), can be exploited alone or in combination with the shape properties for extracting the feature points sets. 

Since the straight application of the HT technique to curves in the 3D space is not trivial (indeed, a curve is represented by the intersection of two surfaces), here we describe the steps necessary  to identify the set of points that are candidate to belong to a feature curve, to simplify their representation using a local projection and to approximate the feature curve. 
Basically, we identify three main steps:

\begin{description}
\item {\em Step 1: Potential feature points recognition:} using different shape properties it extracts the sets of feature points from the  input model; then, points are aggregated into smaller dense subsets; it works in the 3D space, see details in 
Section~\ref{FeaturePointChar}.
\item {\em Step 2: Projection of the feature sets onto best fitting planes:} it computes a projection of each set of points obtained in the step 1 onto a best fitting plane; see Section~\ref{PointCloudProj}.
\item {\em Step 3: Feature curve approximation:} based on a generalization of the HT it computes an approximation of the feature curve; it is applied to each set resulting from step 2;  see Section~\ref{Hough}.
\end{description}
While the first step is done only once, the second and the third ones run over each set of potential feature points. 

\begin{figure}[ht]
\centering
\includegraphics[scale=.6]{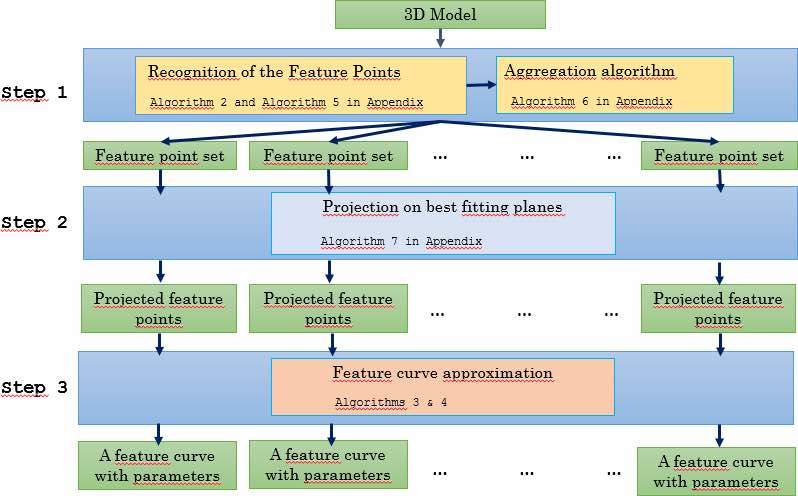}
\caption{A synthetic flowchart of our method.}
\label{fig:workflow}
\end{figure}

For the sake of clarity, a synthetic flowchart of our method is shown in Figure \ref{fig:workflow}. In the boxes the pseudo-code algorithms corresponding to the different actions are referred.

We provide the outline of our feature recognition method in the Main Algorithm~\ref{mainAlg} and we refer to 
Sections \ref{FeaturePointChar}-\ref{Hough} for a detailed description of the procedures.
For the convenience of the reader we sum up the variables of input/output and the notation we use in the 
Main Algorithm \ref{mainAlg}.
The algorithm requires four inputs: 
\begin{itemize}
\item a given 3D model denoted by $\mathcal M$; 
\item a property of the shape, such as curvature
or photometric information, denoted by $prop$ and used to extract the feature points set; 
\item a threshold denoted by $p$ and used for filtering
the feature points; 
\item an HT-regular family of planar curves denoted by $\mathcal F$. 
\end{itemize}
In the body of the Main Algorithm \ref{mainAlg}
the following variables are used: $\mathbb X$, which denotes the set of feature points extracted from the model $\mathcal M$ by means of
the Feature Points Recognition Algorithm \ref{algFeaturePoints}; $\mathbb Y_j$, with $j=1 \ldots m$, which denotes the subset of points 
of $\mathbb X$ sharing some similar properties obtained applying the Aggregation Algorithm \ref{algClustering}; $\mathbb Z_j$, 
with $j=1 \ldots m$, which is the local projection of $\mathbb Y_j$ to its best fitting plane obtained applying Projection 
Algorithm \ref{algProjection}; $\mathcal C_j$, which is the curve of the family $\mathcal F$ that best approximates $\mathbb Z_j$. $\mathcal C_j$ is 
obtained applying the Curve Detection Algorithm \ref{algHough} to $\mathbb Z_j$ with respect to a family of functions $\mathcal F$ in a region $\mathcal T$ of the parameter space which is discretized with a step $d$.
\subsection{Potential feature points recognition} \label{FeaturePointChar}
Using various shape properties (curvatures, parabolic 
points identification or photometric information) the feature points set is extracted from the input model.
In this paper we concentrate on features highlighted by means of curvature functions and/or colorimetric attributes.

The geometric properties are derived using classical curvatures, like {\em minimum}, {\em maximum}, {\em mean}, {\em Gaussian} or {\em total curvatures}. 
We denote these well-known geometric quantities by $C_{min}$, $C_{max}$, $C_{mean}$, $C_{Gauss}$, $C_{tot}$ respectively. Several methods for estimating curvatures over triangle meshes and point clouds exist but, unfortunately, there is not a universal solution that works best for every kind of input \cite{vasa}. 
We decided to keep our implementation flexible, adopting in alternative a discretization of the normal cycles \cite{Cohen03} proposed in the Toolbox graph~\cite{ToolboxGraph} and the statistic-based method presented in \cite{kalogerakis07}. 
%
%
\begin{algorithm}
\SetAlgorithmName{Main Algorithm}
\DontPrintSemicolon
\SetAlgoLined
\SetKwInOut{Input}{Input}
\SetKwInOut{Output}{Output}
\SetKwInOut{return}{return}
\Input{a 3D model $\mathcal M$, the property $prop$ (the type of curvature and/or colour), the threshold $p$ for feature points filtering, an HT-regular family $\mathcal F$ of planar curves}
\Output{a list of curves belonging to the family $\mathcal F$}
 \Begin{  
  \tcc{extracts the feature points set from the model $\mathcal M$}
 $\mathbb X \gets $ Feature Points Recognition Algorithm \ref{algFeaturePoints} applied to $\mathcal M$, $prop$ and $p$\;
 \tcc{groups the points of $\mathbb X$ into smaller dense subsets}
 $\mathbb Y =[\mathbb Y_1,\ldots, \mathbb Y_m]\gets $ Aggregation Algorithm \ref{algClustering} applied to $\mathbb X$\; 
 \For{$j=1,\ldots, m$}{
 \tcc{projects the points of $\mathbb Y_j$ onto the best fitting plane}
 $\mathbb Z_j \gets $Projection Algorithm \ref{algProjection} (see Appendix) applied to $\mathbb Y_j$;\\
 $t \gets $ number of parameters of the curves of $\mathcal F$\;
 \tcc{initializes the region $\mathcal T$ of the parameter space and its discretization step $d$}
$\mathcal T \gets [a_1,b_1] \times \ldots \times [a_t,b_t] \in \mathbb R^t$;
$d \gets (d_1,\ldots,d_t) \in \mathbb R_{>0}^t$\;
 \tcc{computes the curve of $\mathcal F$ that best approximates $\mathbb Z_j$}
 $\mathcal C_j \gets$ Curve Detection Algorithm \ref{algHough} applied to $\mathbb Z_j, \mathcal F, \mathcal T, d$\;
 }
 \Return{$\mathcal C= [\mathcal C_1,\ldots, \mathcal C_m]$}
}
 \caption{Feature curve identification for a given 3D model $\mathcal M$ with respect to a property (different types of 
 curvatures/colour) and a family $\mathcal F$ of curves}\label{mainAlg}
 \end{algorithm}

The photometric properties can be represented in different
colour spaces, such as RGB, HSV, and CIELab spaces. Our choice is to work in the CIELab space \cite{AKK00}, which has been proved to approximate human vision in a good way. In such a space, tones and colours are distinct: 
the $L$ channel is used for the luminosity, which closely matches the human perception of light ($L=0$ yields black and $L=100$ yields diffuse white), whereas the $a$ and $b$ channels 
specify colours~\cite{color2011}. 

The different types of curvatures $C_{min}$, $C_{max}$, $C_{mean}$, $C_{Gauss}$, $C_{tot}$, and the luminosity $L$ are used as scalar real functions defined over the surface vertices.
One (or more) of these properties is given as input (stored in the variable $prop$) in the Feature Points Recognition Algorithm \ref{algFeaturePoints}. 
The \emph{feature points} $\mathbb X$ of the model~$\mathcal M$ are extracted by selecting the vertices at which 
the property $prop$ is significant (e.g high maximal curvature and/or low minimal curvature and/or low luminosity).
This is automatically achieved by filtering the distribution of the function that we decide to use by means
of a filtering threshold $p$ (see Appendix, Algorithm \ref{algFilter}). Note that $p$ is given in input in the Main Algorithm \ref{mainAlg}. Its value varies according to the precision threshold set for the property used to extract the feature points (e.g. in the case of maximum curvature a typical value of $p$ is $80 \%$).
We sum up the described procedure in the Algorithm \ref{algFeaturePoints}.

\begin{figure}[htb]
\begin{center}
\begin{tabular}{ccccc}
\includegraphics[height=5.1cm]{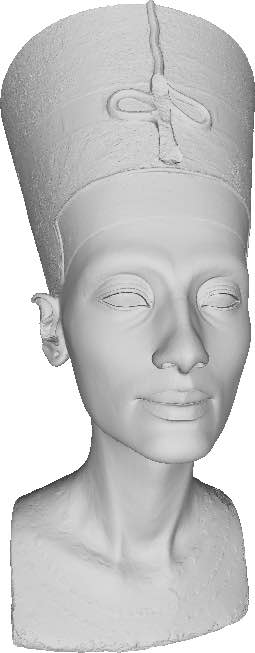} & \hspace{0.5cm} &
\includegraphics[height=5.1cm]{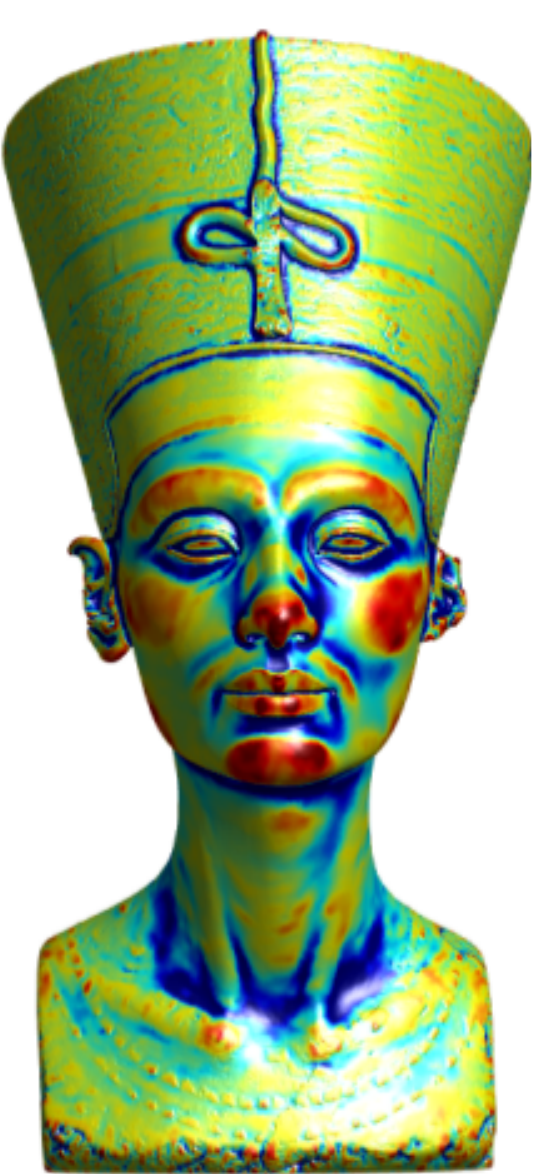} & \hspace{0.5cm}  &
\includegraphics[height=5.1cm]{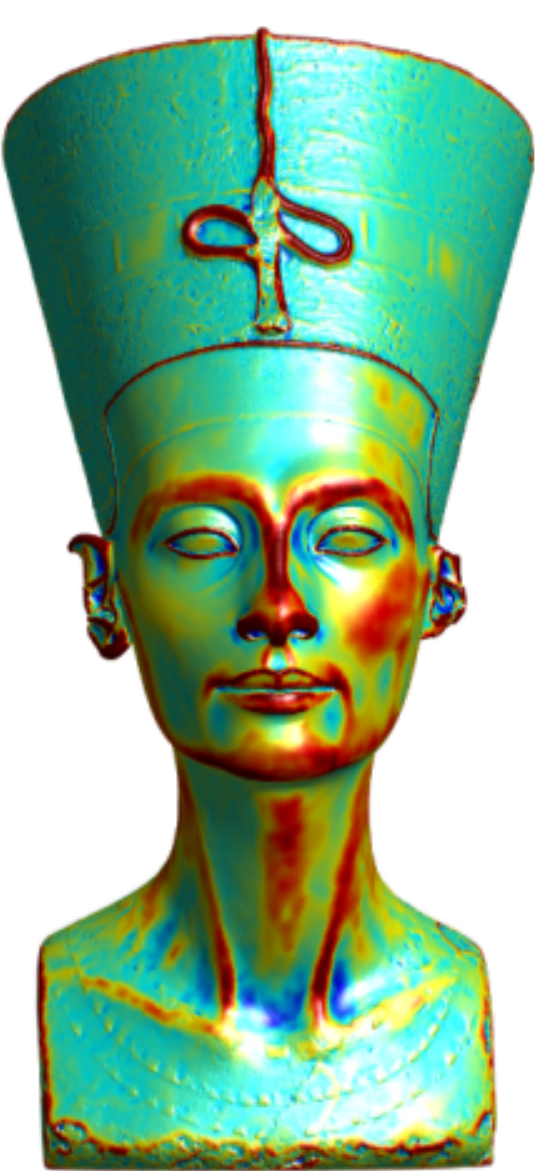}\\
(a) & & (b) & & (c)\\
\includegraphics[height=5.1cm]{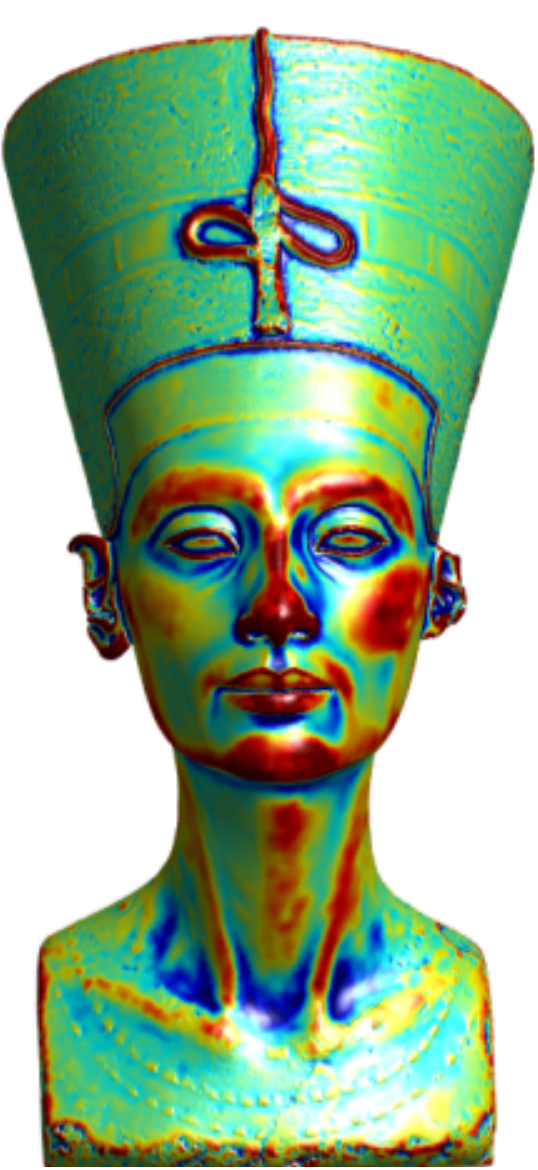} & \hspace{0.5cm}  &
\includegraphics[height=5.1cm]{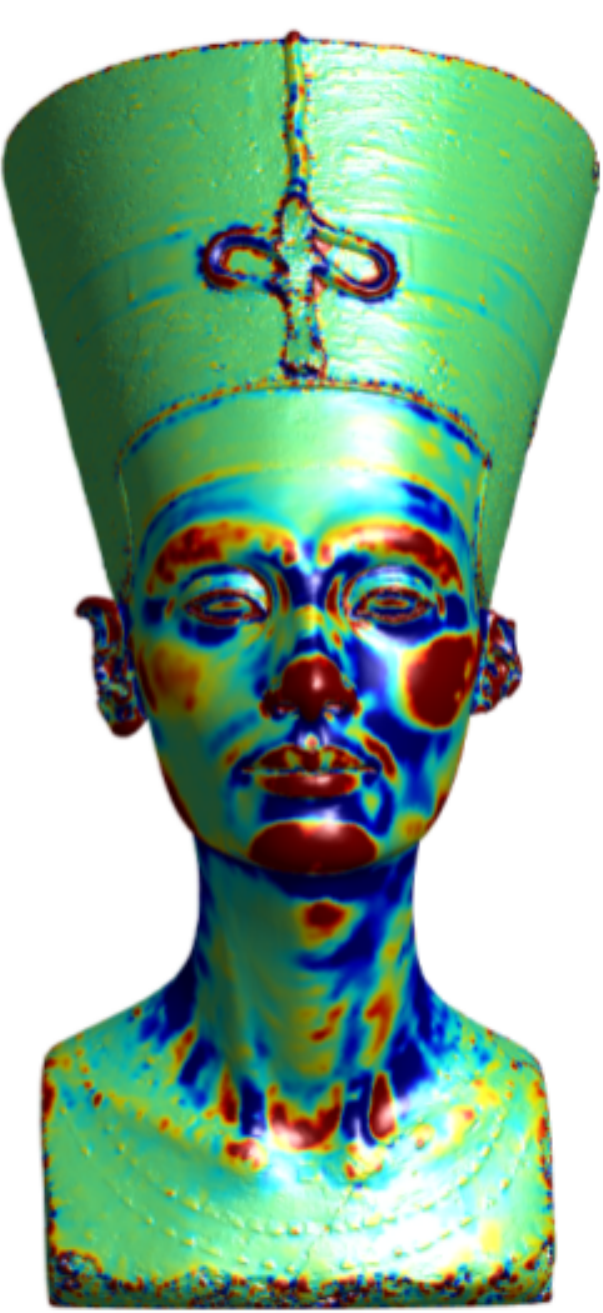} & \hspace{0.5cm}  &
\includegraphics[height=5.1cm]{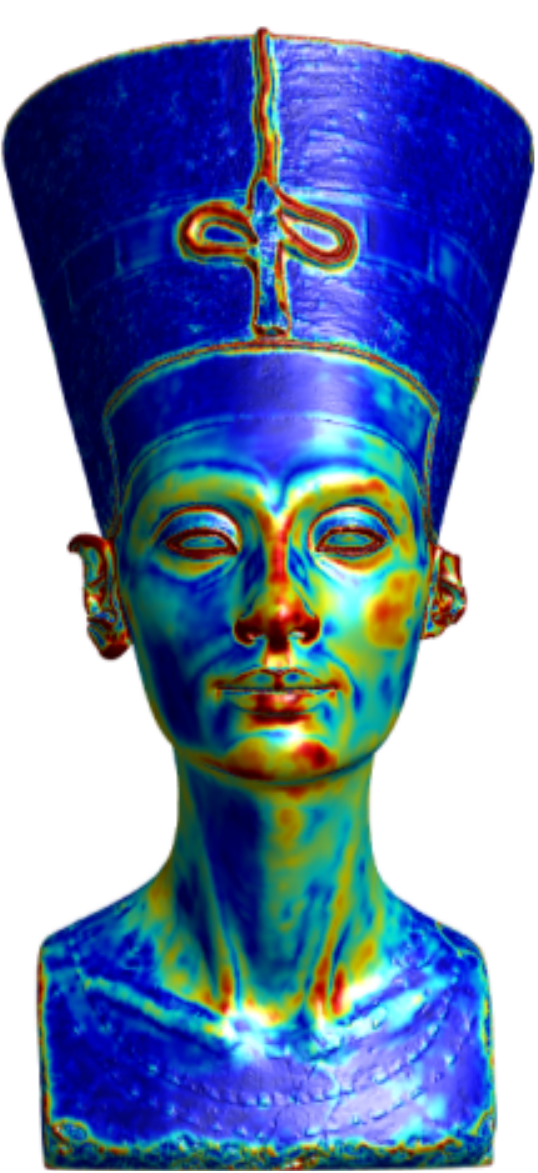}\\
(d) & & (e) && (f)
\end{tabular}
\caption{A 3D model $\mathcal M$ (a) and the visualization (colours range from blue (low) to red (high)) of the values of different curvatures: (b) minimum curvature, (c) maximum curvature, (d) mean curvature, (e) Gaussian curvature and (f) total curvature.}\label{curvatures}
\end{center}
\end{figure}

As an illustrative example we show the steps performed by the extraction of the feature points, when applied to a 3D model $\mathcal M$ given as a triangulated mesh (made up of approximately $10^6$ vertices  and $2\cdot 10^6$ faces) without photometric properties, see Figure \ref{curvatures}(a).

\begin{algorithm}
\SetAlgorithmName{Feature Points Recognition Algorithm}
\DontPrintSemicolon
\SetAlgoLined
\SetKwInOut{Input}{Input}
\SetKwInOut{Output}{Output}
\SetKwInOut{return}{return}
\Input{3D model $\mathcal M$, the property $prop$, that is the type of curvature ($C_{min}$, $C_{max}$, $C_{mean}$, $C_{Gauss}$, $C_{tot}$) and/or the $L$ channel, the filtering threshold $p$}
\Output{Set of feature points $\mathbb X \subset \mathbb R^3$}
   
 \Begin{  
 \tcc{loads $\mathcal M$; faces are optional (available only for meshes)}
[vertices, faces]= {\em load}($\mathcal M$)\; 
\tcc{evaluates curvatures and colour (if available)}
\eIf{$prop$ is a curvature}{
[Cmin, Cmax] = {\em EvaluateCurvatures}(vertices, faces)\;
\Switch{$prop$}{
{\bf case} $C_{mean}$ {\bf do} Cmean=(Cmin+Cmax)/2\;
{\bf case} $C_{Gauss}$ {\bf do} CGauss=Cmin.*Cmax\;
{\bf case} $C_{tot}$ {\bf do} Ctot = {\em abs}(Cmin)+ {\em abs}(Cmax)\;
}
}
{
[L, a, b] = {\em get$\_$Lab}($\mathcal M$)\;
}
$f \gets$ function according to the property $prop$\;
\tcc{builds the histogram of $f$ and filters it using $p$} 
h = {\em histogram}(f)\;
v = {\em Filtering}(h, p) (see Filtering Algorithm \ref{algFilter} in Appendix)\;

\tcc{constructs the feature points set $\mathbb X$}
\For{$j=1\ldots${\em size}(vertices)}{
    {\bf if} f[j] $>$ v {\bf then} add vertices[j] to $\mathbb X$\;
    }
\return{$\mathbb X$}
}
\caption{Extracts the feature points sets $\mathbb X \subset \mathbb R^3$ from the input model $\mathcal M$}
\label{algFeaturePoints}
\end{algorithm}

\begin{figure}[htb]
\begin{center}
\begin{tabular}{ccccc}
\includegraphics[height=2.5cm]{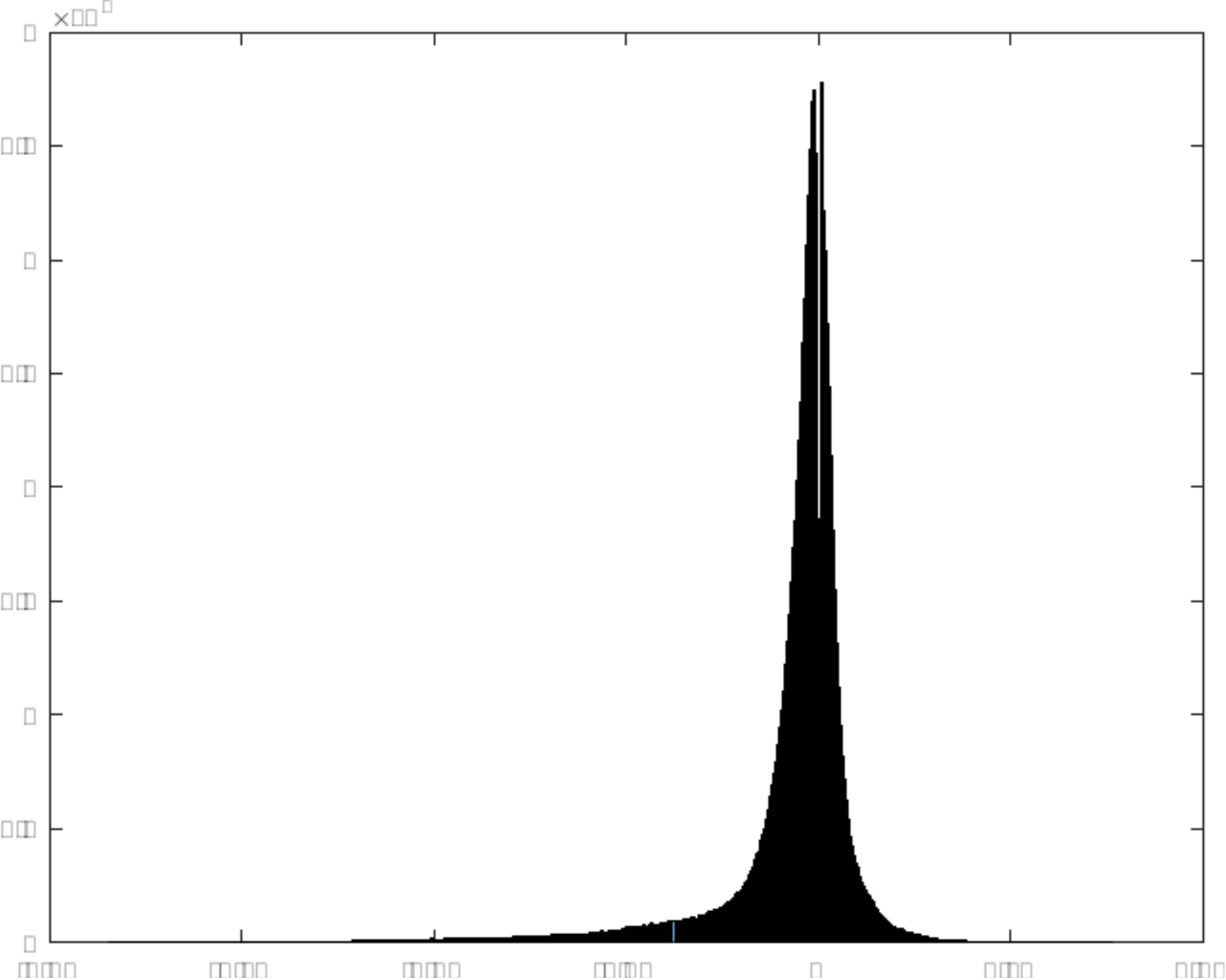} & \hspace{0.5cm} &
\includegraphics[height=2.5cm]{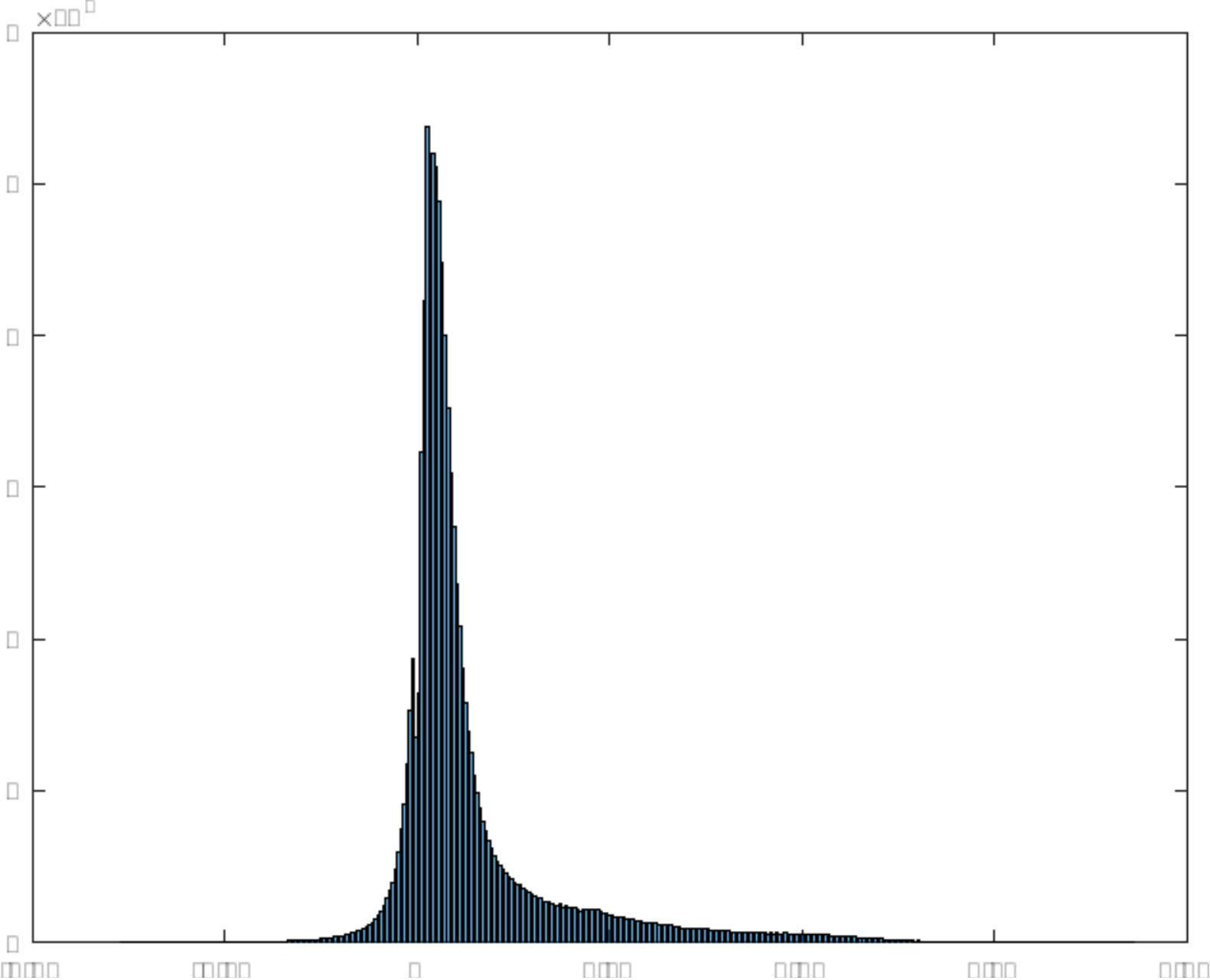} & \hspace{0.5cm}  &
\includegraphics[height=2.5cm]{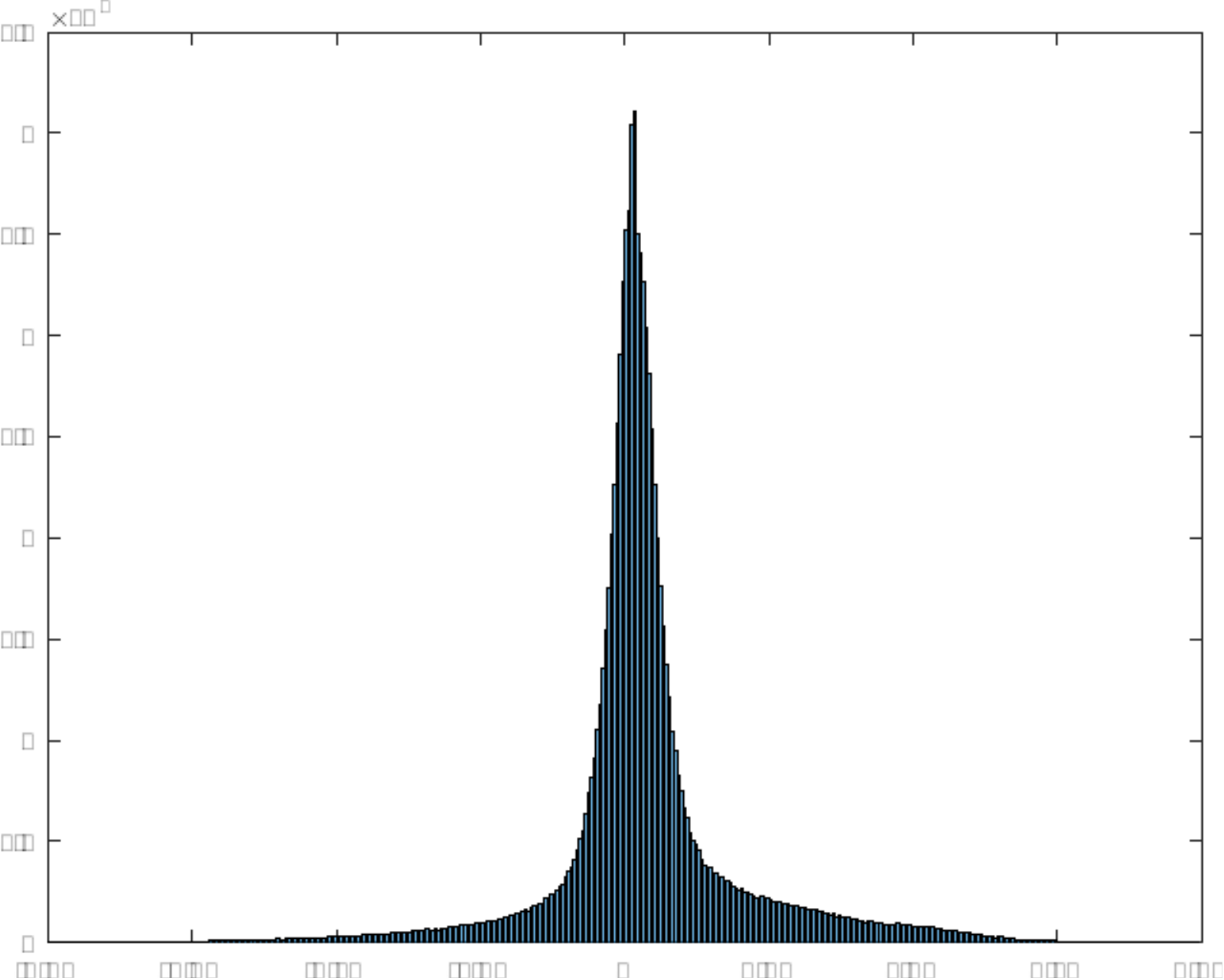} \\
(1) - min curvature & & (2) - max curvature & & (3) - mean curvature\\
&& && \\
\includegraphics[height=2.5cm]{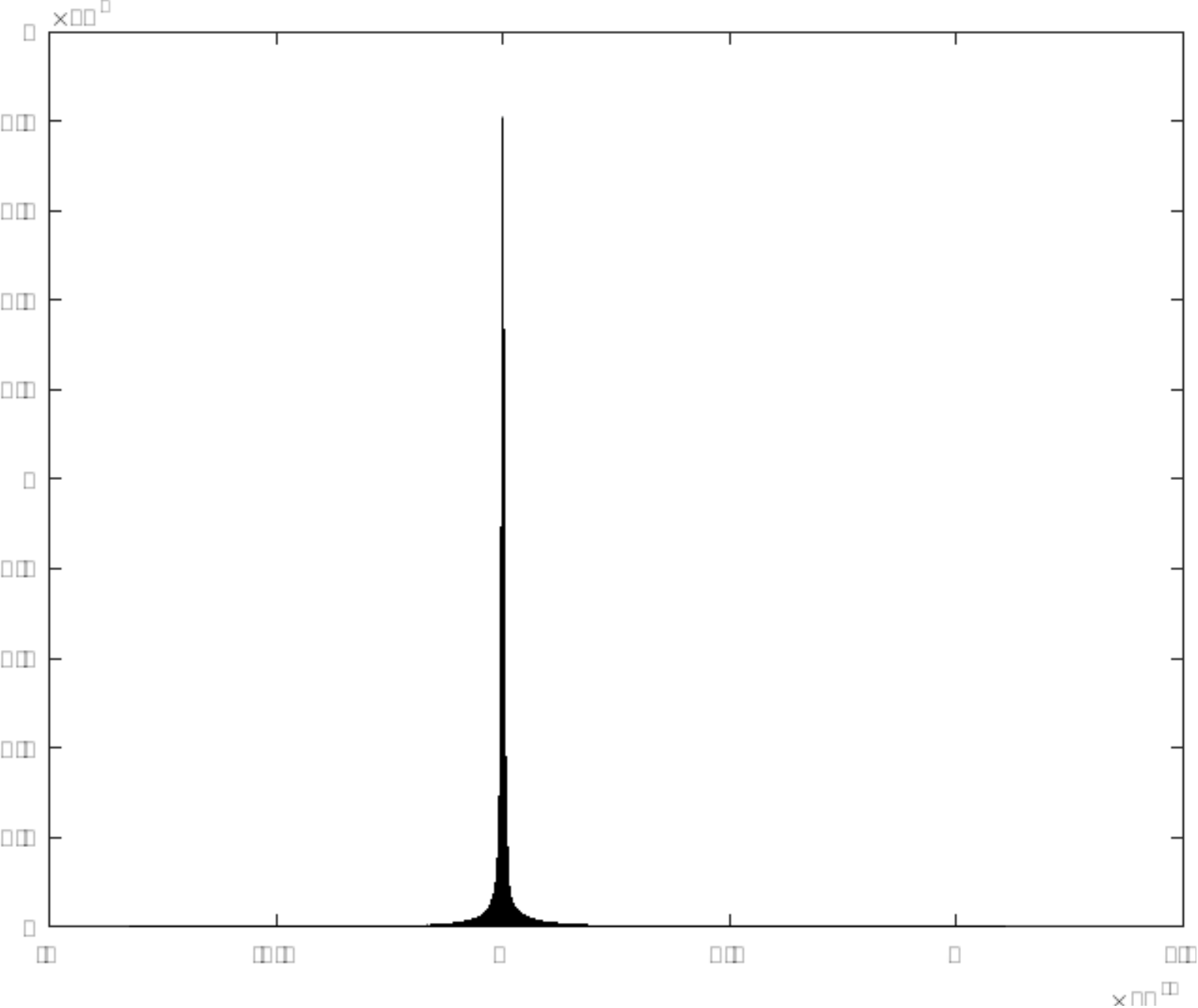} &\hspace{0.5cm}  &
\includegraphics[height=2.5cm]{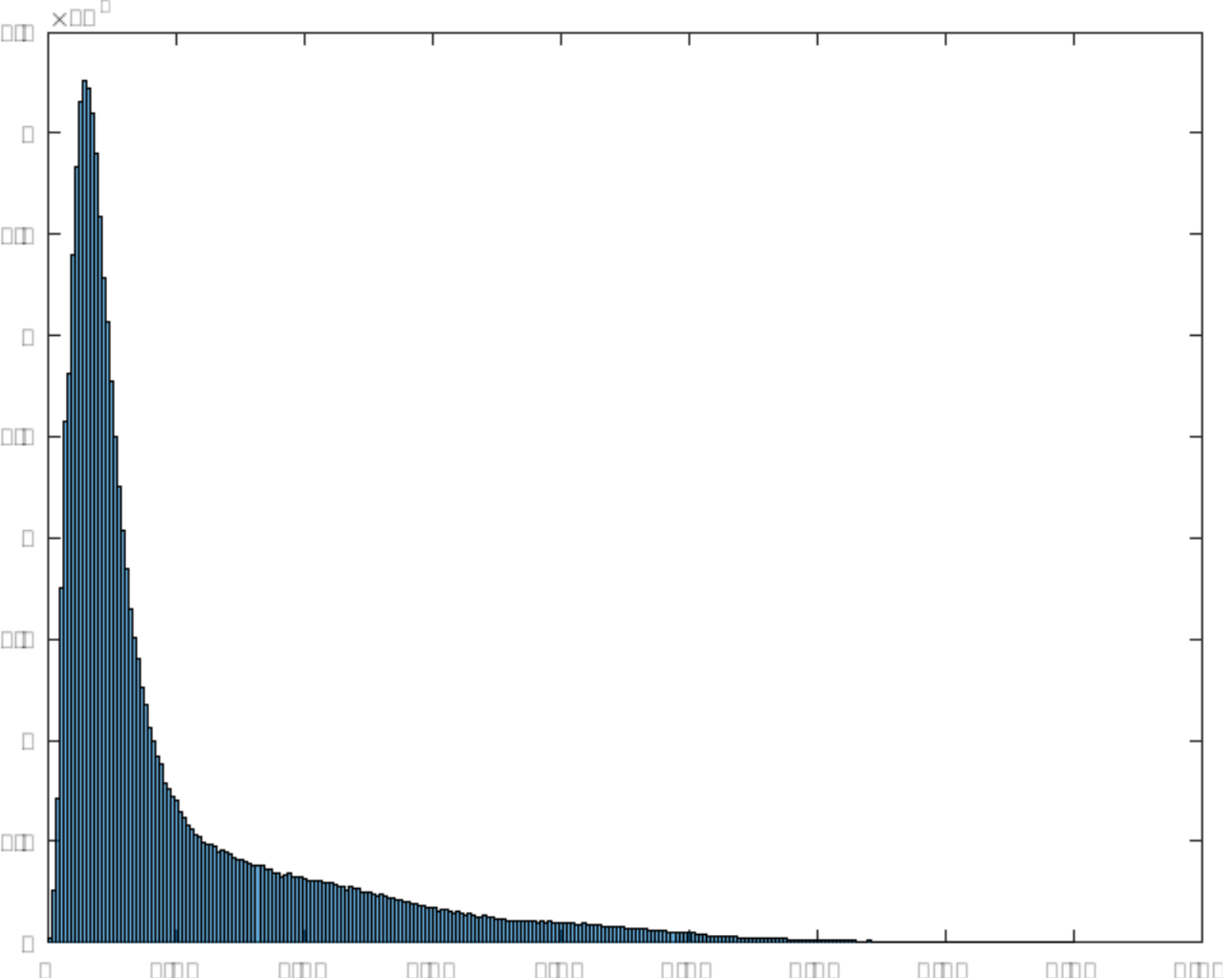} & \hspace{0.5cm} & \\
(4) - Gaussian curvature & & (5) - total curvature
\end{tabular}
\caption{
Histograms of different curvature functions on the 3D model $\mathcal M$ in Figure \ref{curvatures}(a)}\label{histograms}
\end{center}
\end{figure}

\begin{figure}[htb]
\begin{center}
\begin{tabular}{ccc}
\includegraphics[height=3.25cm]{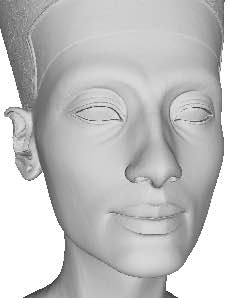} & 
\includegraphics[height=3.25cm]{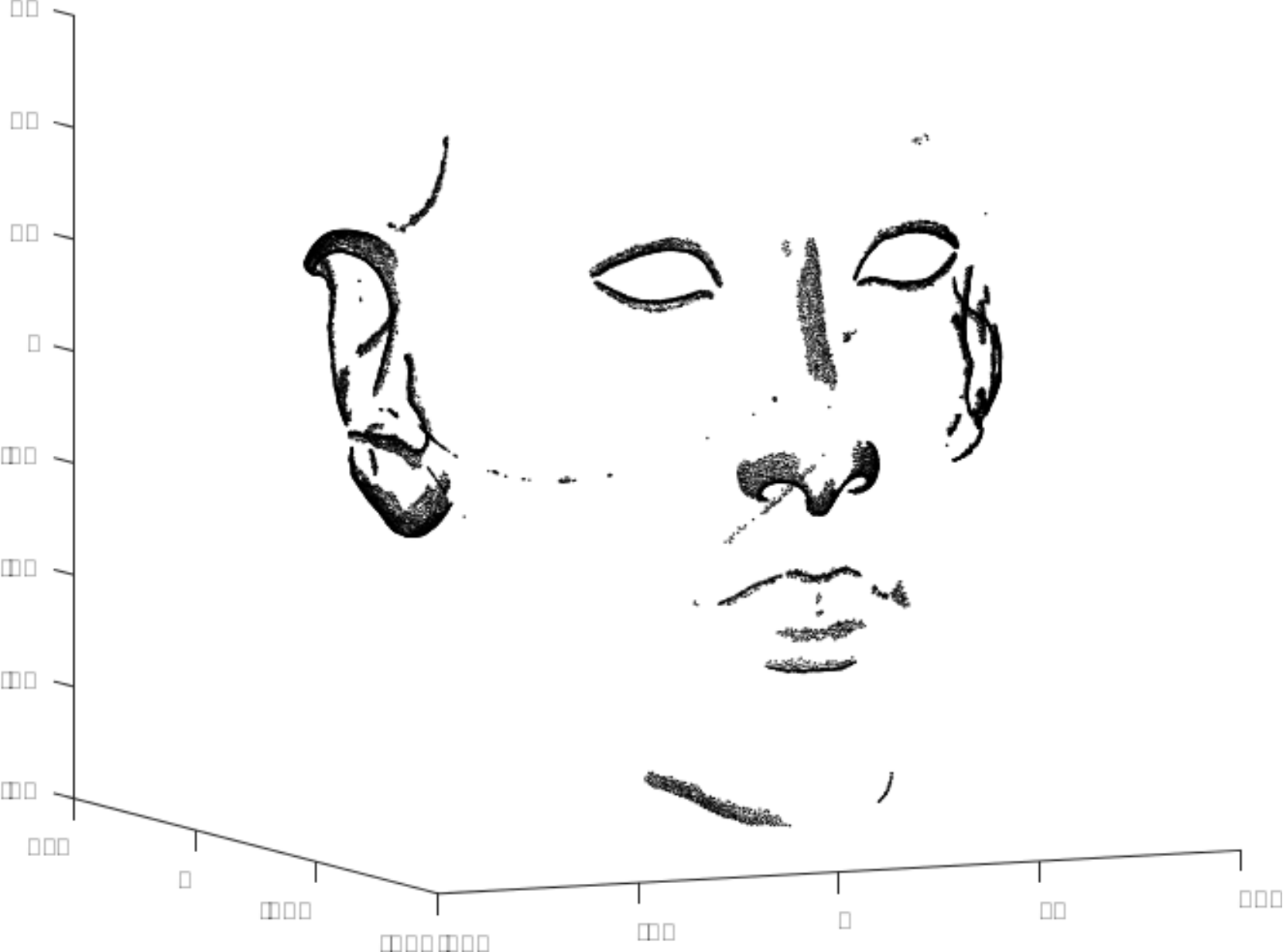} &
\includegraphics[height=3.25cm]{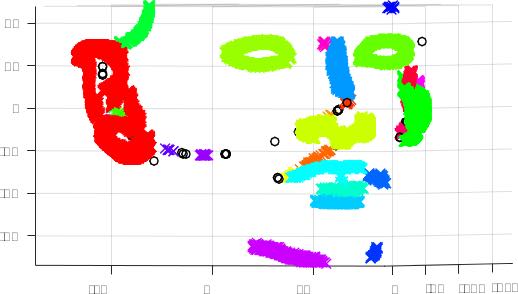}\\
(a) & (b) & (c)
\end{tabular}
\caption{(a) The model $\mathcal M$, (b) the feature points $\mathbb X$ resulting from Algorithm \ref{algFeaturePoints} 
applied to $\mathcal M$ with parameters $prop=2$ and $p=0.9$, and (c) the groups obtained applying Algorithm \ref{algClustering}
to $\mathbb X$  with parameters $K=50$ and $MinPts=5$.}\label{featurePts}
\end{center}
\end{figure}

A visual representation with colours of the different curvature functions computed on the model $\mathcal M$ is shown in Figure \ref{curvatures}(b)-(f); their histograms are shown in Figure \ref{histograms}. In Figure \ref{featurePts}(b) the output of  Algorithm \ref{algFeaturePoints} when applied to $\mathcal M$ using the maximum curvature and filtering the corresponding histogram at $90\%$ is shown.

Once detected, the set $\mathbb X$ of feature points is subdivided into smaller clusters (that is, groups of points sharing 
some similar properties, such as curvature values and/or chromatic attributes) 
by using classical methods of cluster analysis. Here, we adopt a very well-known density model, 
the {\em Density-Based Spatial Clustering 
of Applications with Noise} (DBSCAN) method \cite{EKS*96}, which groups together points that lie closeby
marking as outliers isolated points in low-density regions.
The DBSCAN algorithm requires two additional parameters: a real positive number $\varepsilon$, the threshold 
used as the radius of the density region, and a positive integer $MinPoints$, the minimum number of points 
required to form a dense region. 
In order to estimate the density of the feature set $\mathbb X$ necessary to automatically 
relate the choice of the threshold $\varepsilon$ to the context, we use the {\em K-Nearest Neighbor} (KNN),
a very efficient non parametric method that computes the $k$ closest neighbors of a point in a given dataset \cite{FBF77}.

We sum up the described procedure in the Appendix (Algorithm \ref{algClustering}). Figure \ref{featurePts}(c) represents, with different colours, the outcome of the aggregation algorithm when applied to the feature points $\mathbb X$ in Figure \ref{featurePts}(b).

\subsection{Projection of the feature points sets onto best fitting planes}\label{PointCloudProj}
This step performs a local projection of each feature points set to its best fitting plane. 
This operation does not represent a strong restriction
on our method since we apply such a projection separately to each group (obtained from the aggregation of the feature points) and also because we have already assumed that the kind of features that we are 
looking for can be locally projected onto a plane.
Let $\mathbb Y$ be a $3$-dimensional set of points that can be injectively projected onto a plane.
During this phase the following operations are performed: firstly, the points of the set $\mathbb Y$ 
are shifted to move their centroid onto the origin. Secondly, for the points of~$\mathbb Y$ a best fitting plane $\Pi$ 
is found by computing the multiple linear regression using the least squares method. 
We compute it as follows. We denote by $s$ the cardinality of $\mathbb Y$, and we consider 
the matrix $XY$ of size $s \times 2$ and the column vector $Z$ of size $s \times 1$ whose columns 
respectively contain the $x$-and $y$-coordinates of the points of $\mathbb Y$ and the
the $z$-coordinates of the same points. We apply a {\em linear regression} function to the pair $(XY, Z)$
and get the real values $b_1$ and $b_2$ used to construct the best fitting plane $\Pi$, whose equation 
is $\Pi: z - b_1x - b_2y=0$. 
Finally, the orthogonal transformation $\varphi$ moving the plane $\Pi$ onto the plane $z=0$ is defined 
and applied to the points of $\mathbb Y$ to get the new set $\mathbb Z$. 
We sum up the described procedure in the Appendix, Algorithm \ref{algProjection}. Figure \ref{projection}(b) represents the best fitting plane of the subset $\mathbb Y_8$ depicted in Figure \ref{projection}(a); the new set $\mathbb Z_8$ is shown in Figure \ref{projection}(c).

%
%
%

\begin{figure}[htb]
\begin{center}
\begin{tabular}{ccc}
\includegraphics[height=2.75cm]{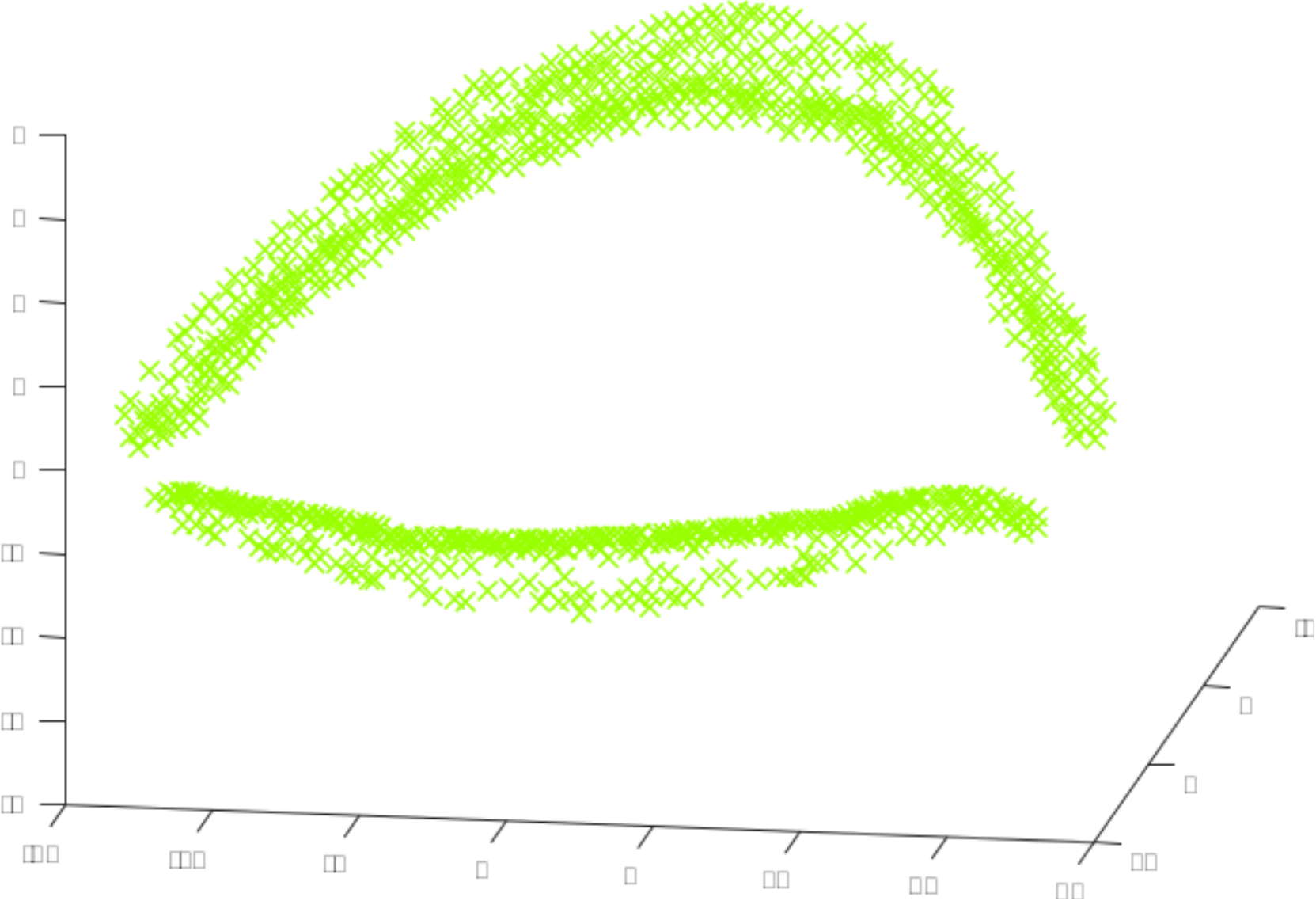} & 
\includegraphics[height=2.75cm]{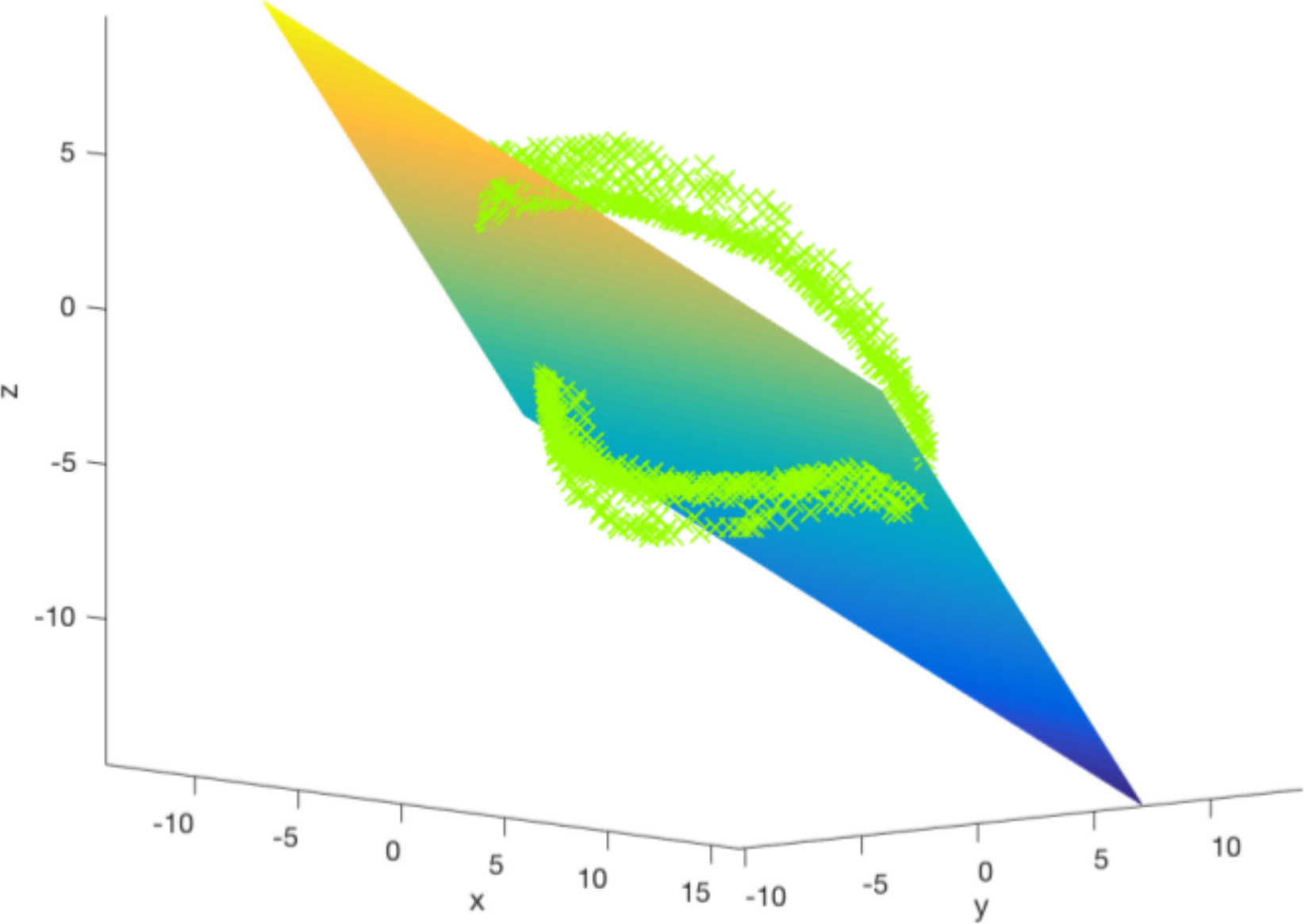}& 
\includegraphics[height=2.75cm]{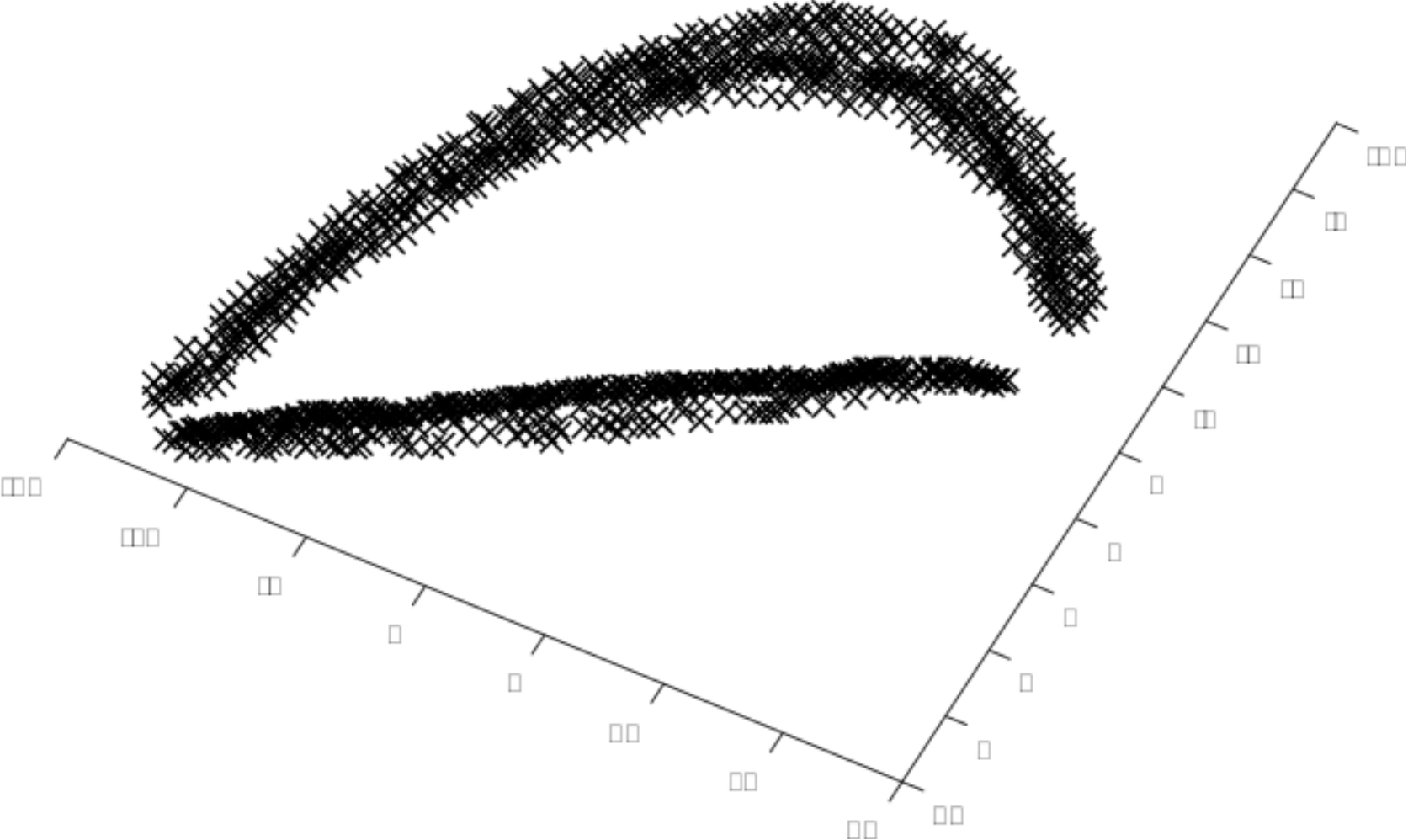}\\
(a) & (b) & (c)
\end{tabular}
\caption{Representation of (a) the cluster $\mathbb Y_8$ (see also Figure \ref{featurePts}(c)), (b) its best fitting plane $\Pi_8$ 
and the subset $\mathbb Z_8$ obtained by projecting $\mathbb Y_8$ onto $\Pi_8$.}\label{projection}
\end{center}
\end{figure}

\subsection{Feature curve approximation}\label{Hough}
Once we have projected every set of points onto its best fitting plane, we apply to the single group elements 
the generalization of the HT technique (see \cite{beltrametti2013hough} and \cite{torrente2014almost}) 
aimed at approximating the feature curve.
With respect to the previous application of \cite{beltrametti2013hough} to images, our approach is novel since we apply the method to each projected 3D feature points set by exploiting a vast catalogue of curves (see Section \ref{Sec:curves}
for a detailed description of the families of curves used in this paper).
Further, we propose a method that does not undergo to any grid approximation of the coordinates of the given points. 

We provide here some details on the curve detection algorithm which is the core of the HT-based technique.
In its classical version \cite{c1962method, DH72}, HT permits to recover the equation of a straight line exploiting a very 
easy mathematical principle:
starting from points lying on a straight line, defined using the common
Cartesian coordinates $x$ and $y$ (or equivalently the polar coordinates $\rho$ and~$\theta$)
and following the usual slope-intercept parametrization with parameters $a$ and $b$
we end with a collection of straight lines in the parameters' space (or equivalently sinusoidal curves
in the variables $\rho$ and~$\theta$) that all intersect in exactly one point. 
The equation of each straight line is also called \emph{Hough transform} of the point, usually 
denoted by $\Gamma_p(\mathcal F)$ (where $\mathcal F$ denotes the chosen family of curves, the straight
lines in this particular case). The coordinates of the unique intersection point of all the Hough transforms identify 
the original line.

Following the approach of \cite{beltrametti2012algebraic}, we consider a collection of curves which includes more complex shapes beyond the commonly used lines, e.g. circles and ellipses 
(see Section \ref{Sec:curves}).
Under the assumption of Hough regularity on the chosen family of curves~$\mathcal F$ (see \cite{beltrametti2013hough}),
the detection procedure can be detailed as follows. 
Let $p_1,\ldots, p_s$ be the set of feature points; in the parameter space we find the (unique) intersection of the Hough 
transforms (the hypersurfaces depending on the parameters) $\Gamma_{p_1}(\mathcal F), \ldots, \Gamma_{p_s}(\mathcal F)$ 
corresponding to the points $p_1,\ldots,p_s$; that is, we compute $\lambda = \cap_{i=1 \ldots s} \Gamma_{p_i}(\mathcal F)$.
Finally, we return the curve of the family $\mathcal F$ uniquely determined by the parameter~$\lambda$.

From a computational viewpoint, the burden of the outlined methodology is represented by the computation of the intersection of the Hough transforms, which is usually implemented using a so called ``voting procedure''. Following \cite{torrente2014almost} the voting procedure can be detailed in the following four steps:

\begin{enumerate}
\item \emph{Fix a bounded region $\mathcal T$ of the parameter space}.\\
This is achieved exploiting typical characteristics of the chosen family $\mathcal F$ of curves, like the bounding box properties or the presence of salient points. In Section \ref{Sec:curves} we detail how to derive the bounding box of the parameter space from the algebraic or polar representation of various families of curves.

\item \emph{Fix a discretization of $\mathcal T$}.\\
This is nontrivial yet fundamental for the detection result which is valid up to the chosen 
discretization step. The choice of the size of the discretization step is currently done as a percentage of the range interval of each parameter. For some recent results on this topic we refer to~\cite{TorrenteBeltramettiSendra}. 
Let $\mathcal T = [a_1,b_1] \times \ldots \times [a_t,b_t]$ be the fixed region of the parameters space 
and $d=(d_1,\ldots,d_t) \in \mathbb R_{>0}^t$ be the discretization step.
For each $k=1,\ldots,t$, we define 
\begin{equation}\label{Jkxk}
 J_k := \left \lceil \frac{b_k-a_k- \frac{d_k}{2}}{d_k} \right \rceil + 1 ,   
\end{equation}
where $\lceil x \rceil = \min\{z \in \mathbb N \:|\: z  \ge x\}$ and
\begin{equation}
\lambda_{k,j_k} :=a_k + j_k d_k
\end{equation}\label{lambdakjk}
with $j_k=0,\ldots, J_k-1$. 
Here $J_k$ denotes the number of considered samples for each component, and $j_k$ the index of the sample.
We denote by ${\bf j}$ the multi-index $(j_1,\ldots, j_t)$,
by ${\lambda}_{\bf j}:=(\lambda_{1,j_1},\ldots, \lambda_{t, j_t})$ the ${\bf j}$-th {\em sampling point}, 
and by
\begin{eqnarray}\label{cells}
{\bf C}({\bf j}):=\left\{( \lambda_1,\ldots, \lambda_t ) \in \mathbb R^t \; \Big| \;  
\lambda_k \in \left[\lambda_{k,j_k}-\frac{d_k}{2}, \lambda_{k,j_k}+\frac{d_k}{2}\right),\; k=1,\ldots,t \right\} 
\end{eqnarray}
{\em the cell} centered at (and represented by) the point ${\lambda}_{\bf j}$.
The discretization of $\mathcal T$ is given by the $J_1\times \cdots \times J_n$ cells 
of type ${\bf C}({\bf j})$ which are  a covering of the region $\mathcal T$.

\item \emph{Construct an accumulator function}
$\mathcal A: \mathcal T \rightarrow \mathbb{N}$, which is defined as 
$\mathcal A=\sum_{1=1 \ldots s} f_{p_i}$ where $f_{p_i}: \mathcal T \rightarrow \mathbb{N}$ is defined as follows:
\[ 
f_{p_i}(c)=\left\{ \begin{array}{ll}
1 & \mbox{if the Hough Transform } \Gamma_{p_i}(\mathcal F)\ \mbox{crosses the cell}\ c;\\
0 & \mbox{otherwise}.
\end{array} \right.
\]
For the evaluation of each $f_{p_i}$ we adopt the \emph{Crossing Cell} algorithm (detailed in Algorithm \ref{algCrossingCell})
which is based on bounds of the evaluation of $f_{p_i}$ theoretically proved in \cite{torrente2014almost}. The function $EvalFirstBound$, resp. $EvalSecondBound$, in the pseudocode represents the evaluation of the bound $B_1$ (resp. $B_2$) of a given polynomial $f$ in $n$ variables w.r.t a cell centered at $p$ with radius $\varepsilon$.
Such bounds depend on the Jacobian and the Hessian matrices of $f$, denoted by ${\rm Jac}_f$ and $H_f$
respectively. They are defined as follows:
\begin{eqnarray}
B_1 &=& \|{\rm Jac}_f(p)^t\|_1 \varepsilon + \frac{n}{2} H \varepsilon^2\label{B1}\\
B_2 &=& \frac{2R}{{\rm J}(c + n^{5/2}H{\rm J}R)}\label{B2}
\end{eqnarray}
where $H=\max_{\{x \in \mathbb R^n : \|x-p\|_\infty \le \varepsilon\}} \|H_f(x)\|_\infty$,  
$R<\min \Big\{\varepsilon, \frac{\|{\rm Jac}_f(p)\|_1}{H} \Big\}$, $c=\max\{2, \sqrt{n}\}$ and 
${\rm J} = \sup_{\{x \in \mathbb R^n : \|x-p\|_\infty < R\}} \|{\rm Jac}_f^\dagger(x)\|_\infty$,
with ${\rm Jac}_f^\dagger$ denoting the Moore-Penrose pseudo-inverse of ${\rm Jac}_f$.
Since the above quantities depend on the Jacobian and the Hessian matrices, $p$ must be a point for which these values are non-trivial. Note that the evaluation is made over a symbolic representation of the Jacobian matrix, its pseudo-inverse 
and the Hessian matrix. This implies that the computational cost of this operation is constant while their symbolic representation is computed only once, in the overall Curve Detection Algorithm \ref{algHough}. 
In our implementation we use the system CoCoA~\cite{CoCoA-5} for the symbolic manipulation of polynomials and matrices.

\item \emph{Identify the cell corresponding to the maximum value of the accumulator function and return the coordinates of its center}.\\
Following the general theory, the HT regularity guarantees that the maximum of the accumulator function is unique. In addition, being based on local maxima, the voting strategy permits to identify curves also from partial and incomplete data, even if the missing part is significant, see the examples in Figure \ref{partial}(I.c-d) and in Figure \ref{partial}(II.c-d).  
\end{enumerate}

\begin{algorithm}
\SetAlgorithmName{Crossing Cell Algorithm}
\DontPrintSemicolon
\SetAlgoLined
\SetKwInOut{Input}{Input}
\SetKwInOut{Output}{Output}
\SetKwInOut{return}{return}
\Input{The symbolic representation of $f$ in the variables $x_1,\ldots,x_n$, a point $p \in \mathbb R^n$, 
a tolerance $\varepsilon$, the symbolic expression of the Jacobian ${\rm Jac}_f$ of $f$, the Moore-Penrose pseudo-inverse ${\rm Jac}_f^\dagger$ 
of ${\rm Jac}_f$, and the Hessian matrix $H_f$ of $f$.}
\Output{an element of $\{0, 1$, \emph{undetermined}$\}$}

\Begin{
$B_1 \gets EvalFirstBound(f, p, \varepsilon, {\rm Jac}_f)$ (see formula (\ref{B1}))\;
$B_2 \gets EvalSecondBound(f, p, \varepsilon, {\rm Jac}_f^\dagger, H_f)$ (see formula (\ref{B2}))\;
\eIf{$abs(f(p)) > B_1$}{returnValue=0}{
{\bf if} $abs(f(p)) < B_2$ {\bf then} returnValue=1\\ {\bf else} returnValue= {\emph{undetermined}} {\bf end}
}
\return{returnValue}
}
\caption{Returns $1$ or $0$ if $f=0$ crosses or not the cell of radius $\varepsilon$ centered at $p$. 
If not decidable it returns \emph{undetermined}.}\label{algCrossingCell}
\end{algorithm}

The Curve Detection Algorithm \ref{algHough} sums up the curve detection's procedure described in the previous steps $1$-$4$.
Note that a prototype implementation in CoCoA of the Curve Detection Algorithm \ref{algHough} is freely available\footnote{http://www.dima.unige.it/~torrente/recognitionAlgorithm.cocoa5}.

\begin{figure}[htb]
\begin{center}
\begin{tabular}{ccc}
\includegraphics[width=0.35\linewidth]{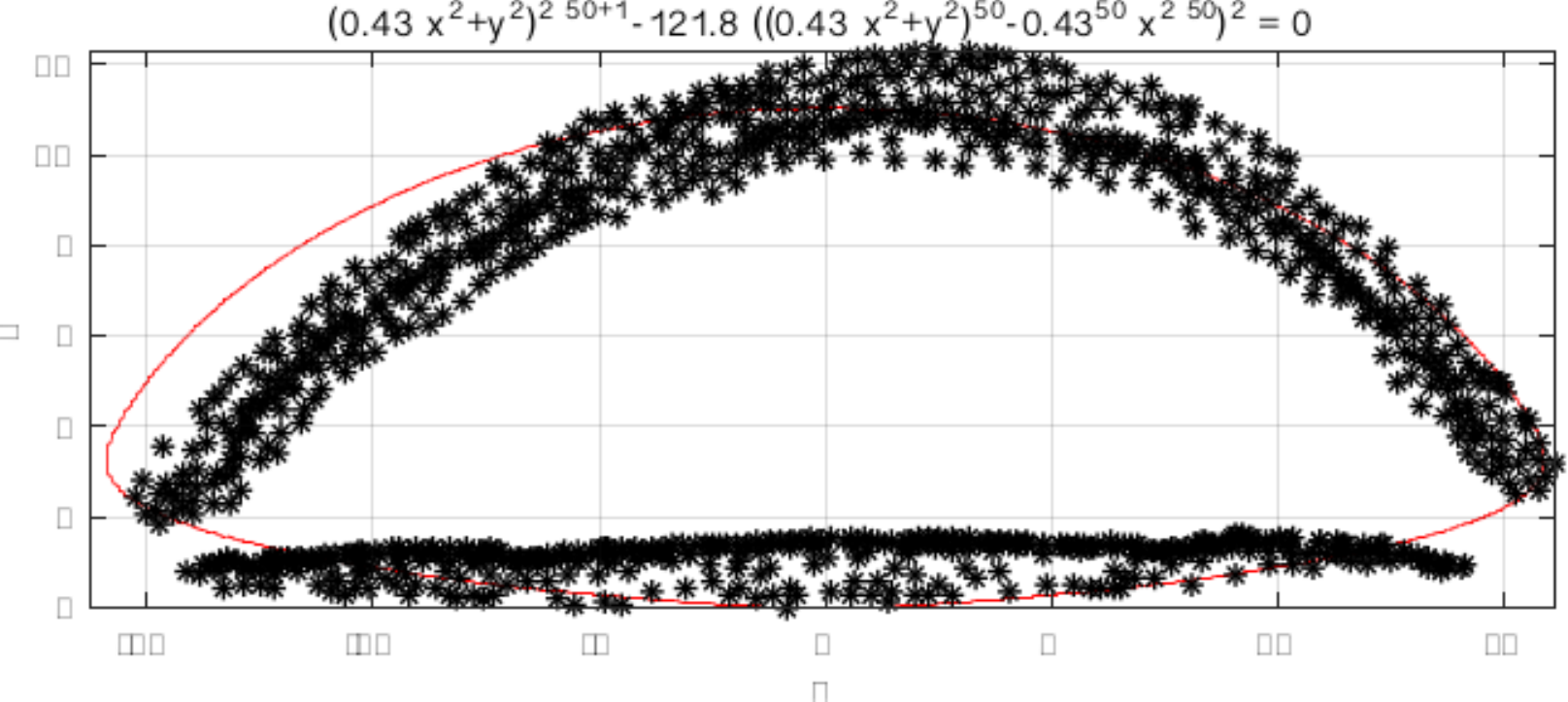} &
\hspace{1cm} &
\includegraphics[width=0.2\linewidth]{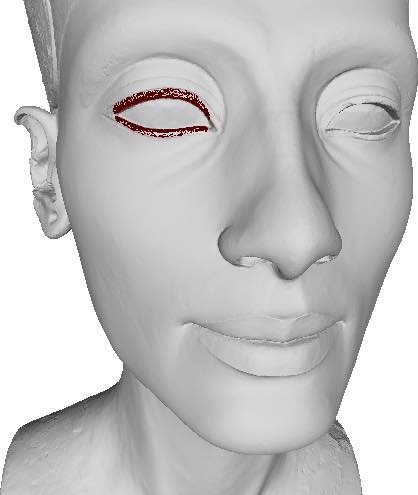}\\
(a) & & (b)
\end{tabular}
\caption{Representation of the set $\mathbb Z_8$ and the geometric petal curve (in red) computed by Algorithm 
\ref{algHough} (a) and visualization of the curve (in red) on the original model of Figure \ref{curvatures}(b).}
\label{nefertitiFinale}
\end{center}
\end{figure}

To give an idea of the outcome of the Algorithm \ref{algHough}, we have run it on the set $\mathbb Z_8$
(represented in Figure~\ref{projection}(c)) made of $981$ points.
In this example we use the {\em geometric petal} curve, an HT-regular family of curves presented both in polar and 
in cartesian form, see Section \ref{Sec:curves} for details.
By exploiting the bounding box of the set $\mathbb Z_8$ and properties of the geometric petal curve, we consider 
$\mathcal T =  [121,122] \times [0.43, 0.45]$ and $d = (0.025, 0.005)$.
The Algorithm \ref{algHough} applied to this framework computes the geometric petal curve depicted in 
Figure \ref{nefertitiFinale}(a); the feature curve is shown on the 3D model as the red line in Figure \ref{nefertitiFinale}(b).

\begin{algorithm}
\SetAlgorithmName{Curve Detection Algorithm}
\DontPrintSemicolon
\SetAlgoLined
\SetKwInOut{Input}{Input}
\SetKwInOut{Output}{Output}
\SetKwInOut{return}{return}
\Input{A finite set $\mathbb X =\{p_1,\ldots, p_s\} \subset \mathbb R^2$, an HT-regular family 
$\mathcal F=F(x,y, \lambda_11, \ldots, \lambda_t)$ of curves, 
a region $\mathcal T=[a_1,b_1]\times\ldots \times [a_t,b_t] \subset \mathbb R^t$ of the parameter space, 
a discretizaton step $d=(d_1,\ldots,d_t) \in \mathbb R_{>0}^t$}
\Output{a curve $\mathcal C$ of the family $\mathcal F$}
   
\Begin{  
\tcc{discretizes the region $\mathcal T$ with step $d$}
Initialize $J_k$, with $k=1,\ldots t$, (see formula (\ref{Jkxk}))\;
Initialize $\lambda_{k,j_k}$, with $k=1,\ldots t$ and $j_k=0, \ldots, J_k-1$ (see formula (\ref{lambdakjk}))\;
Initialize $\lambda_{\textrm{\bf j}}$ and $C(\textrm{\bf j})$, with $\textrm{\bf j} \in J_1\times \ldots \times J_t$ 
(see formula (\ref{cells}))\;
\tcc{constructs of the multi-matrix $\mathcal A$}
$\mathcal A \gets$ zero matrix of size $J_1\times \ldots \times J_t$;\\
$Jac \gets$ Jacobian matrix of $F(x,y, \lambda_1, \ldots,\lambda_t)$ w.r.t $\lambda_1, \ldots,\lambda_t$\;
${\rm Jac}_f^\dagger \gets$ Moore-Penrose pseudo-inverse of $Jac$\;
$H_f \gets$ Hessian matrix of $F(x,y, \lambda_1, \ldots,\lambda_t)$ w.r.t $\lambda_1, \ldots,\lambda_t$\;
\For{$i=1,\ldots,s$}{
\For{each $\textrm{\bf{j}} \in J_1\times \ldots \times J_t$}{
$\mathcal A(\textrm{\bf{j}}) \gets \mathcal A(\textrm{\bf{j}}) + 
\textrm{Crossing Cell}(F(p_i), \lambda_{\textrm{\bf{j}}}, \frac{d}{2},{\rm Jac}_f(p_i),{\rm Jac}_f^\dagger(p_i),H_f(p_i))$  (Alg. \ref{algCrossingCell})
}
}
\tcc{computation of the maximum of $\mathcal A$}
${\bf \bar j} \gets \max(\mathcal A$);\\
\Return{$\mathcal C = \mathcal C(\lambda_{\bf \bar j})$, the curve of $\mathcal F$ with parameters $\lambda_{\bf \bar j}$}
}
\caption{Computes the curve $\mathcal C$ of the family $\mathcal F$ best detecting the profile 
 highlighted by the points of the set $\mathbb X$}\label{algHough}
\end{algorithm}

\begin{figure}
\begin{center}
\begin{tabular}{ccccccc}
  \includegraphics[width=0.18\linewidth]{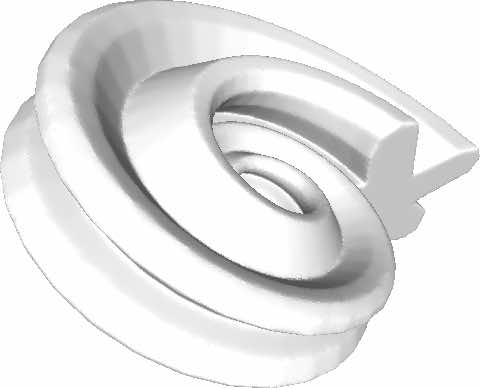} & &
   \includegraphics[width=0.2\linewidth]{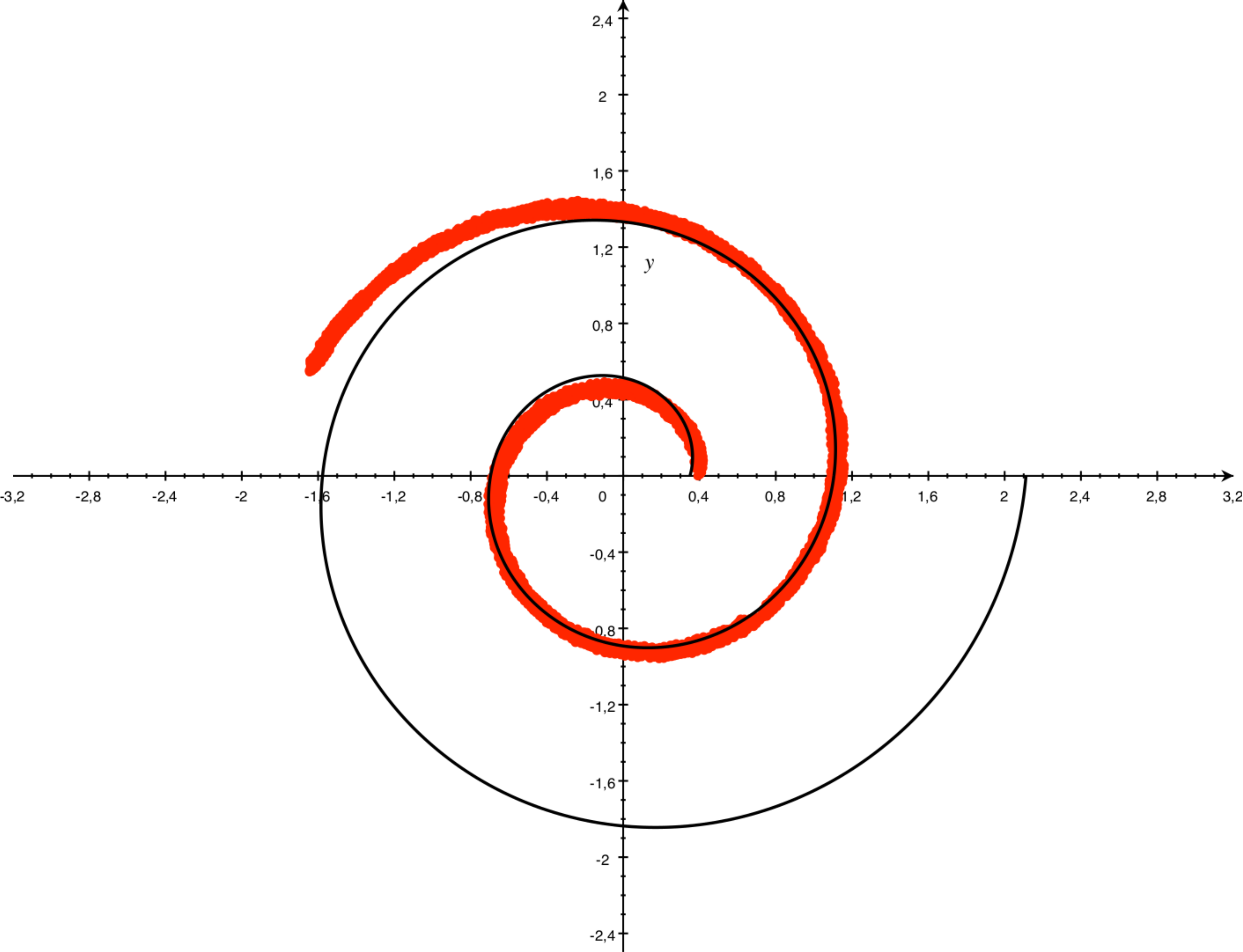} & &
 \includegraphics[width=0.18\linewidth]{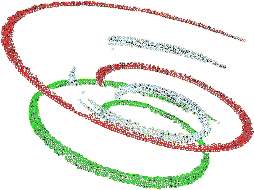} & &
  \includegraphics[width=0.18\linewidth]{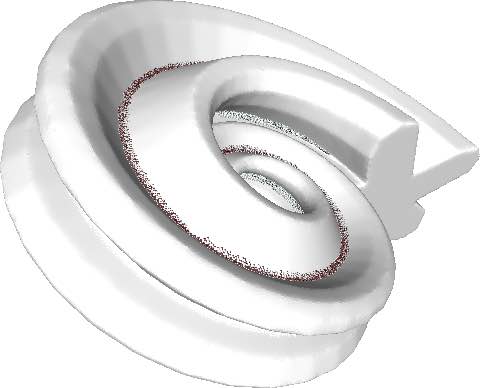}  \\
  (a) &  & (b) & & (c) & & (d)
  \end{tabular}
  \centering
   \caption{A 3D model (a), recognition of a spiral (b), two views of the spiral-like curves detected by our method (c,d).}
 \label{fig:spiral}
 \end{center}
 \end{figure}

We outline the steps of the Main Algorithm~\ref{mainAlg} 
when applied to the $3D$ model (made up of $152850$ vertices and $305695$ faces) represented in Figure~\ref{fig:spiral}(a). 
In this case, in order to detect the spiral-like curves we work with the generalized family~$\mathcal F$ of Archimedean spirals (see Section \ref{Sec:curves} for more details).
The Feature Points Recogniton Algorithm \ref{algFeaturePoints} is applied using the mean curvature 
and returns a set $\mathbb X$ made up of $12689$ points.
Then, the Aggregation Algorithm \ref{algClustering} (in Appendix) subdivides the points of $\mathbb X$ into $8$ groups  $\mathbb Y_1,\ldots, \mathbb Y_8$.
For shortness, we give details for the first group, $\mathbb Y_1$, which is made up of $2382$ points. 
Exploiting some geometrical properties of the Archimedean spirals (see again Section \ref{Sec:curves}), we fix and
discretize a suitable region of the parameter space, and apply the Curve Detection Algorithm \ref{algHough}.
The corresponding spiral curve is depicted in Figure \ref{fig:spiral}(b) and shown on the 3D model as the red line 
in Figure \ref{fig:spiral}(c-d). Other detected spirals are marked using different colours in Figure \ref{fig:spiral}(c-d).

\subsection{Computational complexity}\label{computComplexity}
Here we briefly analyse the computational complexity of the method. The feature points recognition function 
(Algorithm \ref{algFeaturePoints}) depends on the chosen procedure, in case we adopt the curvature estimation proposed 
in \cite{Cohen03} the computational complexity is $O(n\log n)$, where $n$ represents the number of points of the model. 
The computational cost for converting RGB coordinates into CIELab ones is linear ($O(n)$). Denoting $n_f$ the number of the feature points (in general $n_f \ll n$), the aggregation  into groups using DBSCAN 
(see the Algorithm \ref{algClustering} in the Appendix)  
takes $O(n{_f}^2)$ operations in the worst case \cite{EKS*96} (on average it takes $O(n_f \log n_f)$ operations, and thus 
it is overcome by the KNN search operation detailed in the Algorithm \ref{algClustering} that costs $O(kn_f \log n_f)$, see \cite{Friedman:1977}.  

After the aggregation of the feature points, the projection (Algorithm \ref{algProjection} in the Appendix) and the curve detection 
(Algorithm \ref{algHough}) algorithms are applied to each group, separately (note that the sum $s_f$ of the elements in 
the groups is, in general, smaller than $n_f$ because of the presence of outliers and small groups).

The cost of the curve detection algorithm is dominated by the size of the discretization of the region $\mathcal{T}$, as detailed in the Algorithm \ref{algHough}. Such a discretization consists of $M=\prod_{k=1}^t J_k$ elements, where $t$ is the number of parameters (in the curves proposed in this paper, $t=2,3$) and $J_k$ is the number of subdivisions for the $k$th
parameter, see formula (\ref{Jkxk}). The evaluation of the HT on each cell is constant and the Jacobian, pseudo-inverse and Hessian matrices are symbolically computed once for each curve; therefore, the cost of fitting a curve to each group with our HT method is $O(M)$.
Since the curve fitting is repeated for $m$ groups, the overall cost of our method is 
$O(\max (n \log n, ms{_f}^2, mM) )$ where $n$, $m$, $s_f$ and $M$ represent, respectively, the number of points of the 3D model, the number of groups of feature points and their maximum size, the size of the discrete space on which we evaluate the accumulator function.

\section{Families of curves}
\label{Sec:curves}
In this section we list the families of curves used in our experiments, focusing in each case on their main properties and characteristics (parameter dependency, boundedness, computation of the bounding box). For every family of curves we 
explicitly describe how to derive from its algebraic representation the parameters used as input for the Curve Detection Algorithm \ref{algHough}.
The selected curves form an atlas, which is both flexible and open: in fact, these curves are modifiable (for instance, 
by adding, stretching or scaling parameters) and the insertion of new families of curves is always possible. Note that the atlas comprises families of curves defined using Cartesian or polar coordinates, since our framework works on both cases.

Our collection also includes some elementary, but likewise interesting, families of algebraic curves like straight lines, 
circles and ellipses, which can be exploited to detect 
for instance eyes contours, pupils shapes and lips lines. Their equation are linear (straight lines) and 
quadratic (ellipses and circles) and depend on $3$ parameters at most.

In the following, we list the other families of the collection; these curves mainly come from~\cite{shikin1995handbook} that 
contains a rich vocabulary of curves many of which are suitable for our approach, too. The curves employed in our
experiments (see Section \ref{examples}) are:  the \emph{curve of Lamet}, the \emph{citrus curve}, the 
\emph{Archimedean spiral}, the {curve with $m$-convexities} and the \emph{geometric petal} curve.

\begin{figure}[htb]
\begin{tabular}{ccc}
\includegraphics[width=0.3\linewidth]{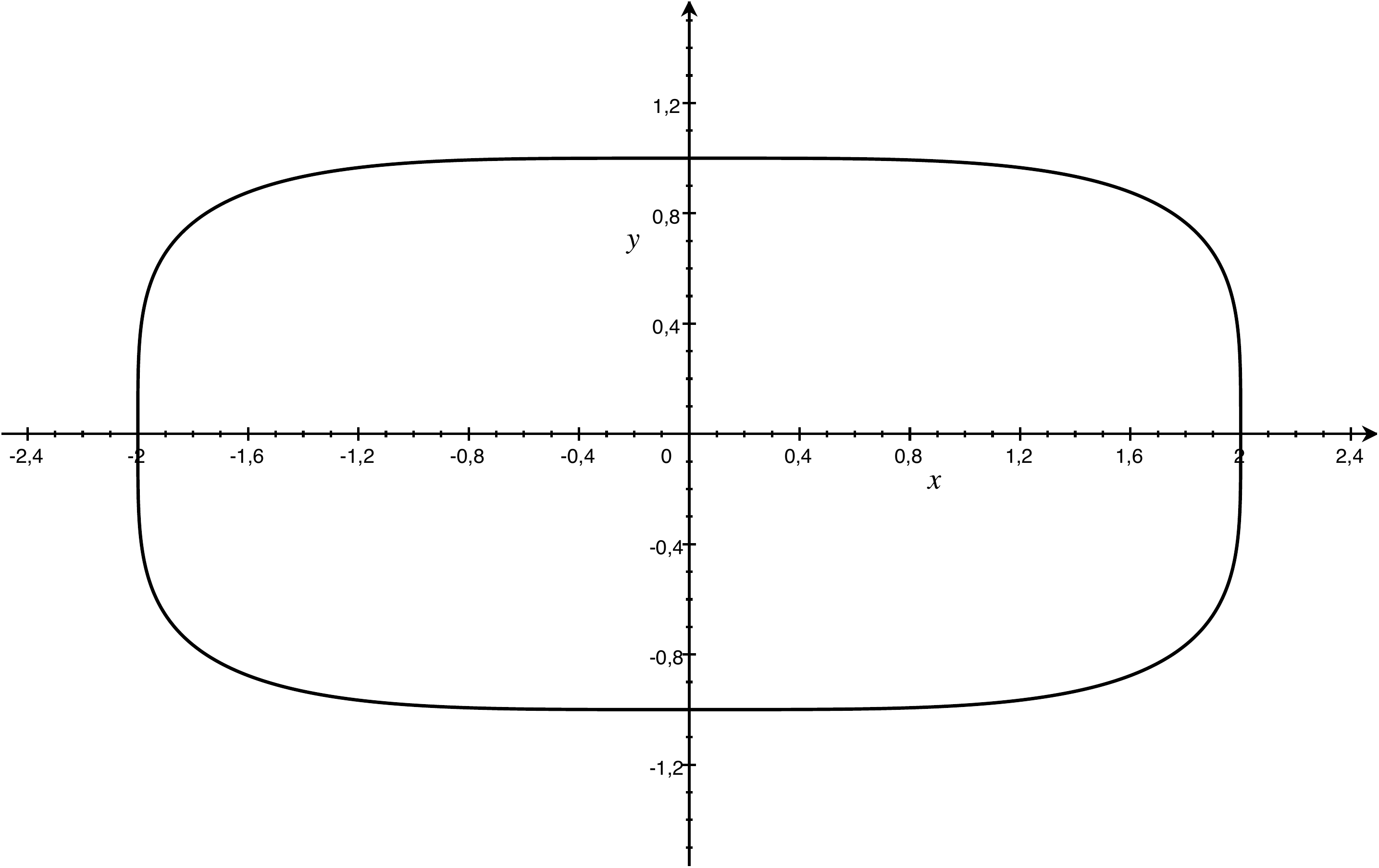} & 
\includegraphics[width=0.3\linewidth]{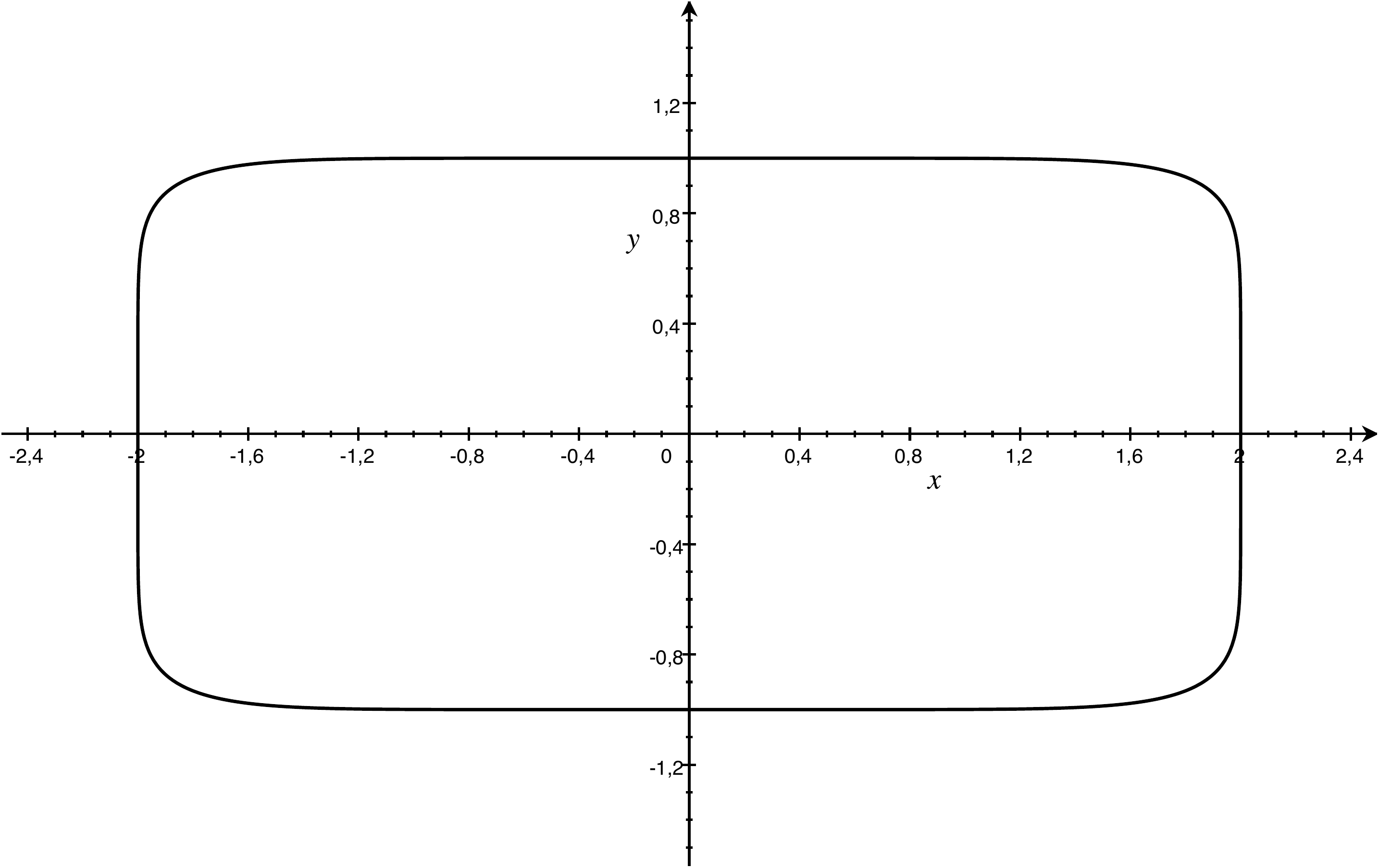} &
\includegraphics[width=0.3\linewidth]{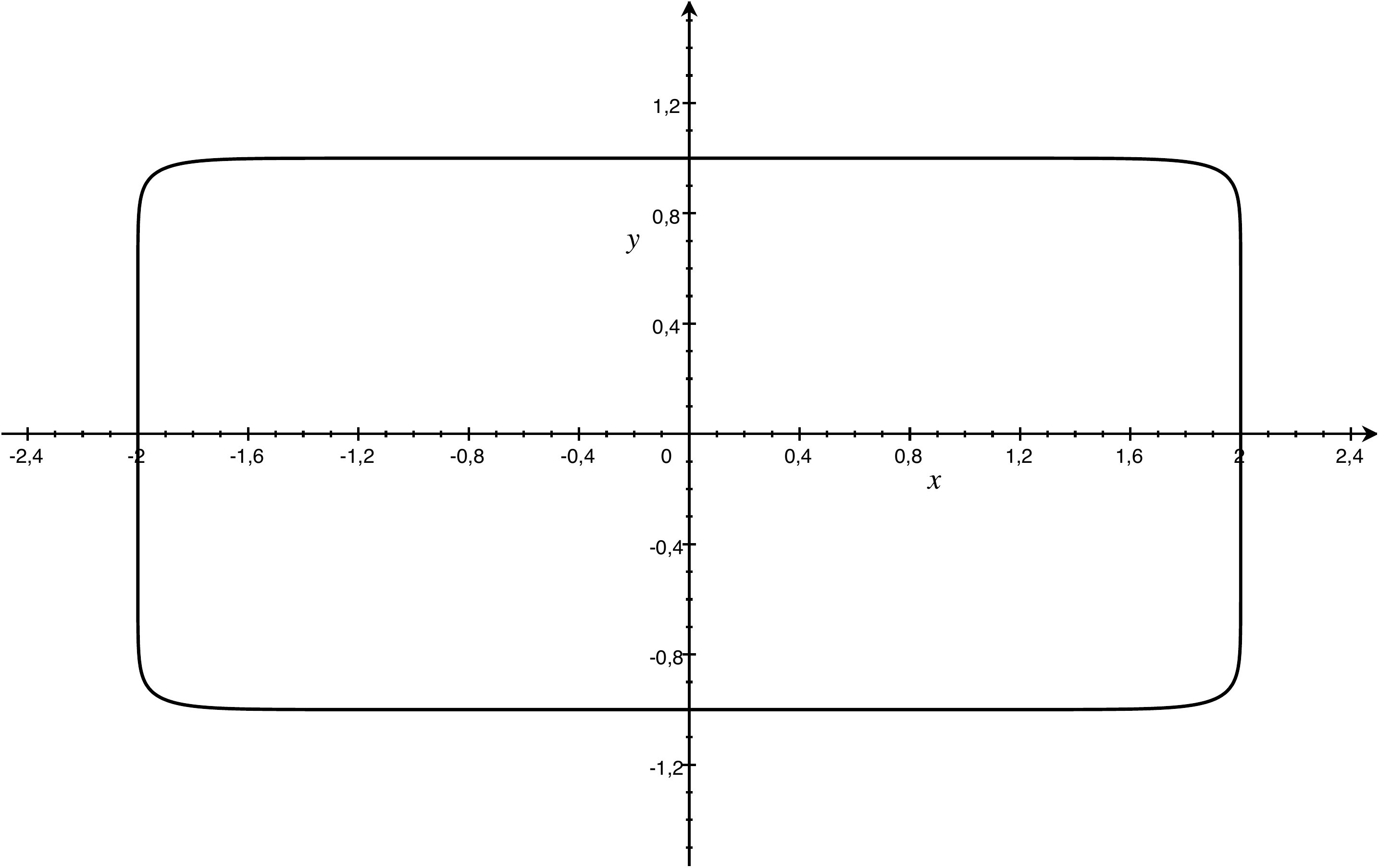}\\
(a) & (b) & (c)
\end{tabular}
\caption{Curve of Lamet with 
$a=2$, $b=1$ and: (a) $m=4$, (b) $m=8$, and (c) $m=16$.}\label{LametCurve}
\end{figure}

The \emph{curve of Lamet} is a curve of degree $m$, with $m$ an even positive integer, whose outline is a rectangle with rounded corners.
Its Cartesian equation is:
$$
\frac{x^m}{a^m}+\frac{y^m}{b} = 1
$$
or equivalently in polynomial form:
$bx^m+a^my^m = a^m b$,
with $a,b \in \mathbb{R}_{>0}$ and it is a bounded connected closed curve with two axes of symmetry (the $x$ and the $y$ axes).
The curve of Lamet is contained in the rectangular region $[-a,a]\times[-b^{1/m}, b^{1/m}]$.
Some examples are provided in Figure \ref{LametCurve} using different values of the parameters $a$, $b$ and $m$.

Another interesting shape (for instance when looking for a mouth, an eye feature or a leaf like decoration) 
is given by the sextic surface of equation
$$
a^4(x^2+z^2)+(y-a)^3y^2=0
$$
with $a \in \mathbb{R}$, called the \emph{zitrus} (or \emph{citrus}) 
\emph{surface} by Herwig Hauser \cite{Imaginary}.
The citrus surface has bounding box $[-\frac{a}{8}, \frac{a}{8}] \times [0,a] \times [-\frac{a}{8}, \frac{a}{8}]$, 
centroid at $(0,\frac{a}{2},0)$ and volume $\frac{1}{140}\pi a^3$. 

\begin{figure}[htb]
\begin{center}
\begin{tabular}{ccc}
\includegraphics[width=0.4\linewidth]{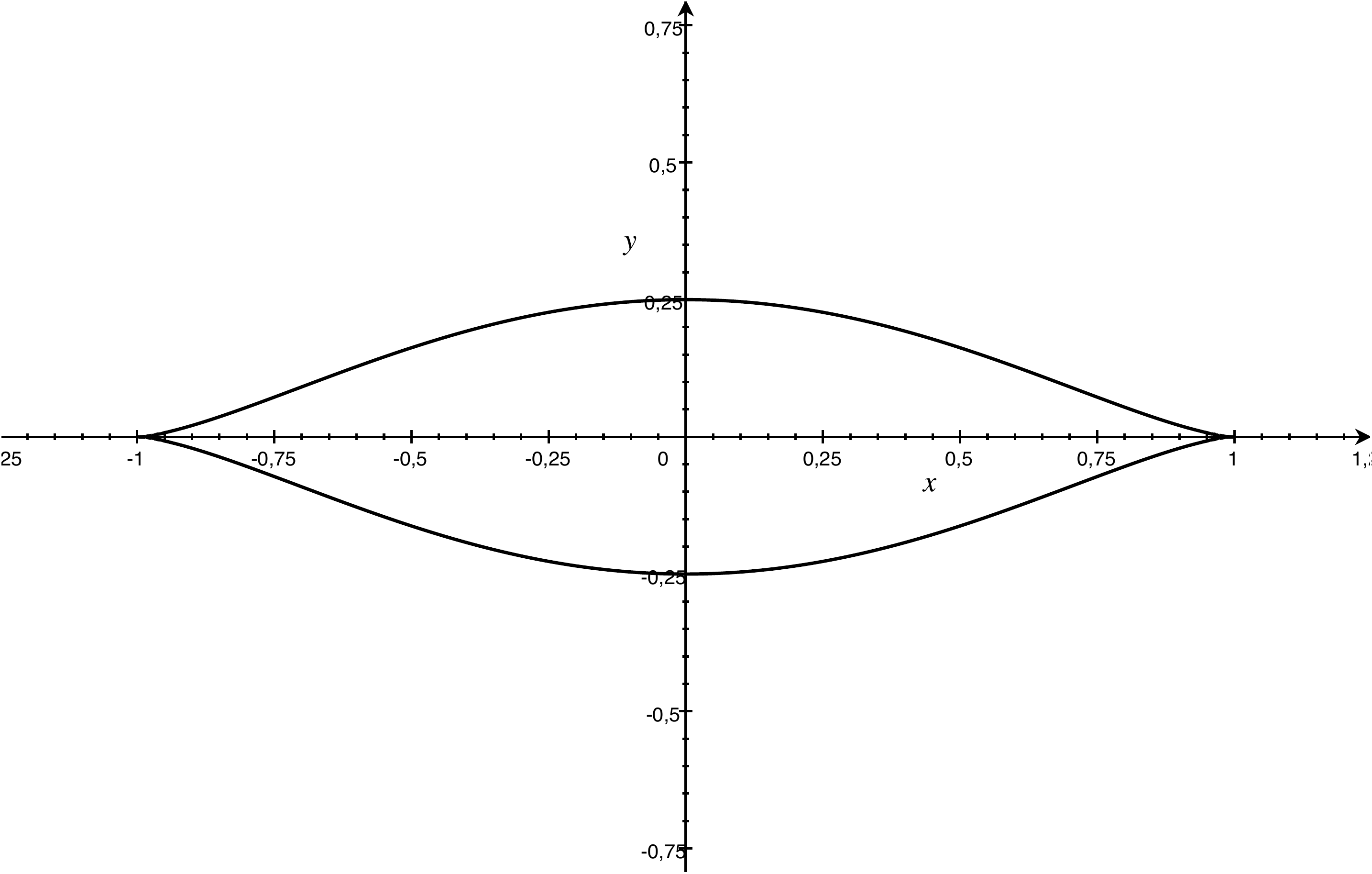} & &
\includegraphics[width=0.4\linewidth]{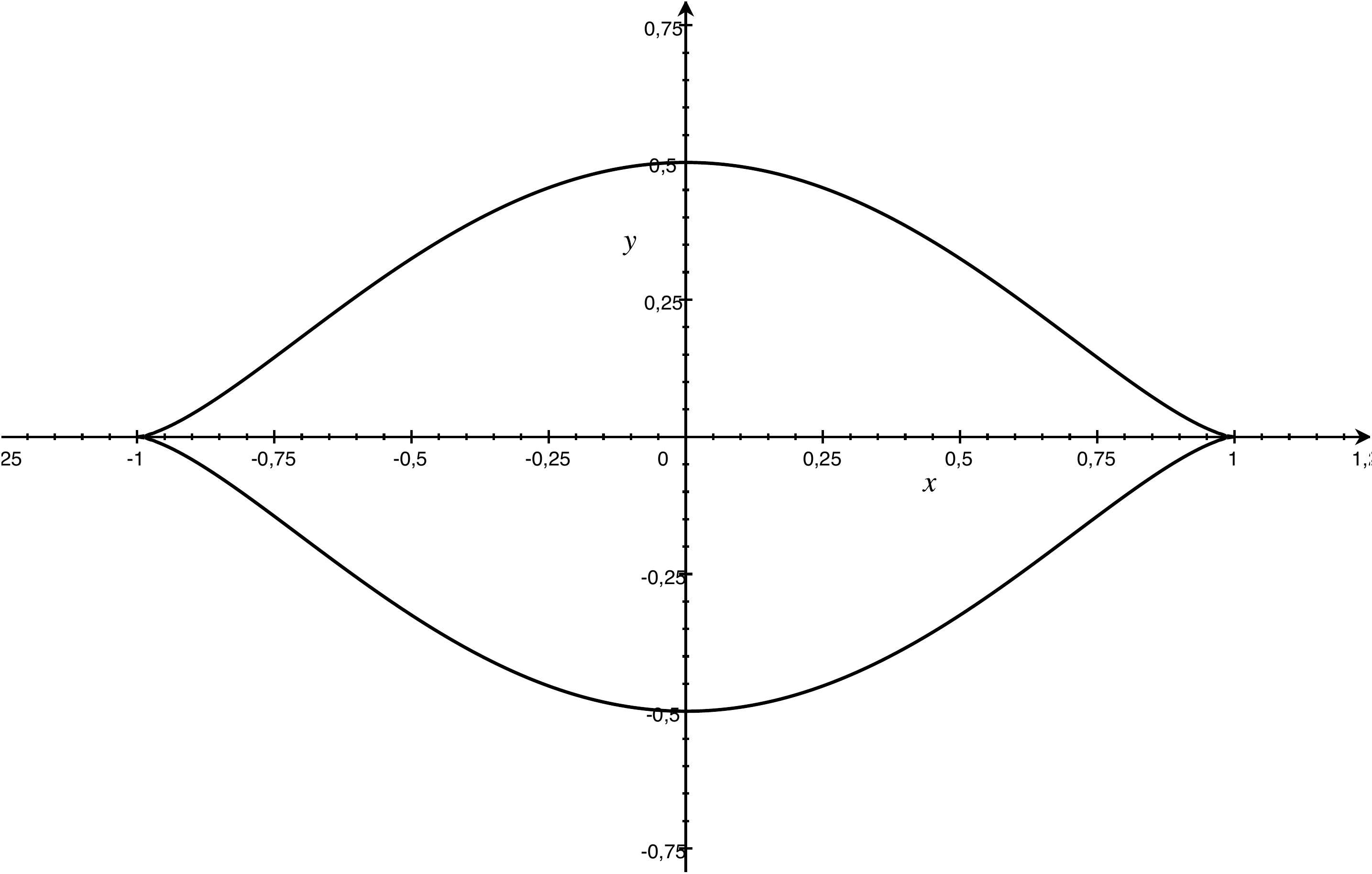} \\
(a) & & (b)
\end{tabular}
\caption{The citrus curve of equation: (a) $f_a(x,y)=0$ with $a=2$, $c=1/4$, (b) $f_{a,c}(x,y)=0$ with $a=2$, $c=1/2$.}\label{citrusCurve}
\end{center}
\end{figure}

We derive the \emph{citrus curve} of equation $f_a(x,y)=0$ as the intersection of a rotation of $\pi/2$ of the citrus surface with the plane $z=0$,
where
$f_a(x,y)$ is the following sextic polynomial
\begin{eqnarray*}
f_a(x,y) = a^4y^2 + \left(x-\frac{a}{2} \right)^3 \left(x+\frac{a}{2} \right)^3
\end{eqnarray*} 
with $a \in \mathbb{R}$ (see Figure \ref{citrusCurve}(a)).
The citrus curve is a symmetric bounded curve with bounding box 
$[-\frac{a}{2}, \frac{a}{2}] \times [-\frac{a}{8}, \frac{a}{8}]$.

To include shapes with a different ratio, we introduce another citrus curve
whose equation is $f_{a,c}(x,y)=0$ (which is simply stretched or shortened along the $y$-axis) where $f_{a,c}(x,y)$ is given by:
\begin{eqnarray*}
f(x,y) = a^4c^2y^2 + \left(x-\frac{a}{2} \right)^3 \left(x+\frac{a}{2} \right)^3
\end{eqnarray*}
with $a,c \in \mathbb{R}$ (see Figure \ref{citrusCurve}(b)).
Note that this is again a symmetric bounded curve with bounding box 
$[-\frac{a}{2}, \frac{a}{2}] \times [-\frac{a}{8c}, \frac{a}{8c}]$.

\begin{figure}[htb]
\begin{center}
\begin{tabular}{ccc}
\includegraphics[width=0.4\linewidth]{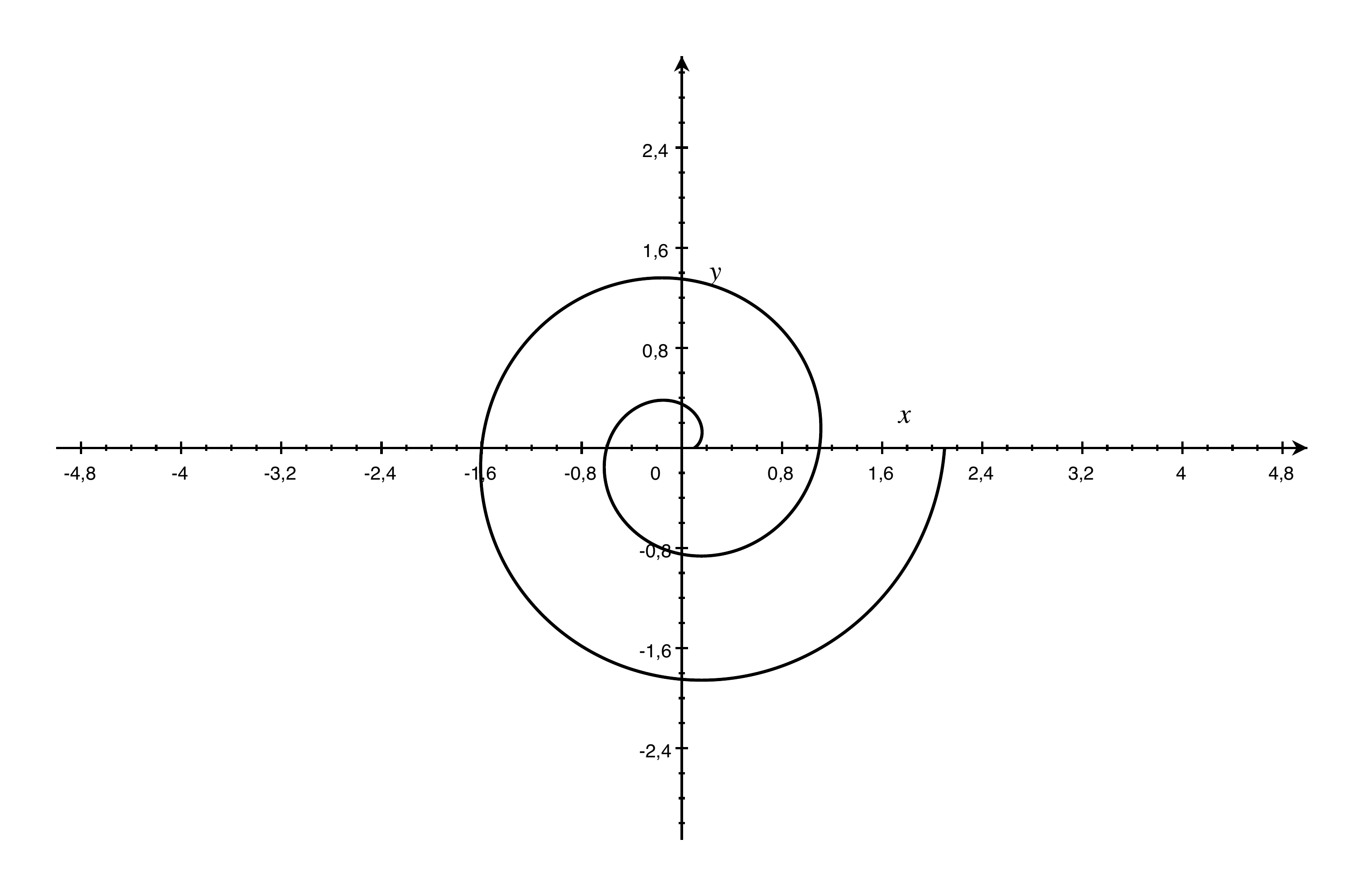} & &
\includegraphics[width=0.4\linewidth]{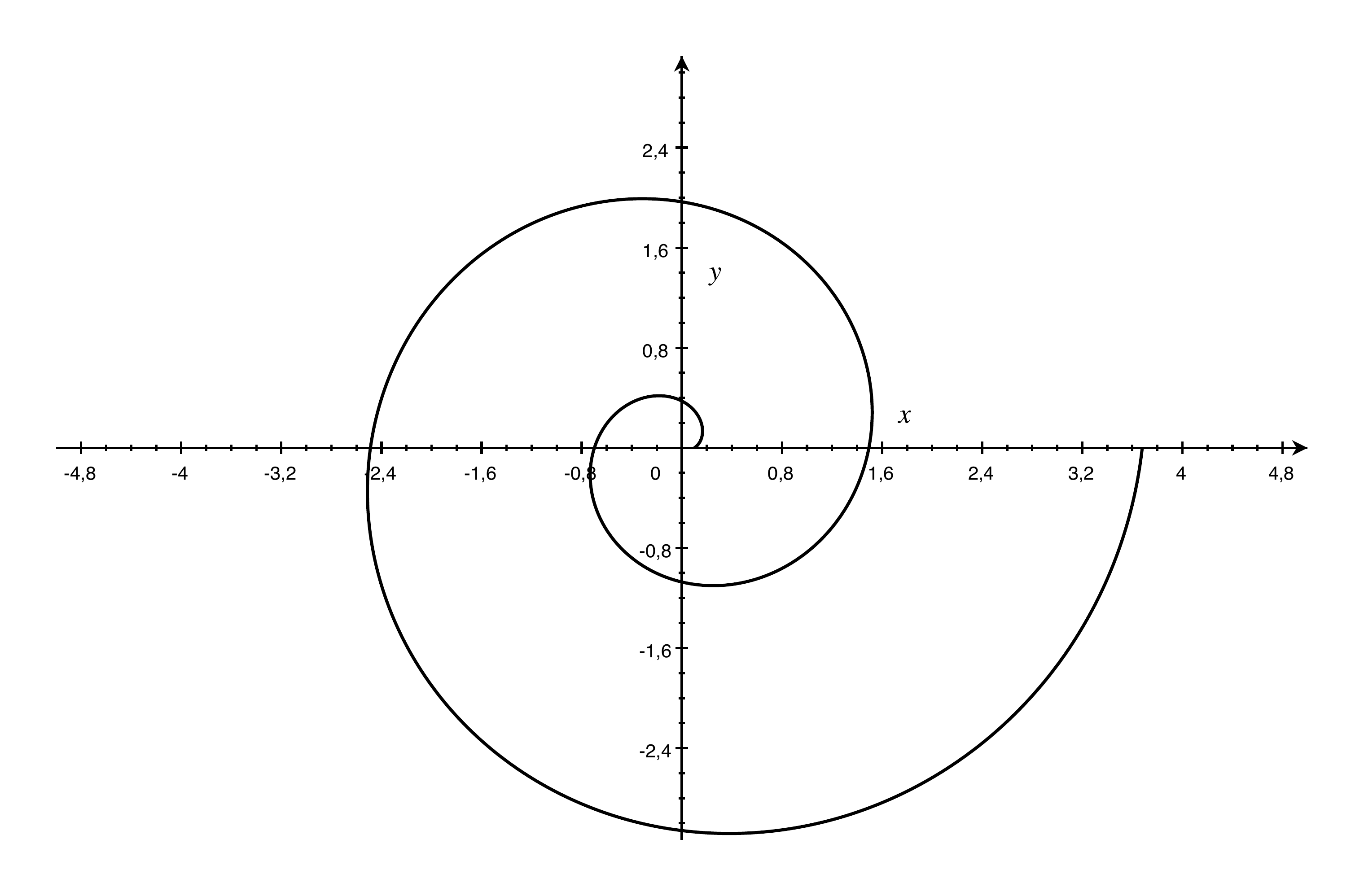}\\
(a) & & (b)
\end{tabular}
\caption{Archimedean spirals with parameters: (a) $a=1/10$, $b=1/2 \pi$ and (b) $a=1/10$, $b=1/2 \pi$, 
$c=1/100$.}\label{ArchSpiral}
\end{center}
\end{figure}

The \emph{Archimedean spiral} (or \emph{arithmetic spiral})
can be found in human artefact decorations 
as well as in nature. Its polar equation is:
$$
\rho = a+b \theta
$$ 
with $a,b \in \mathbb{R}$.
The Archimedean spiral is an unbounded connected curve with a single singular point (the \emph{cessation point} $(a,0)$).
Two consecutive turnings of the spiral have a constant separation
distance equal to $2\pi b$, hence the name arithmetic spiral. Another peculiar aspect is that, though its unboundedness 
nature, the $k$th turning of the spiral is contained in a region bounded by two concentric circles of radii $a+2(k-1)\pi b$ and
$a+2k \pi b$. In particular, the first turning is contained in the circular annulus of radii $a$ and $a+2\pi b$.
An example is provided in Figure \ref{ArchSpiral}(a). 

We extend the Archimedean spiral by weakening its constant pitch property, introducing an extra parameter $c$; its polar equation becomes:
$$
\rho = a+b \theta+c\theta^2
$$ 
with $a, b, c \in \mathbb{R}$.
Analogously to the Archimedean spiral, this generalized family is still an unbounded connected curve with a single 
singular point (the \emph{cessation point} $(a,0)$). An example is provided in Figure \ref{ArchSpiral}(b). 

In order to show the behaviour of the Main Algorithm \ref{mainAlg}, the generalized version of
the Archimedean spirals has been employed in Section \ref{Sec:overview}. Some computational details have
been provided in the case of the set $\mathbb Y_1$, represented in Figure \ref{fig:spiral}(b).
Exploiting the Cartesian coordinates of the cessation point $(0.3915, 0.0048)$ of the set $\mathbb Y_1$, 
and the fact that the first turning of the spiral is contained in the circumference of radius $1.1542$ centered at the origin, 
we considered the region of the parameter space $T=[0.35,0.45] \times [0.1,0.15] \times [0.0032,0.0048]$ and the discretization step $(0.01, 0.01, 0.004)$. 
Figure \ref{fig:spiral} presents a set of feature curves recognized with the extended Archimedean spiral. With reference to the example shown in Figure \ref{fig:spiral}(b), the maximum of the accumulator function is $44$ (the second maximum has value $29$) and corresponds to the cell with center $(\frac{7}{20},\frac{1}{10},\frac{2}{625})$.

\begin{figure}[htb]
\begin{center}
\begin{tabular}{ccc}
\includegraphics[width=0.3\linewidth]{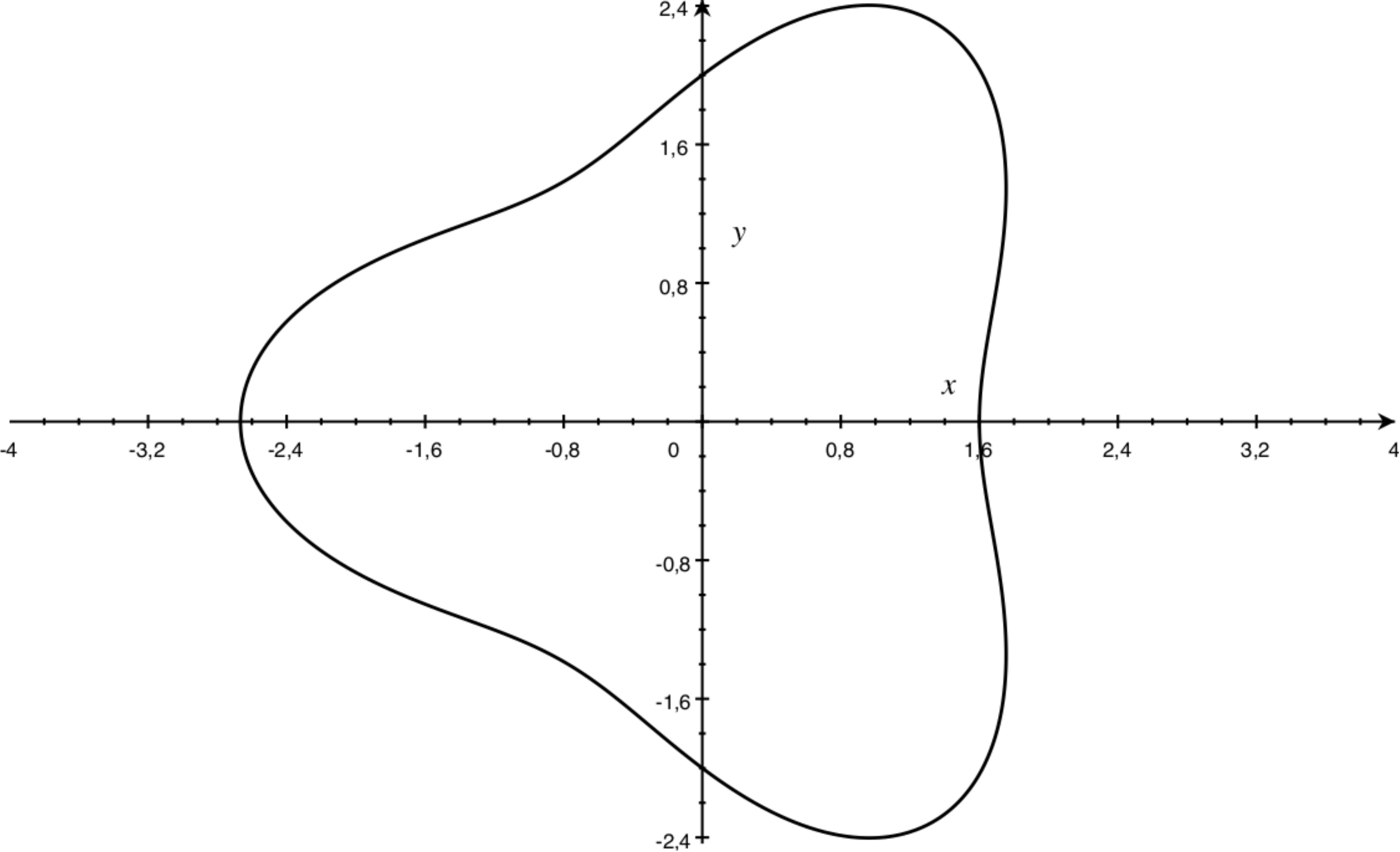} & 
\includegraphics[width=0.3\linewidth]{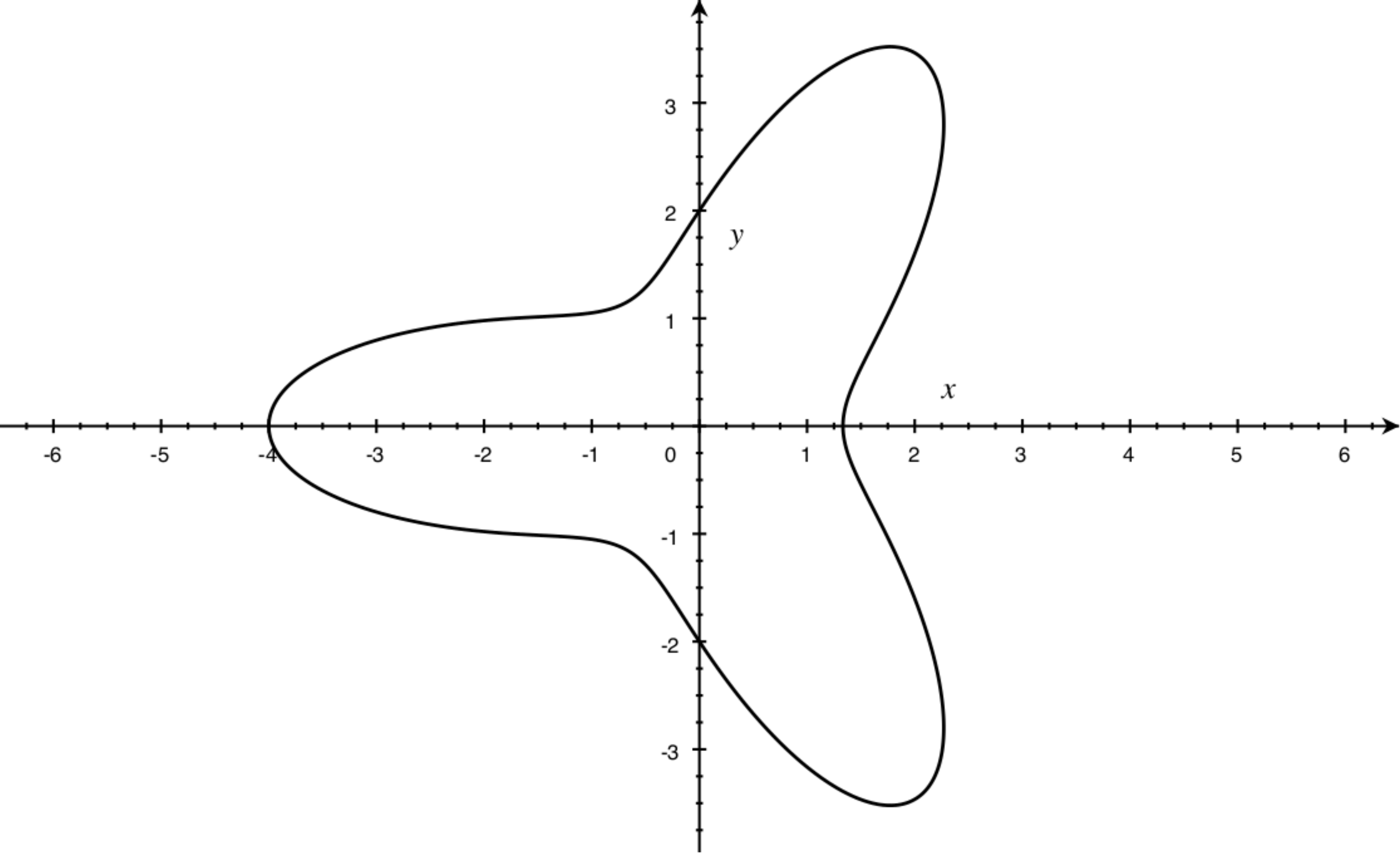} & 
\includegraphics[width=0.3\linewidth]{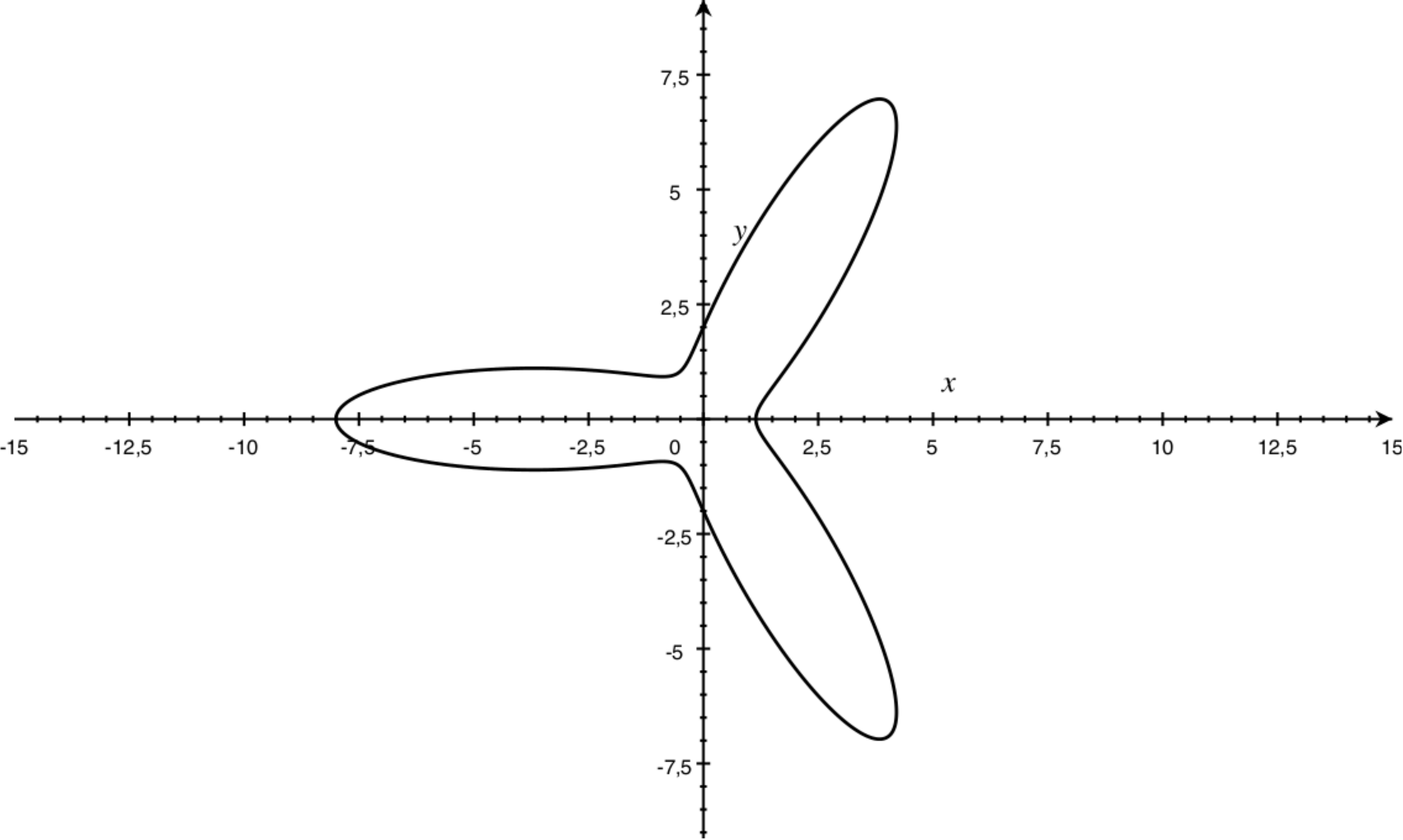} \\
(a) & (b) & (c)\\
\\
\includegraphics[width=0.3\linewidth]{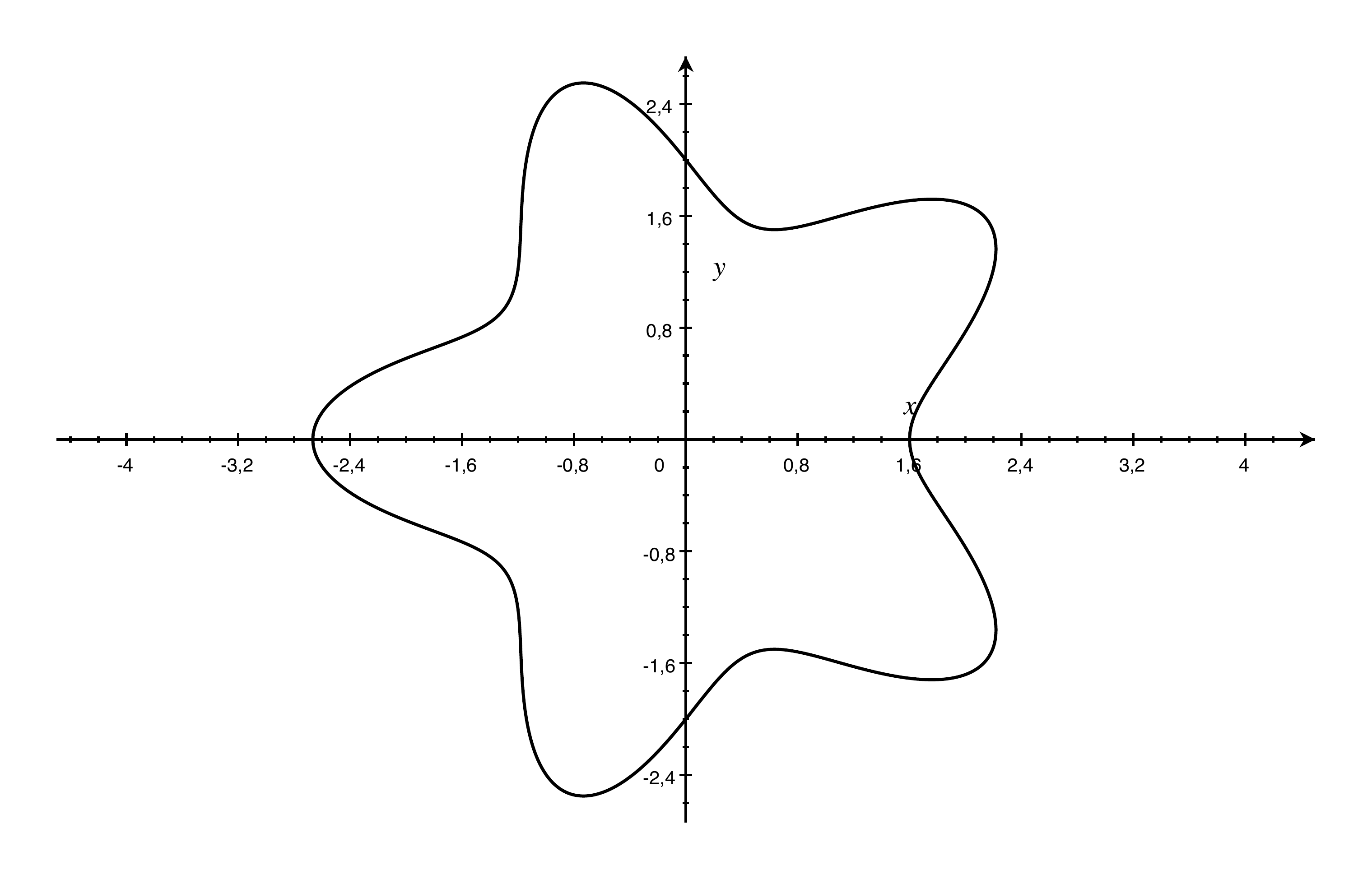} & 
\includegraphics[width=0.3\linewidth]{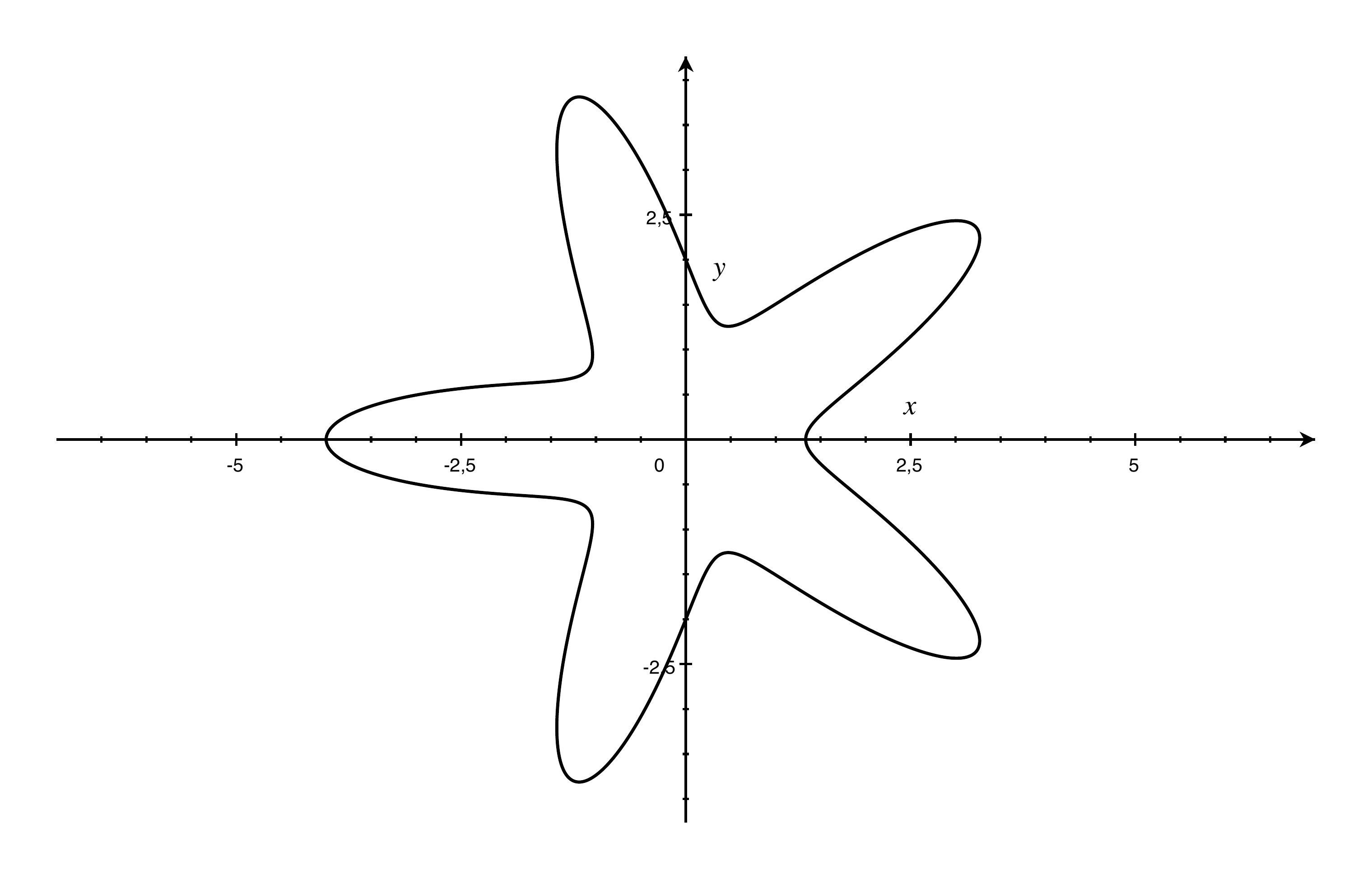} & 
\includegraphics[width=0.3\linewidth]{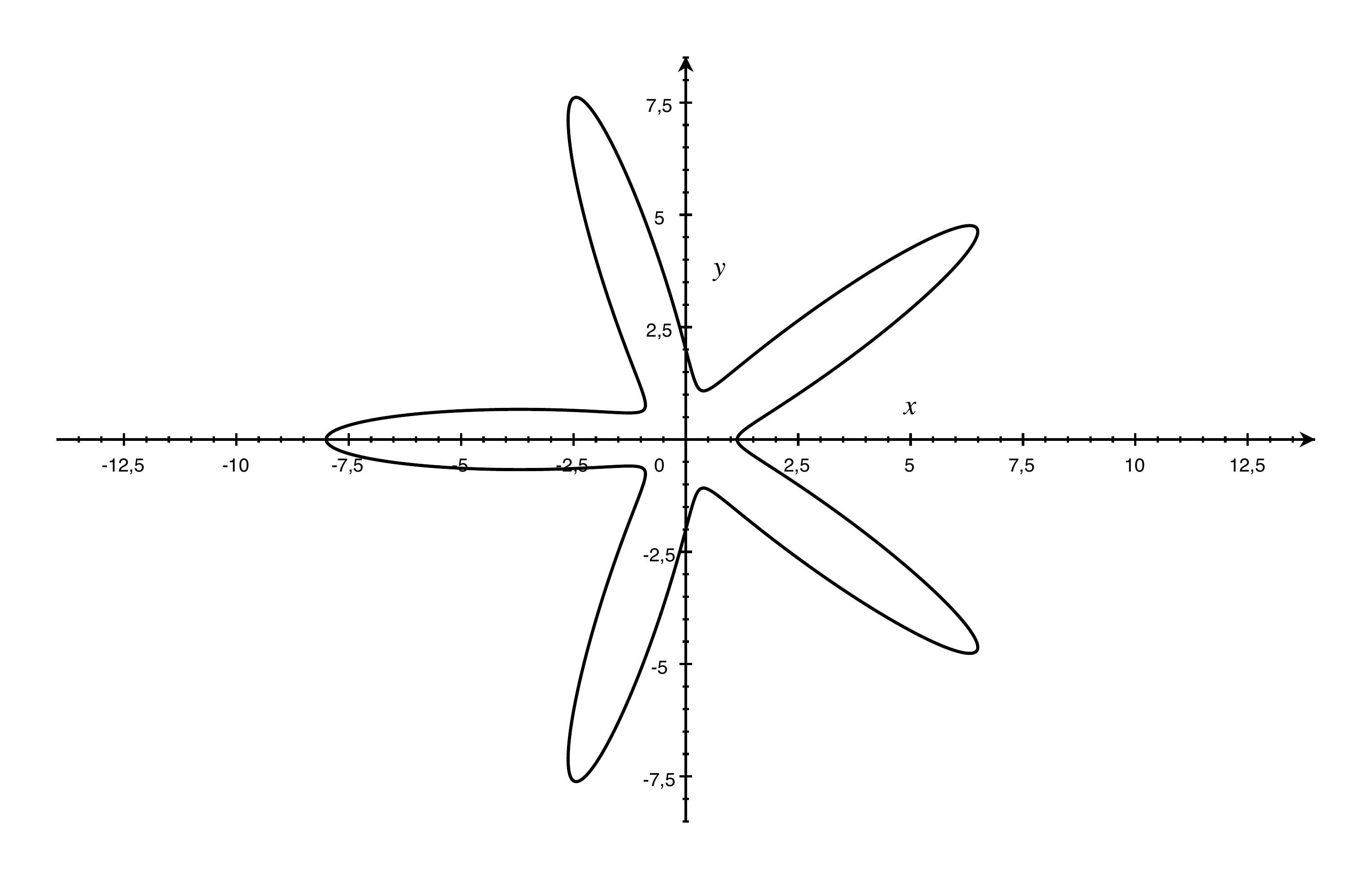} \\
(d) & (e) & (f)
\end{tabular}
\caption{Curves with $m$-convexities with $a=2$ and (a) $b=1/4$, $m=3$,
(b) $b=1/2$, $m=3$, (c) $b=3/4$, $m=3$, (d) $b=1/4$, $m=5$,
(e) $b=1/2$, $m=5$, (f) $b=3/4$, $m=5$.}\label{convexCurve}
\end{center}
\end{figure}

Figure \ref{convexCurve} shows another family of curves.
The \emph{curve with $m$-convexities} defined by the polar equation
$$
\rho = \frac{a}{1+b \cos(m \theta)}
$$ 
with $a,b \in \mathbb{R}_{>0}$, $b<1$, and $m \in \mathbb{N_+}$, $m\ge 2$.
The curve with $m$-convexities is a bounded connected closed curve with $m$ axes 
of symmetry (the straight lines of equation $x \sin \frac{\pi}{m} k - y \cos \frac{\pi}{m} k=0$, 
with $k=0,\ldots, m-1$). This curve is contained in a region bounded by two concentric circles of radii $\frac{a}{1+b}$ and $\frac{a}{1-b}$. 
The shape of the curve with $m$-convexities strongly depends on the values of its parameters.
In particular, the parameter $a$ plays the role of a scale factor, while the values of $b$ tune 
the convexities' sharpness. 
Some examples of curves with $m$-convexities are provided in Figure \ref{convexCurve} where in the first row parameter $m=3$, and in the second row parameter $m=5$.

\begin{figure}[htb]
\begin{center}
\begin{tabular}{ccc}
\includegraphics[width=0.4\linewidth]{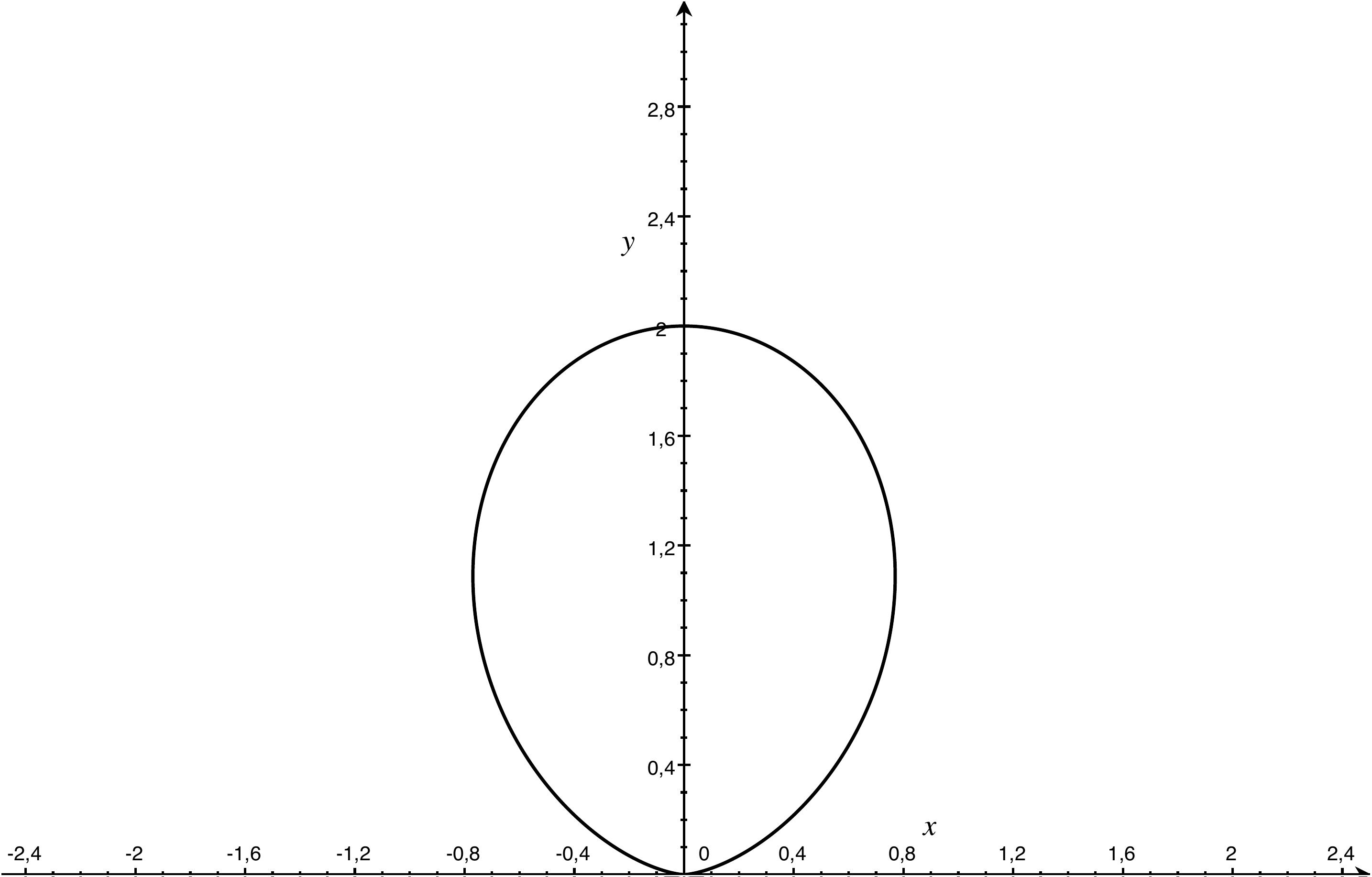} & &
\includegraphics[width=0.4\linewidth]{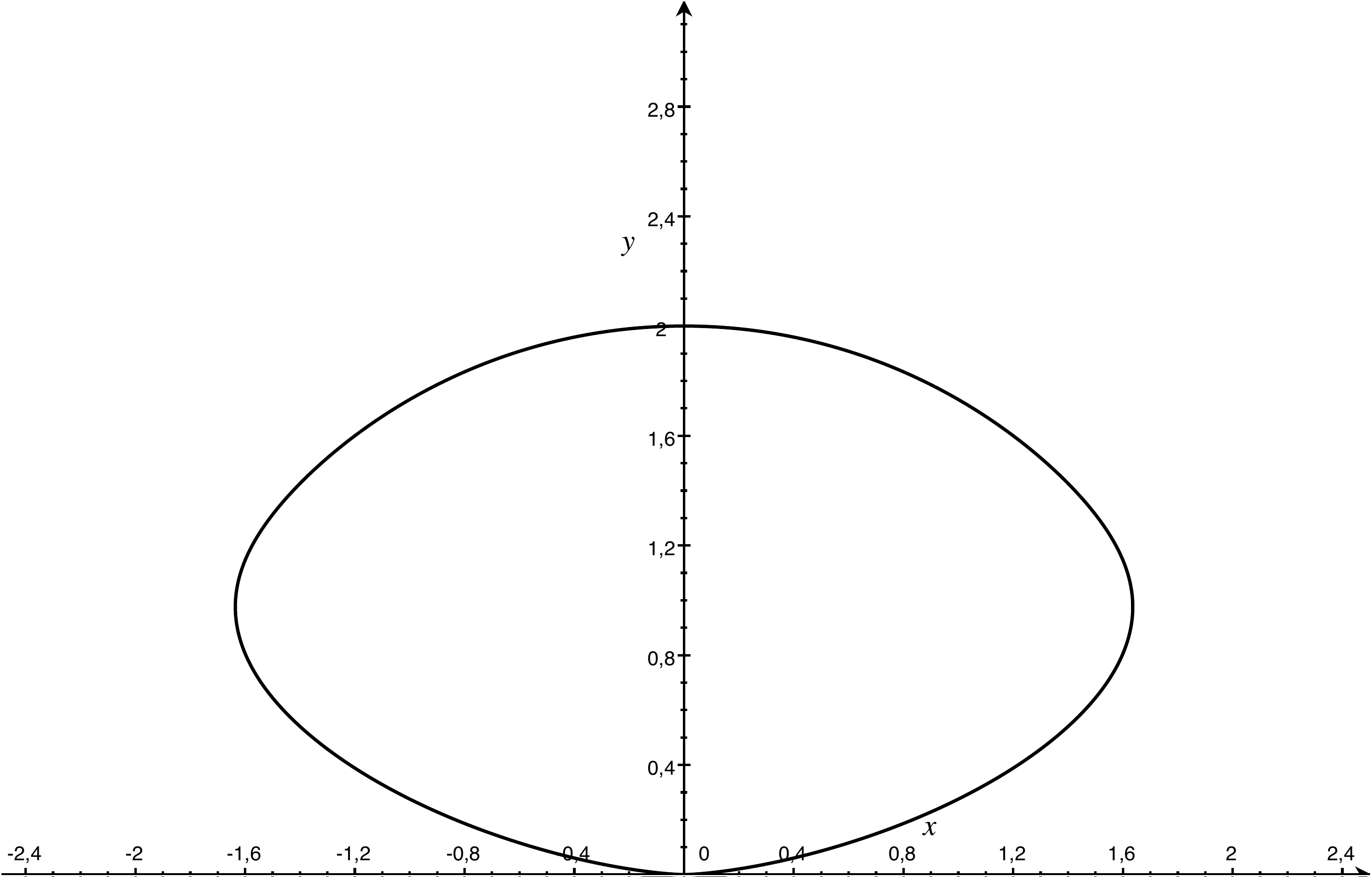}\\
(a) & & (b)\\
\\
\includegraphics[width=0.4\linewidth]{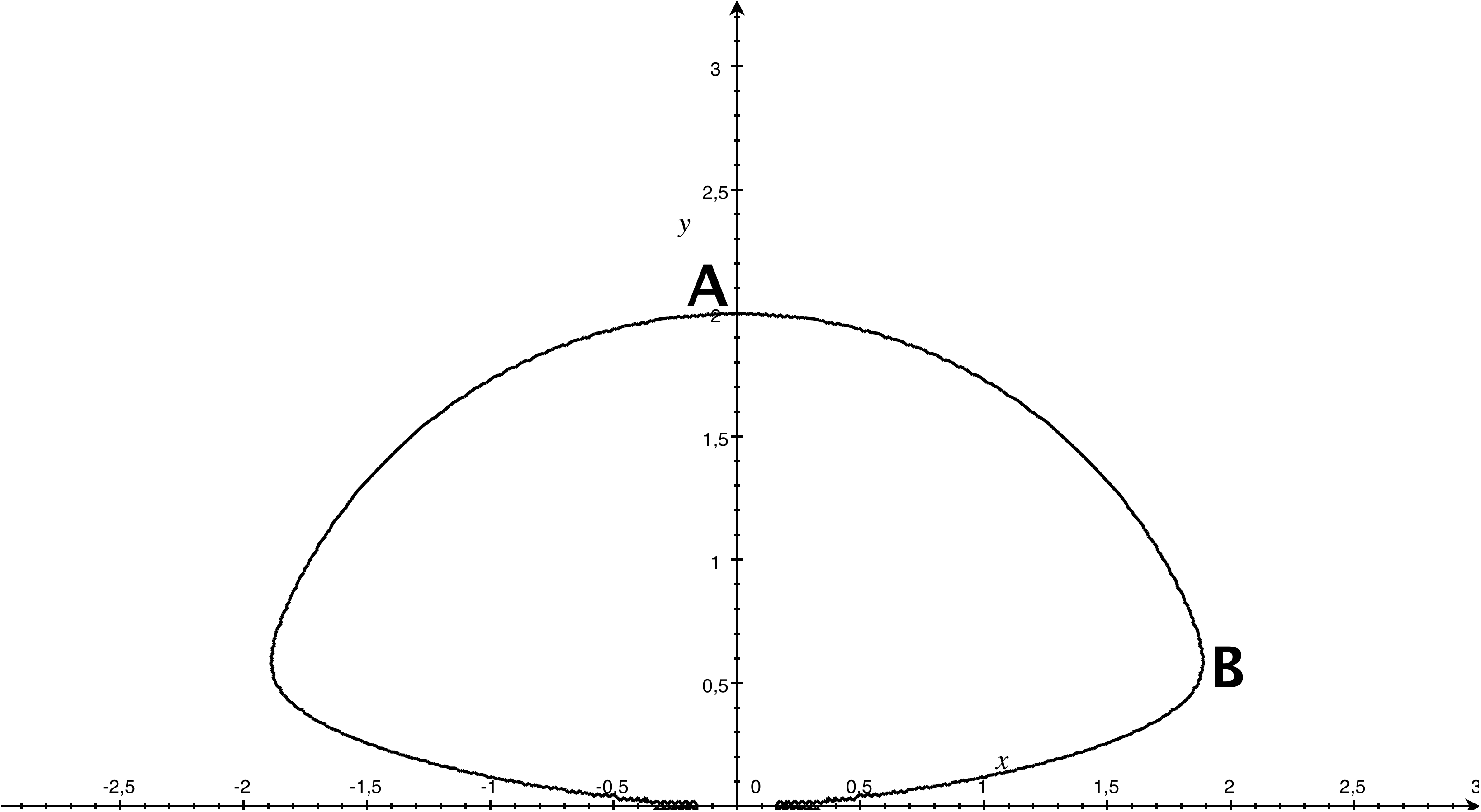} & &
\includegraphics[width=0.4\linewidth]{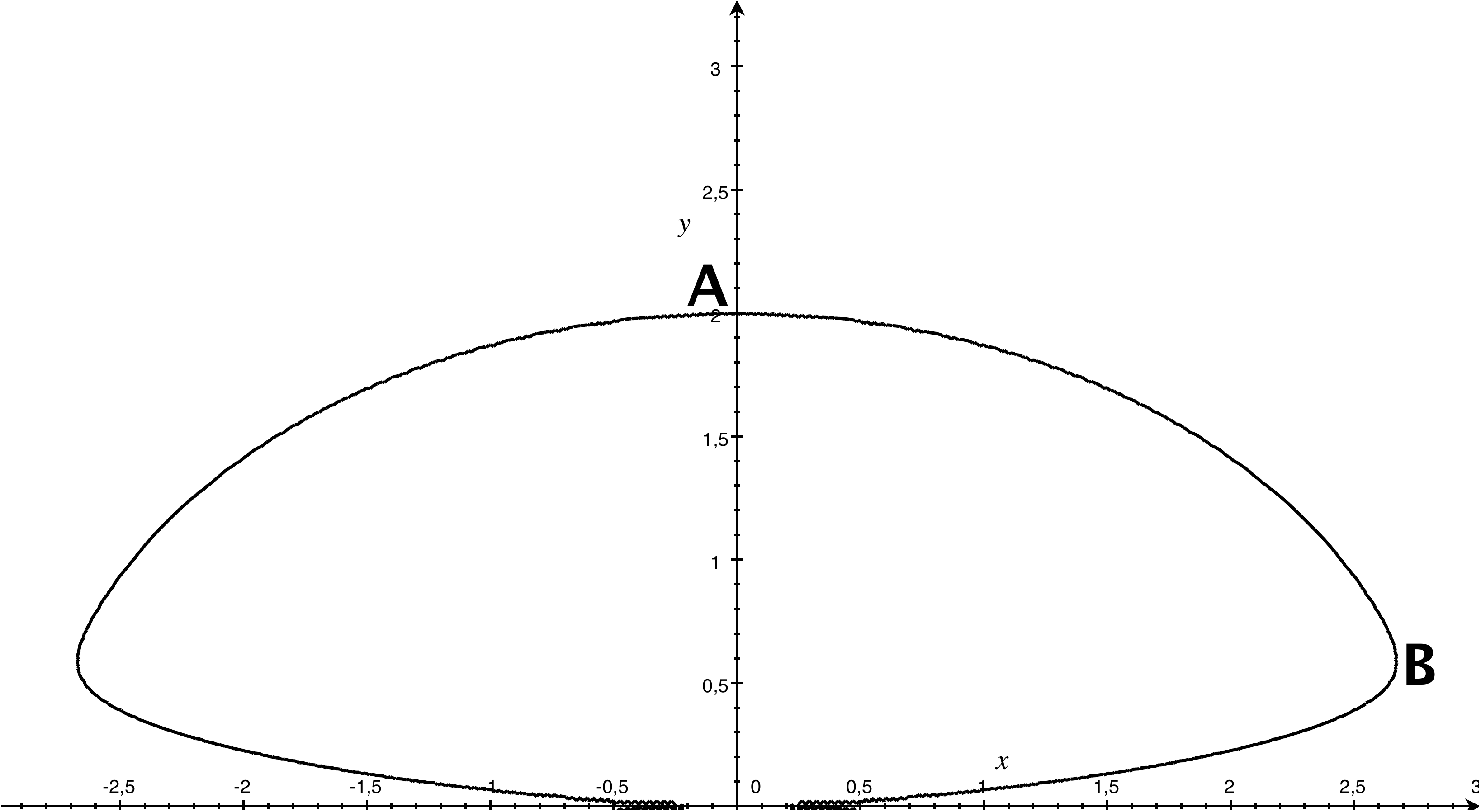}\\
(c) & & (d)
\end{tabular}
\caption{Geometric petal curves with parameters $a=2$, $b=-2$ and
(a) $n=1$, (b) $n=10$.
Geometric petal curves defined by $g_{a,c}(x,y)=0$ with parameters
$a=4$, $n=50$ and (c) $c=1$, (d) $c=1/2$.}
\label{geomPetal}
\end{center}
\end{figure}

The so-called \emph{geometric petal} curve resembles an eye contour line,  for particular values of the parameters. 
Its polar equation is:
$$
\rho = a+b \cos^{2n} \theta
$$ 
with $n \in \mathbb{N}_+$ and $a,b \in \mathbb{R}$.
The geometrical petal is a bounded symmetric curve with a singularity at the origin.
Some examples are provided in Figure \ref{geomPetal}(a)-(b) where the values of the parameters are set as follows: $a=2$, $b=-2$ and $n=1, 10$.

For our purposes, we can restrict to the case $b=-a$.
In this case, we observe that the curve is completely contained inside the circle
of radius $\sqrt{2}a$.
We pass to the Cartesian equation using the standard substitutions
$\rho=\sqrt{x^2+y^2}$ and $\cos \theta = x/\sqrt{x^2+y^2}$; further, in order to lower
the parameters degree, we replace $a$ by $\sqrt{a}$. The Cartesian equation of the 
geometric petal is $g_a(x,y)=0$ where 
$$
g_a(x,y) = (x^2+y^2)^{2n+1}-a[(x^2+y^2)^n-x^{2n}]^2
$$
From an analytic intersection of the curve with the Cartesian axes, we compute the bounding box of the curve as: $\left[ -\frac{2n}{2n+1}\sqrt{a}\sqrt[2n]{\frac{1}{2n+1}},\frac{2n}{2n+1}\sqrt{a}\sqrt[2n]{\frac{1}{2n+1}}\right] \times \left[-\sqrt{a},\sqrt{a}\right]$.

Most of times, like in the extraction/localization of eyes contours, we need a shape which is more stretched along the $x$-axis (see Figure \ref{geomPetal}(c)-(d)). We stretch the geometric petal by scaling the $x$-variable by a factor of $\sqrt{c}$, where $c \in \mathbb{R}_{>0}$. The new Cartesian equation of the curve is $g_{a,c}(x,y)=0$ 
where
\begin{eqnarray}\label{geomPetalFormula}
g_{a,c}(x,y) = (cx^2+y^2)^{2n+1}-a[(cx^2+y^2)^n-c^nx^{2n}]^2.
\end{eqnarray}
with bounding box $\left[ -\frac{2n}{2n+1}\sqrt{\frac{a}{c}}\sqrt[2n]{\frac{1}{2n+1}},\frac{2n}{2n+1}\sqrt{\frac{a}{c}}\sqrt[2n]{\frac{1}{2n+1}}\right] \times \left[-\sqrt{a},\sqrt{a}\right]$.

To give an idea of the outcome of Algorithm \ref{algHough}, the stretched version of the geometric petal curve 
(equation \ref{geomPetalFormula}) has been employed in Section \ref{Sec:overview}.
In that example, we fixed $n=50$ (indeed the shape of the geometric petal curve changes with this parameter, 
see again Figure~\ref{geomPetal}) and, exploiting the bounding box of the set $\mathbb Z_8$ (see Figure \ref{projection}(c)),
we considered a region $\mathcal T =  [121,122] \times [0.43, 0.45]$ of the parameter space. 

In the following, we show how to reduce the number of parameters of this curve. We observed that the value of the exponent parameter $n$ is related to the bounding box of the curve; indeed it has to satisfy
the following condition:
$$
\frac{2 n}{2n+1} \left( 1- \sqrt[n]{\frac{1}{2n+1}}\right)^{1/2} = \frac{y_B}{y_A}
$$ 
where $y_A$ and $y_B$ are the $y$-coordinates values of the points 
$A$ and $B$ (see Figure \ref{geomPetal} (c)-(d)). 
By using the previous relation, it is possible to estimate the value of $n$, thus working with a 
curve which depends on the two parameters $a$ and $c$. 

%
%

\subsection{Compound curves}
\label{sec:compound}
To grasp compound shapes we use more families of curves simultaneously.
We represent compound curves as a combination of different families of curves and define their equation simply as the product of (two or more) curves' equations.
In the following we show some examples of possible compositions.

\begin{figure}[htb]
\begin{center}
\begin{tabular}{ccc}
\includegraphics[width=0.4\linewidth]{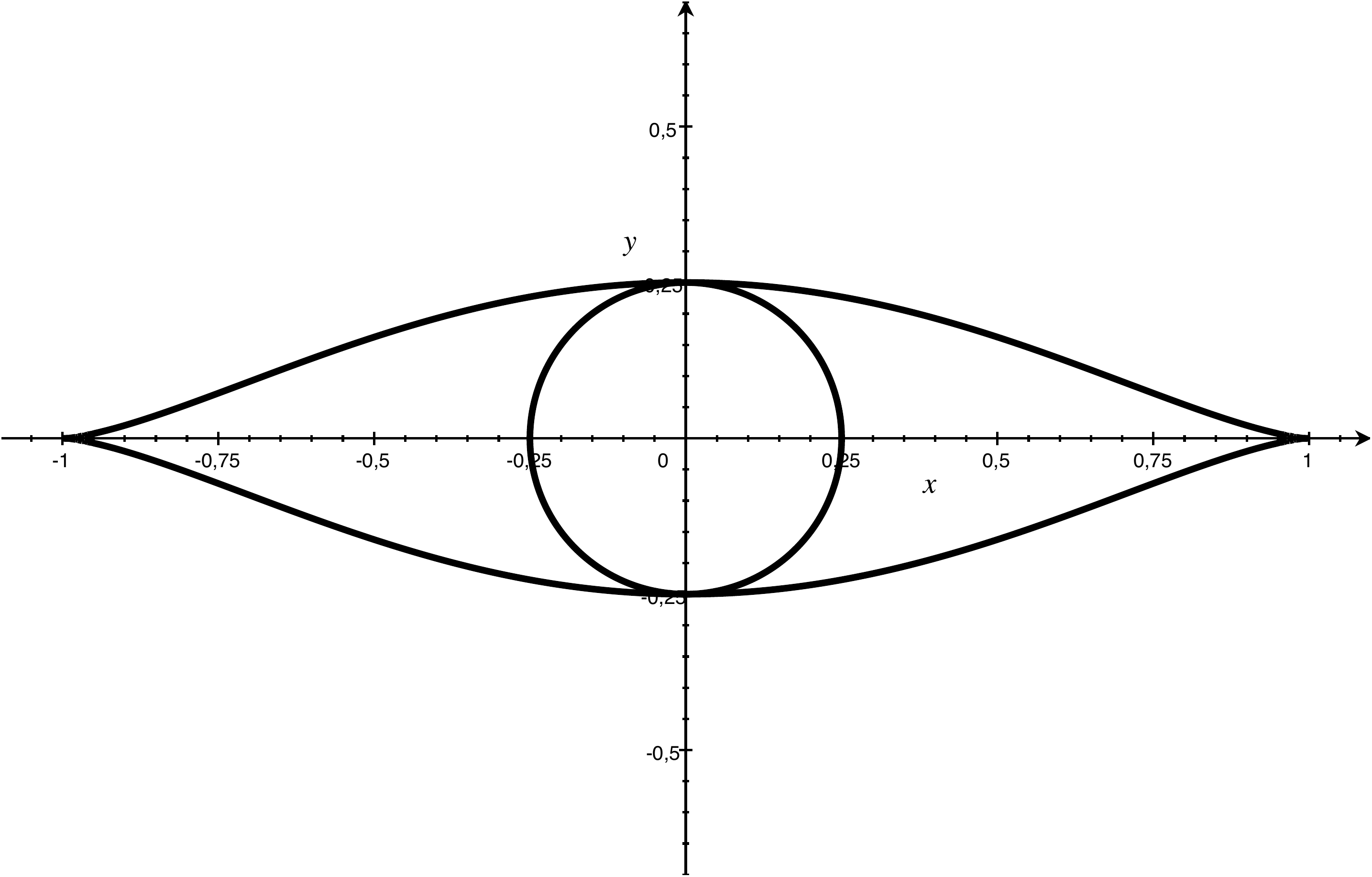} & &
\includegraphics[width=0.4\linewidth]{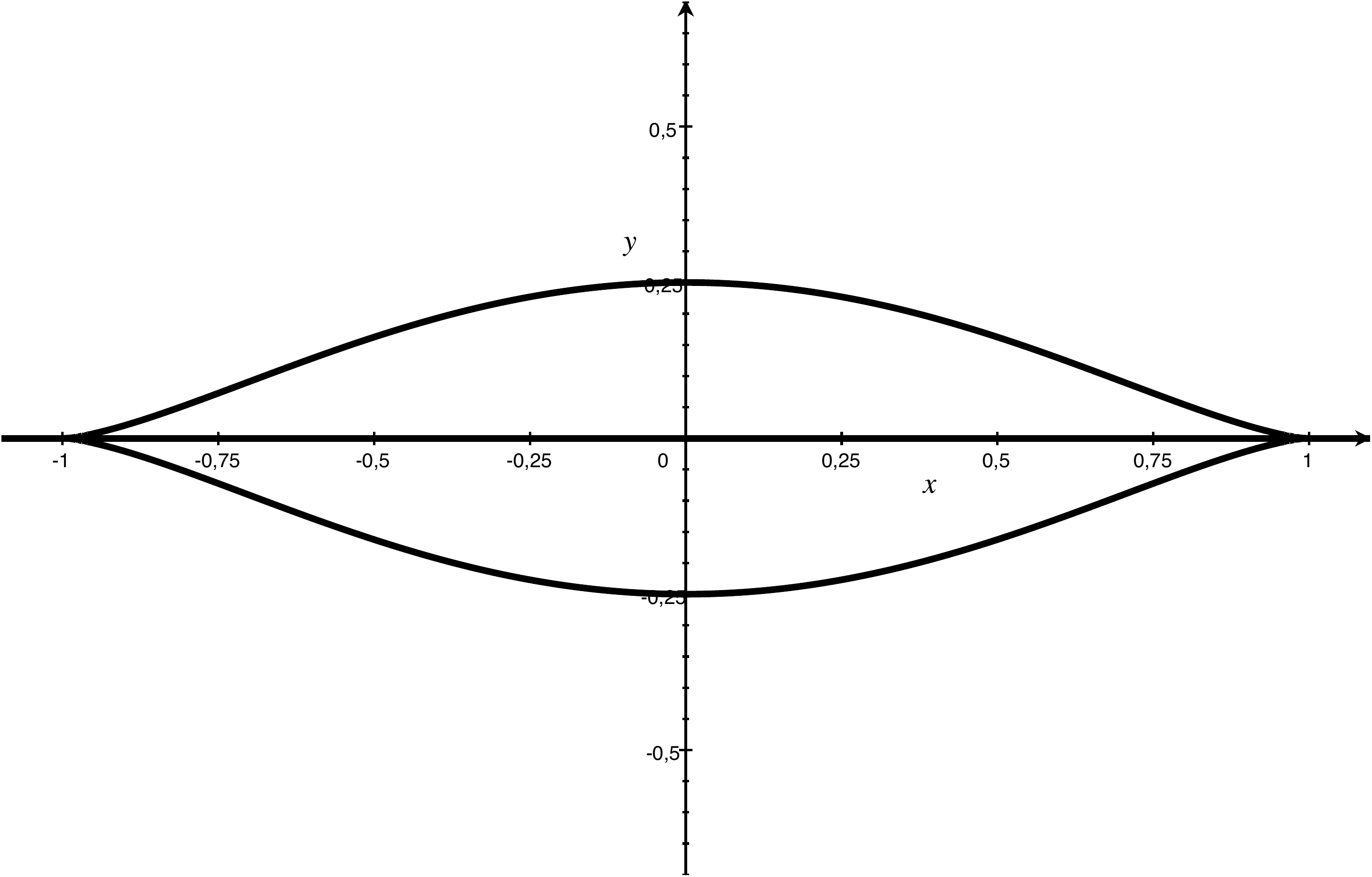}\\
(a) & & (b)\\
\\
\includegraphics[width=0.4\linewidth]{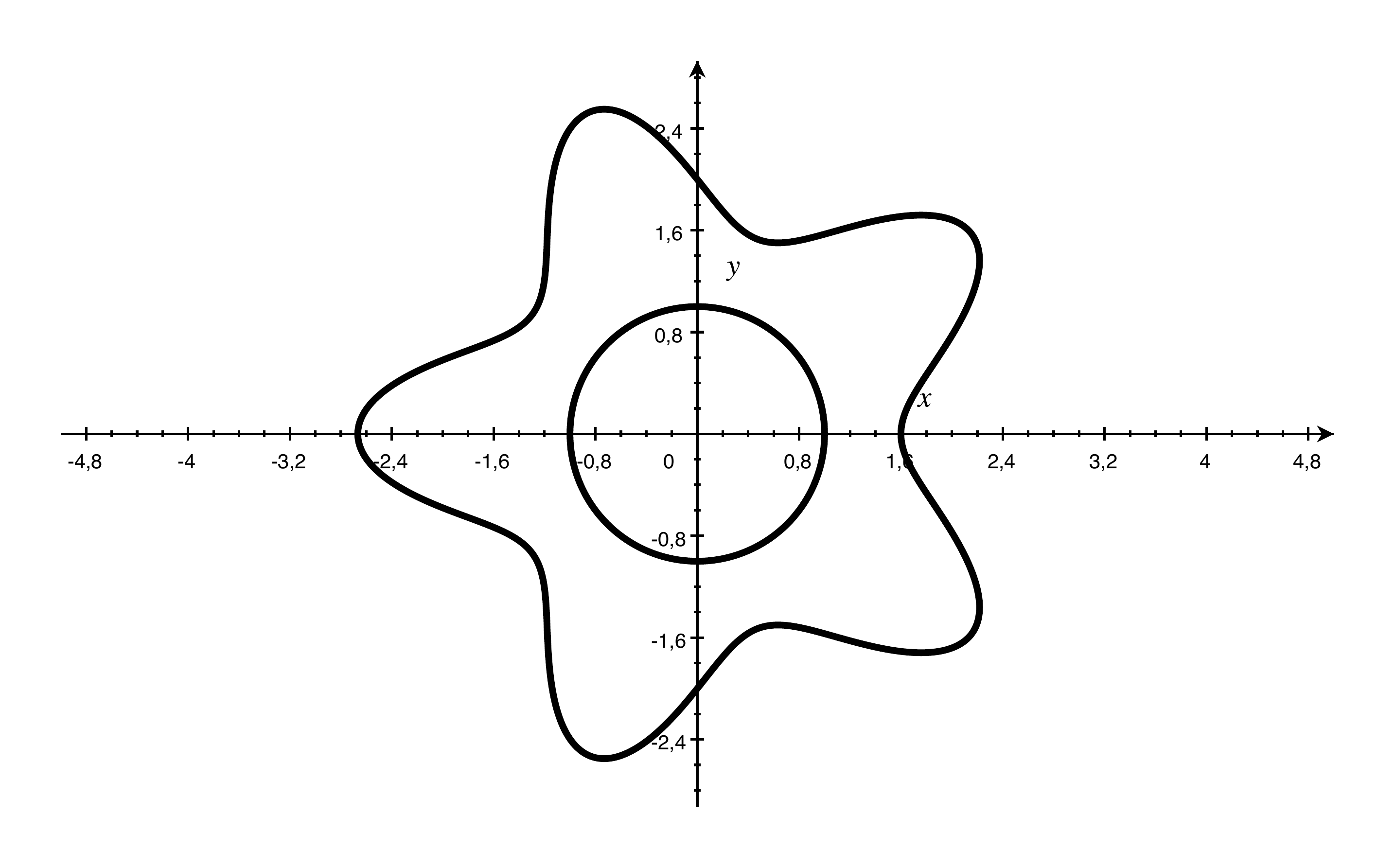} & &
\includegraphics[width=0.4\linewidth]{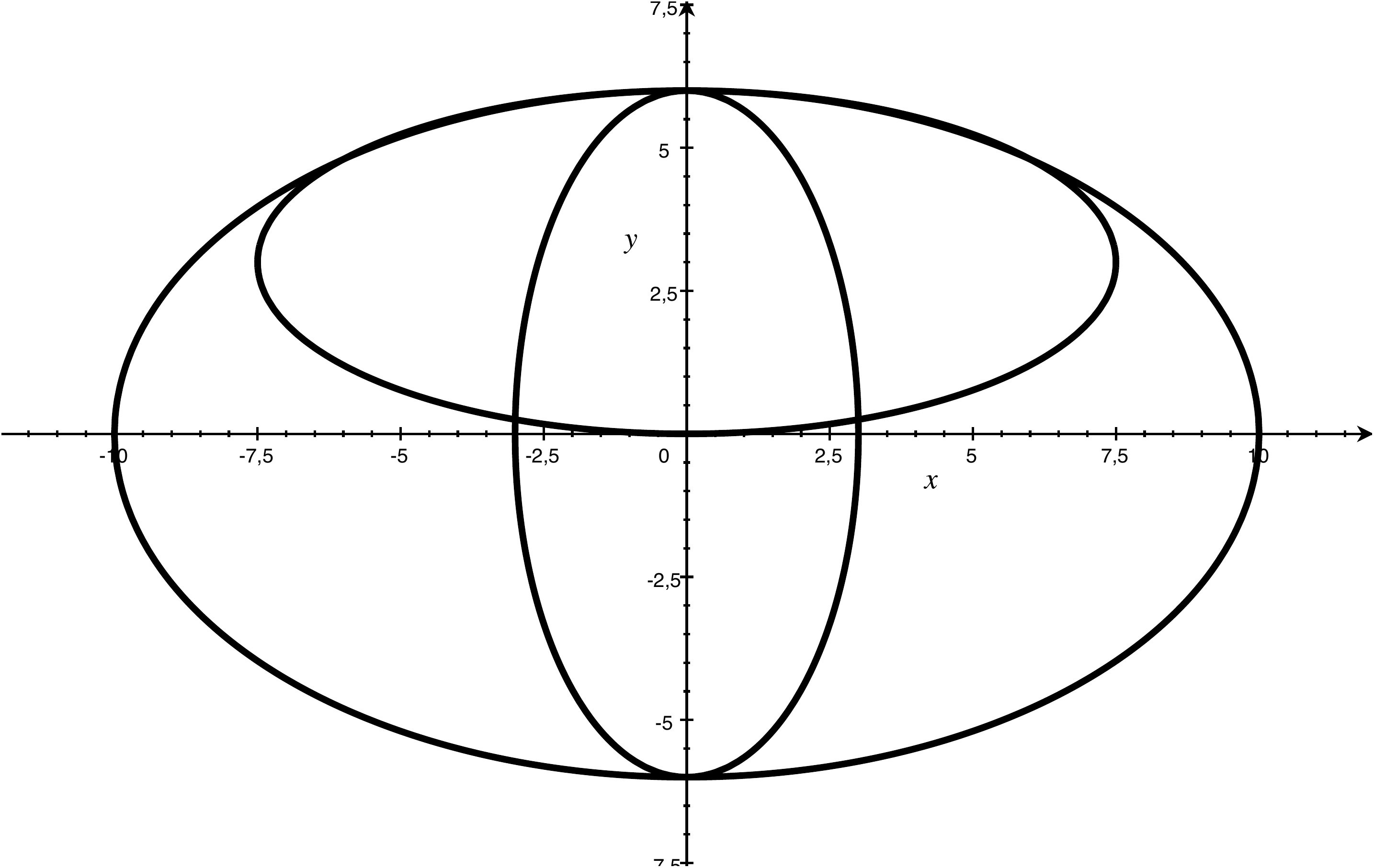}\\
(c) & & (d)
\end{tabular}
\caption{Compound curves: (a) citrus curve with $a=2$ coupled with a circumference; 
(b) citrus curve with $a=2$ coupled with the line $y=0$;
(c) curve with $5$-convexities coupled with a circumference with $a=2$, $b=1/4$, $r=1$;
(d) three ellipses with $a=10$, $b=6$.}\label{compoundCurves}
\end{center}
\end{figure}

Combining a citrus curve and a circle we get the new family of curves of equation:
\begin{eqnarray*}
\left (a^4y^2 + \left(x-\frac{a}{2} \right)^3 \left(x+\frac{a}{2} \right)^3 \right) \left(x^2+y^2 -\frac{a^2}{64}\right)=0
\end{eqnarray*}
with $a \in \mathbb{R}_>0$.
An example is provided in Figure \ref{compoundCurves}(a) where the parameter
is $a=2$.

Another possibility is to couple a citrus curve with a line (we choose one of the two axes of symmetry, 
for instance the $x$ axis) whose equation is:
\begin{eqnarray*}
y\left (a^4y^2 + \left(x-\frac{a}{2} \right)^3 \left(x+\frac{a}{2} \right)^3 \right)=0
\end{eqnarray*}
with $a\in \mathbb{R}_>0$.
An example is provided in Figure \ref{compoundCurves}(b) where the parameter
is $a=2$. An application of the use of this compound curve is given in Figure \ref{combineFig}(b).

It is also possible to associate a curve with $5$-convexities with a circumference; the resulting polar equation is:
\begin{eqnarray*}
\left ( \rho - \frac{a}{1+b \cos(5\theta)}\right ) (\rho-r)=0
\end{eqnarray*}
with $a,b, r \in \mathbb{R}_{>0}$, $b<1$ and $r\le \frac{a}{1+b}$. 
An example is provided in Figure \ref{compoundCurves}(c).

Finally, starting with three ellipses we construct a family of algebraic whose equation is:
\begin{eqnarray*}
\left( \frac{x^2}{a^2}+\frac{y^2}{b^2}  -1\right) \left( \frac{16x^2}{9a^2}+\frac{(2y-b)^2}{b^2}  -1\right)
\left( \frac{4x^2}{b^2}+\frac{y^2}{b^2}  -1\right)=0
\end{eqnarray*}
with $a, b \in \mathbb{R}_{>0}$. 
An example is provided in Figure~\ref{compoundCurves}(d). 

\section{Examples and conclusive remarks}
\label{examples}
Our method has been tested on a collection of artefacts and models collected from the web, 
the AIM@SHAPE repository \cite{VISIONAIR}, the STARC repository \cite{STARC} and the 3D dataset of the EPSRC project \cite{projChevarria}. 

\begin{figure}[htbp]
\centering
\begin{tabular}{ccccccc}
I. & &
\includegraphics[height=2.2cm]{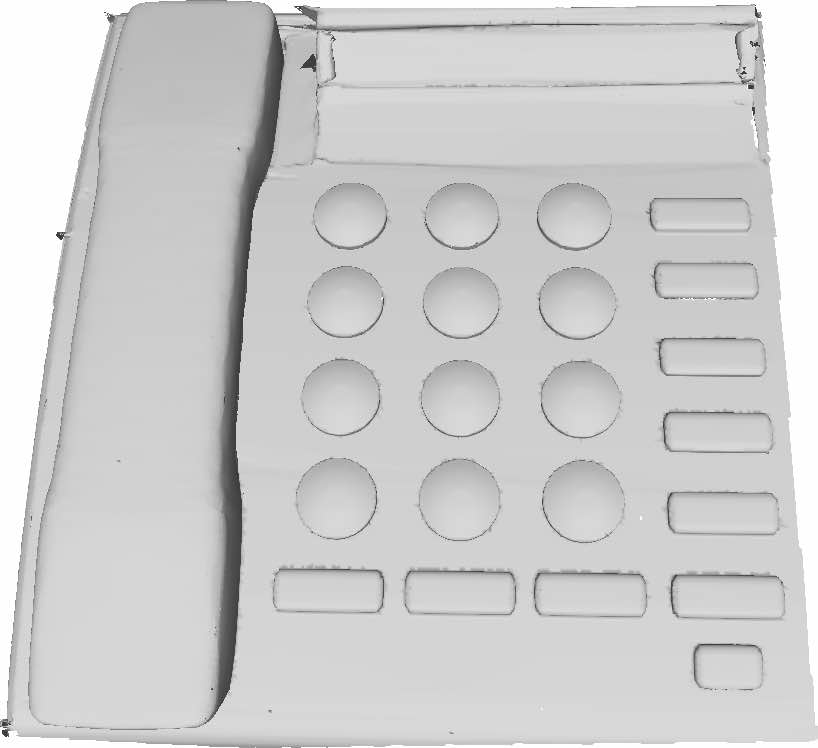} & &
\includegraphics[height=2.2cm]{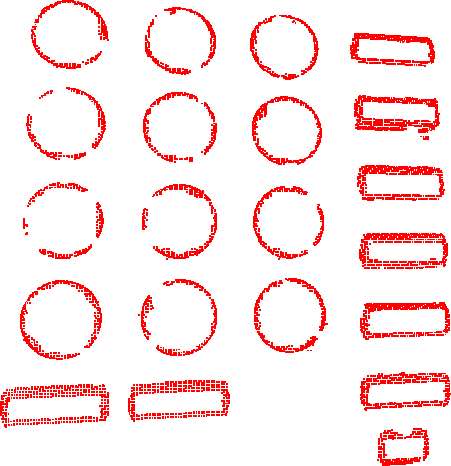} & &
\includegraphics[height=2.2cm]{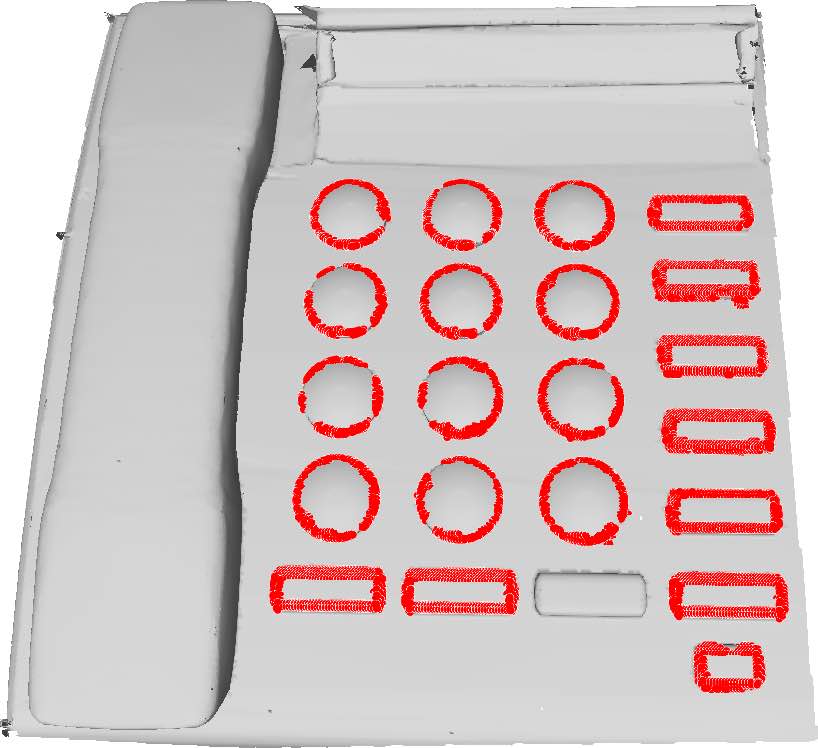} \\
&& (a) & & (b) & & (c)\\
\\
II. & & 
\includegraphics[height=3cm]{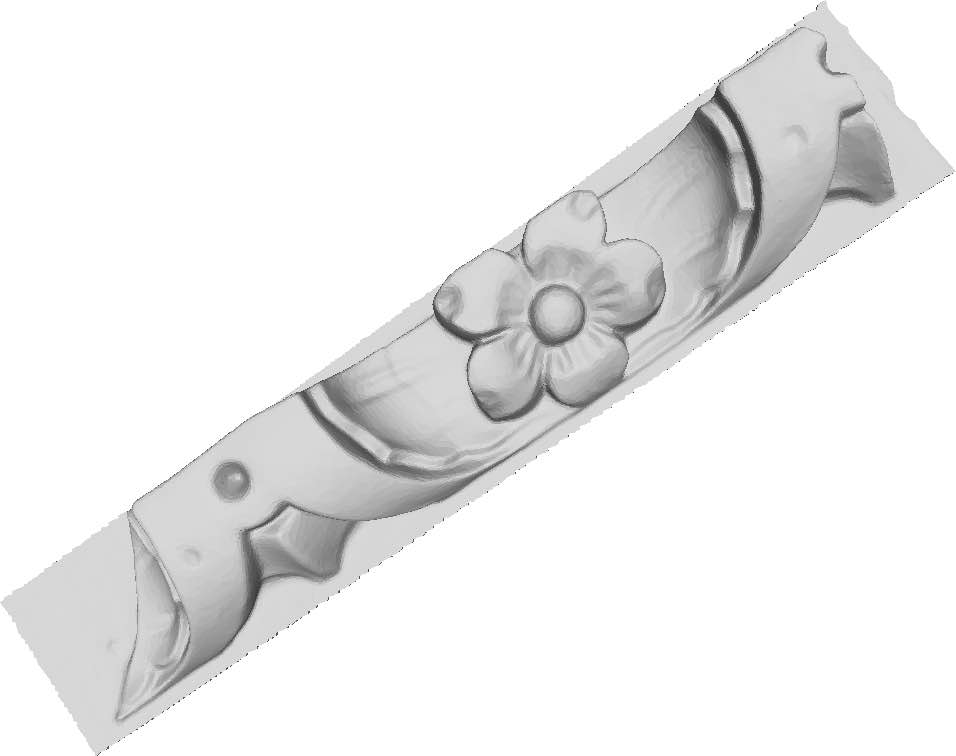} & &
\includegraphics[height=1.5cm]{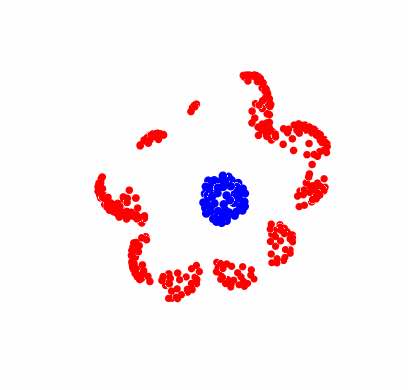} & &
\includegraphics[height=3cm]{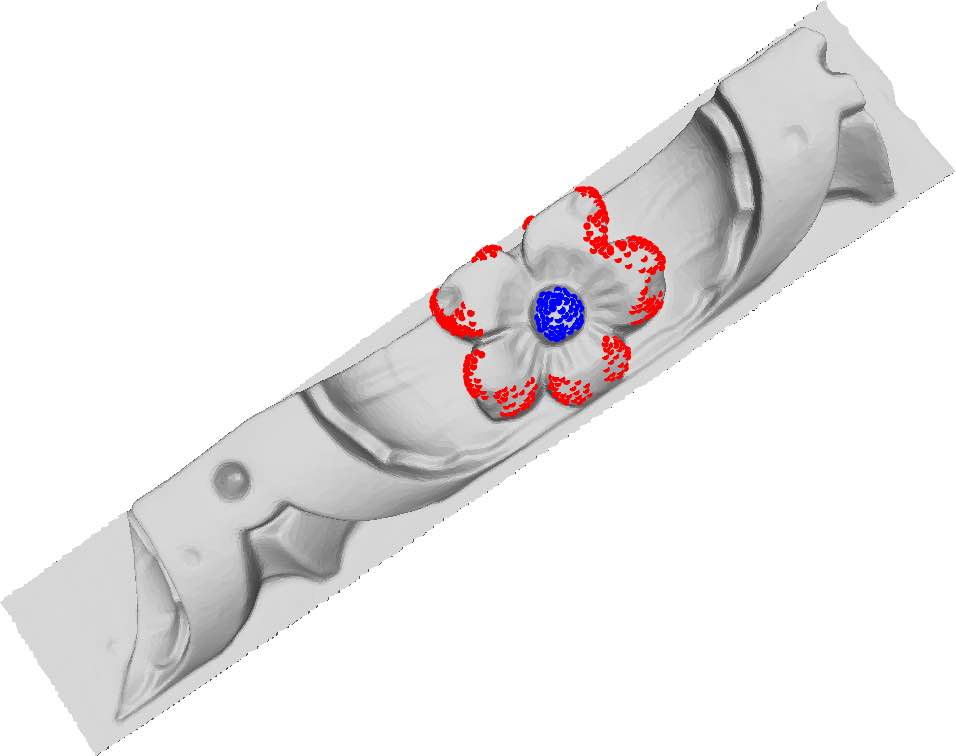} \\
&& (a) & & (b) & & (c)\\
\\
III. & &
\includegraphics[height=1.5cm]{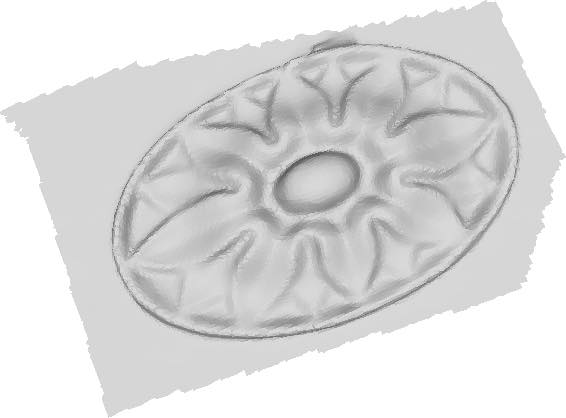} & &
\includegraphics[height=1.5cm]{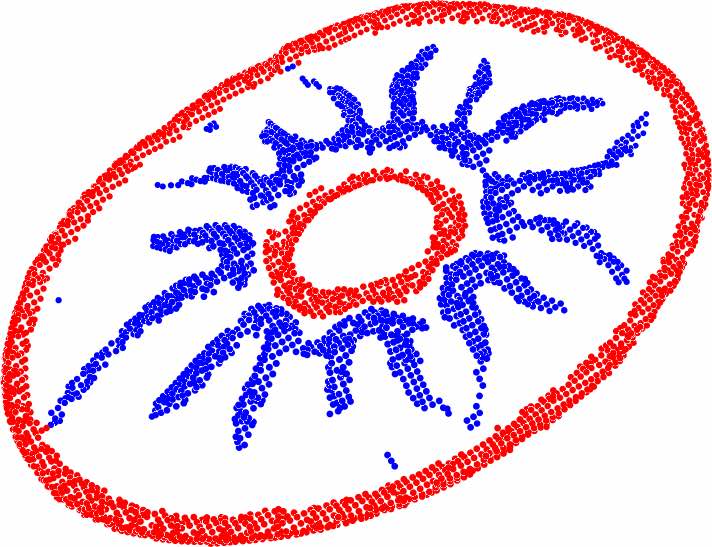} & &
\includegraphics[height=1.65cm]{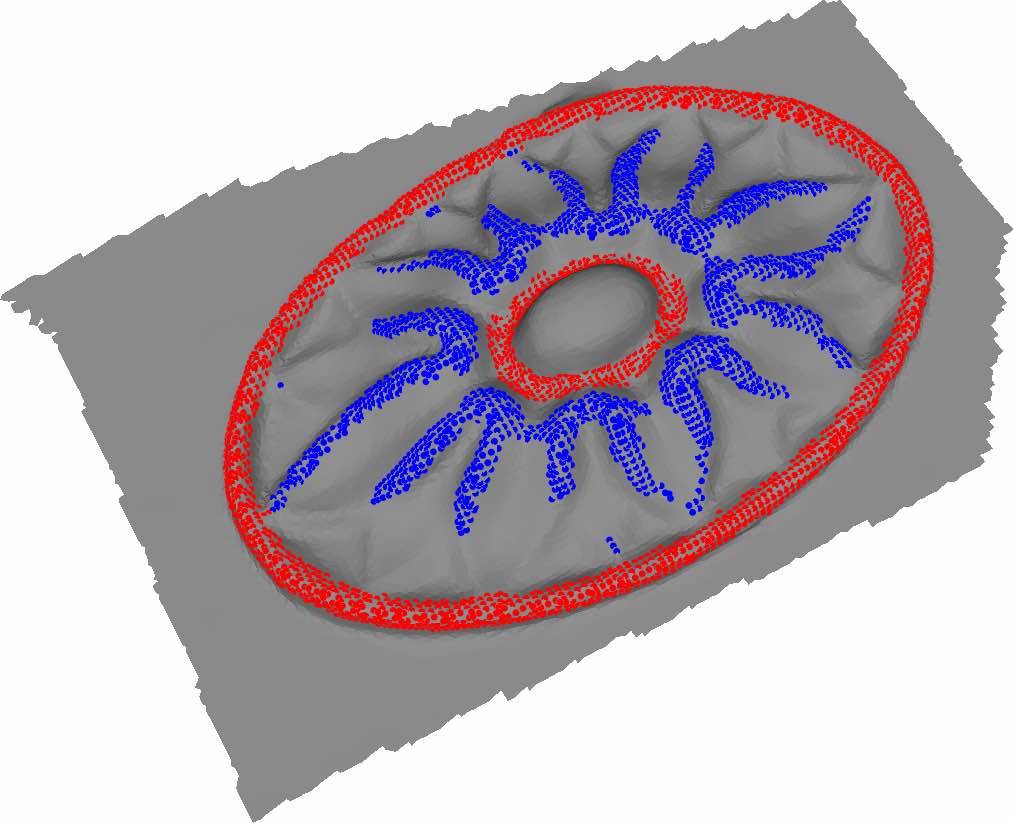} \\
&& (a) & & (b) & & (c)\\
\\
IV. & &
\includegraphics[height=2.6cm]{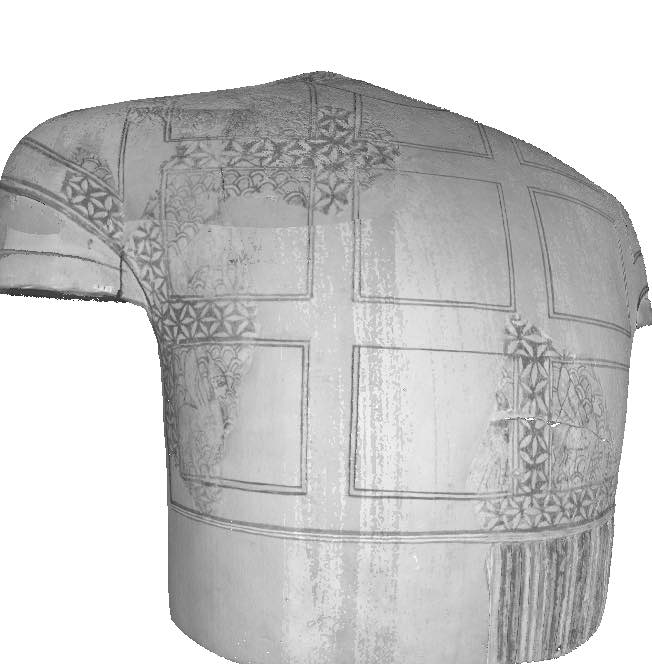} & &
\includegraphics[height=1.5cm]{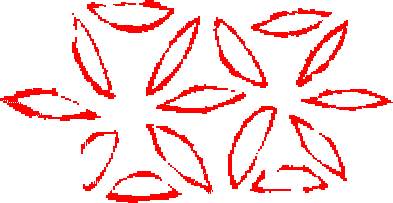} & &
\includegraphics[height=2.2cm]{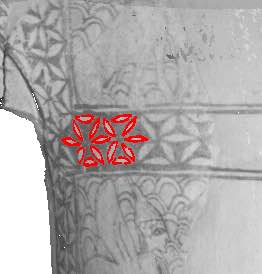} \\
&& (a) & & (b) & & (c)\\
\end{tabular}
\caption{I. circles and curves of Lamet on a phone model;  
II. a circle and a curve with $5$-convexities on an architectural ornamental artefact (from the 3D dataset of the EPSRC 
project \cite{projChevarria}); 
III. a curve with $8$-convexities contained in a region delimited by two concentric circles on an architectural 
ornamental artefact (from the 3D dataset of the EPSRC project \cite{projChevarria}); 
IV. decorations on a bust (from the STARC repository \cite{STARC}).
}\label{figMultipleII}
\end{figure}

\begin{figure}[htbp]
\centering
\begin{tabular}{ccccc}
I. &
\includegraphics[height=3cm]{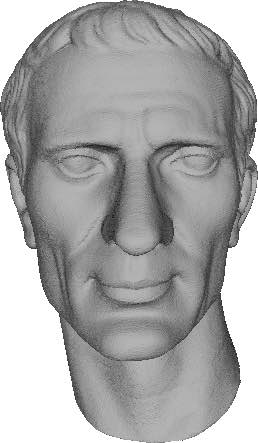} &
\includegraphics[height=1.5cm]{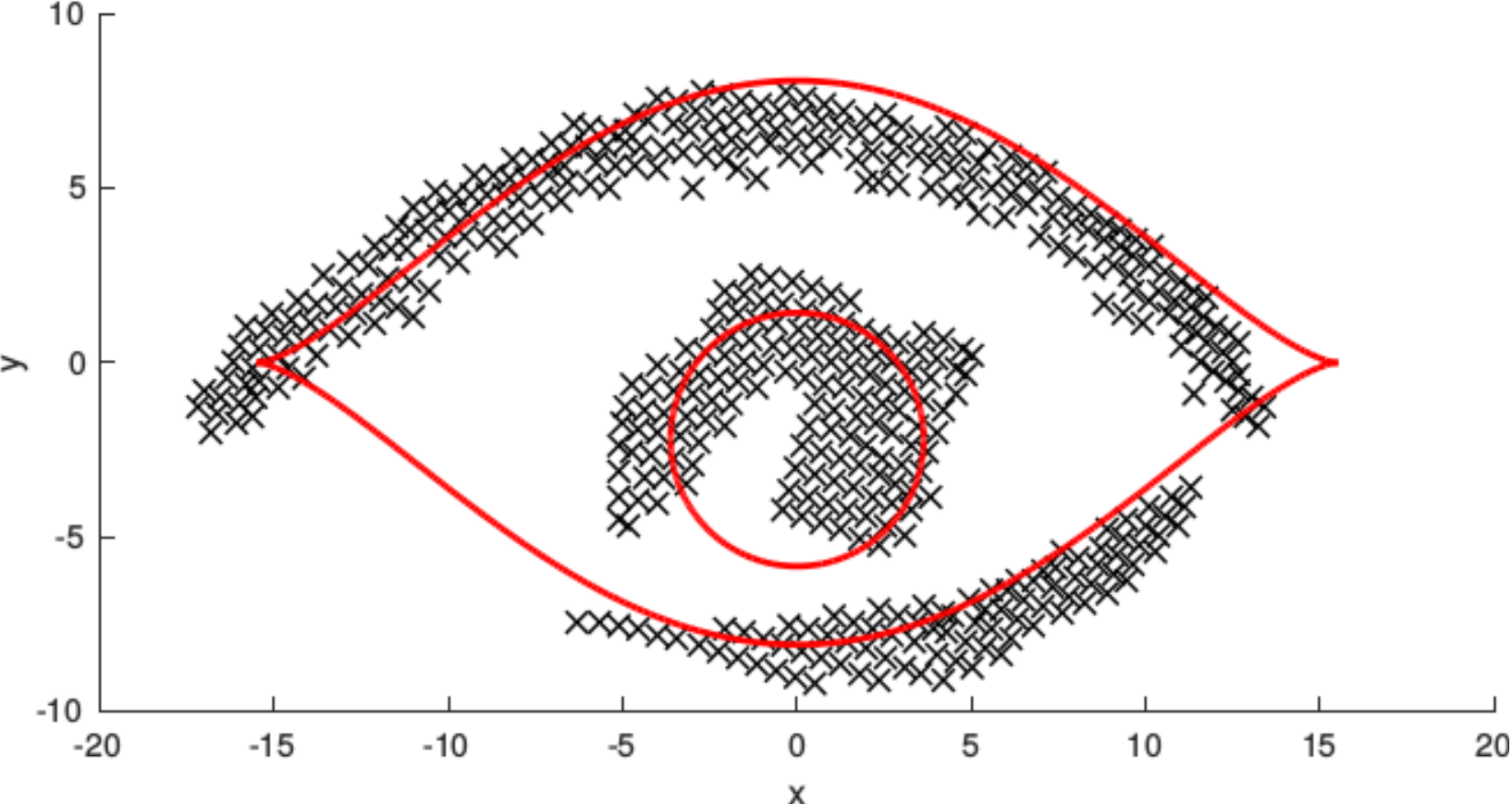} &
\includegraphics[height=1.5cm]{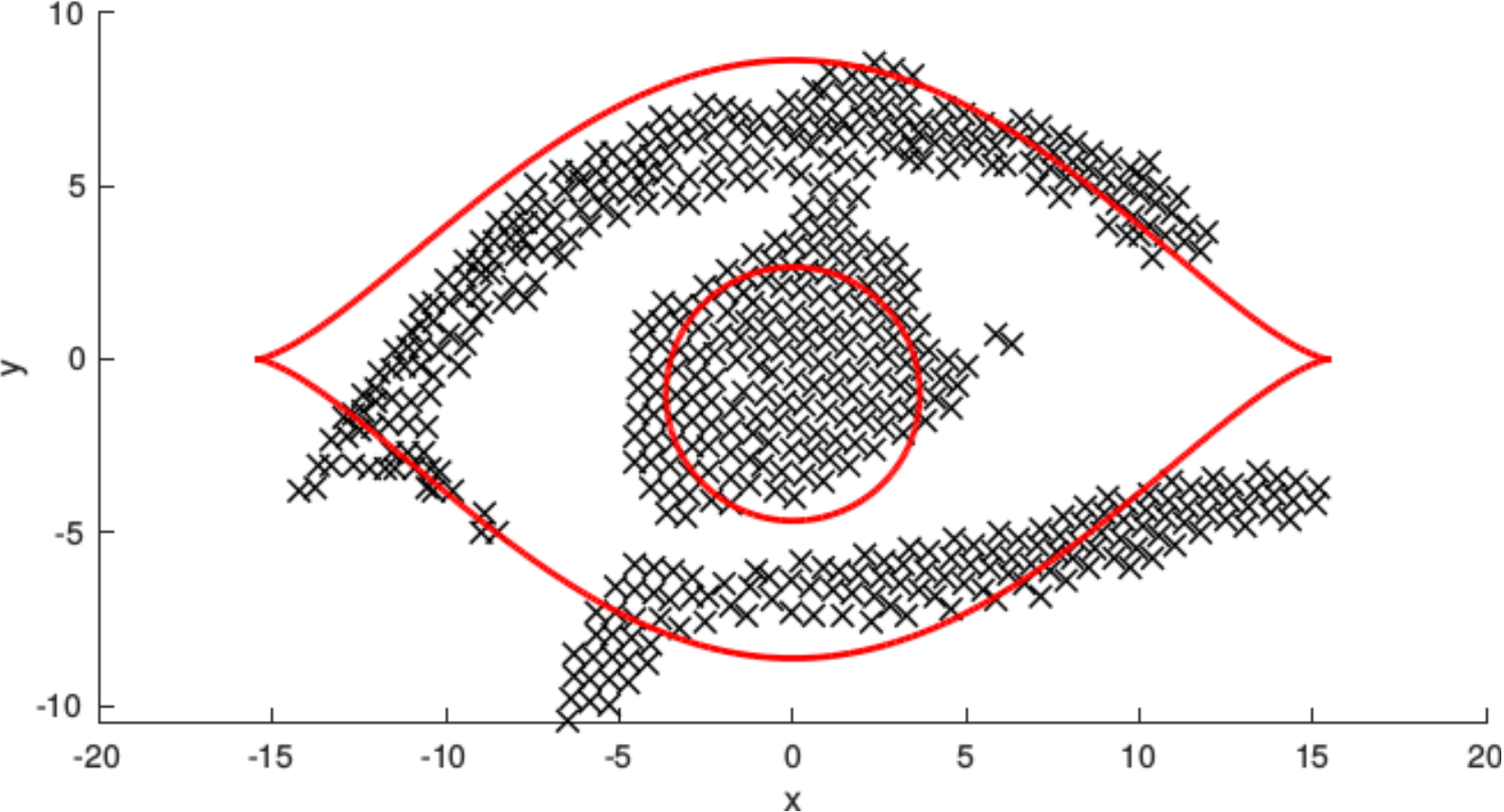} & 
\includegraphics[height=3cm]{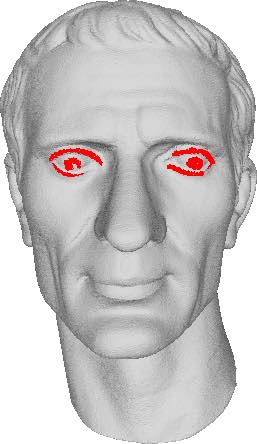}\\
& (a) & (b) & (c) & (d)\\
\\
II. &
\includegraphics[height=3cm]{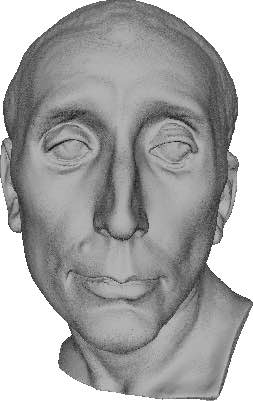} &
\includegraphics[height=1.5cm]{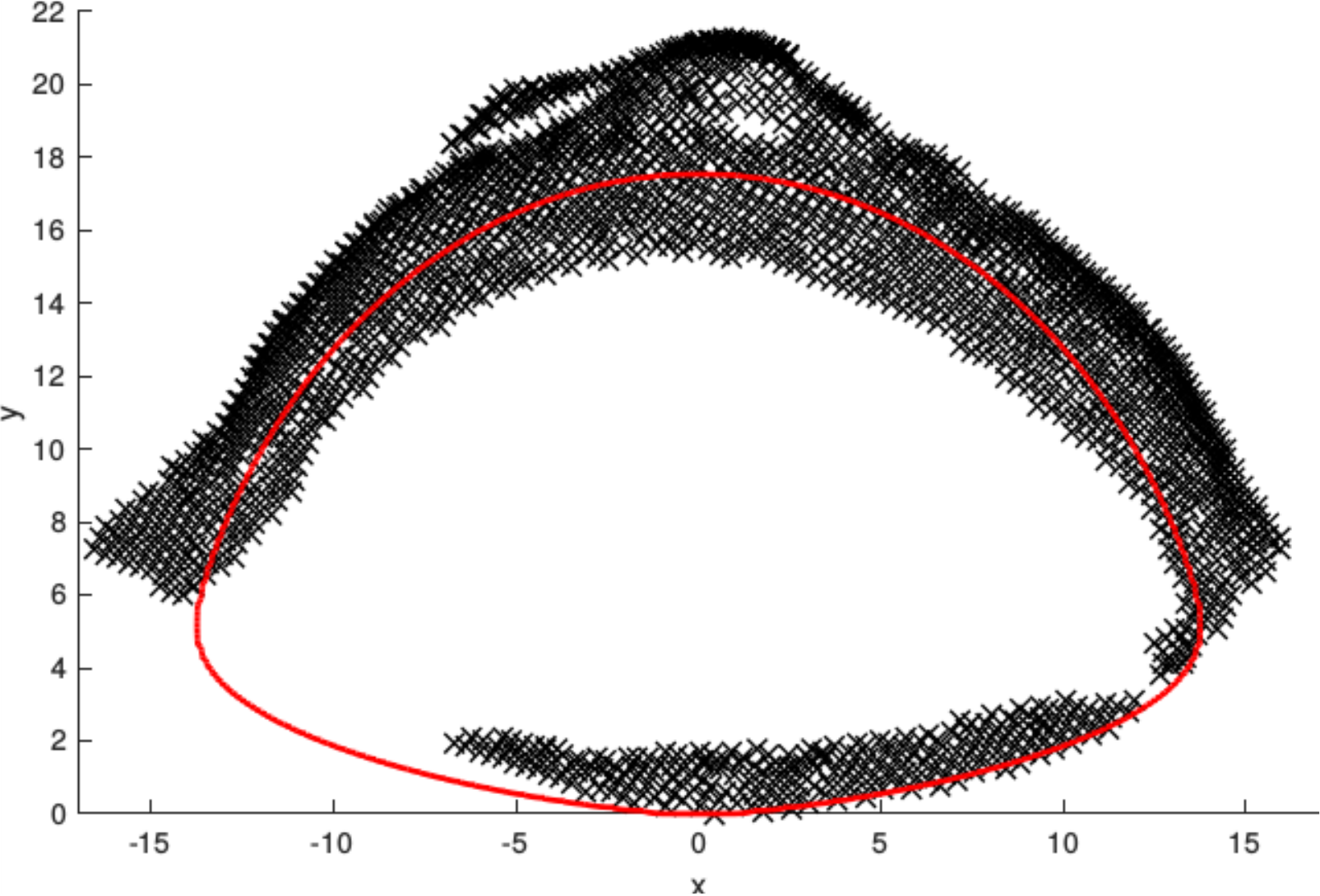} &
\includegraphics[height=1.4cm]{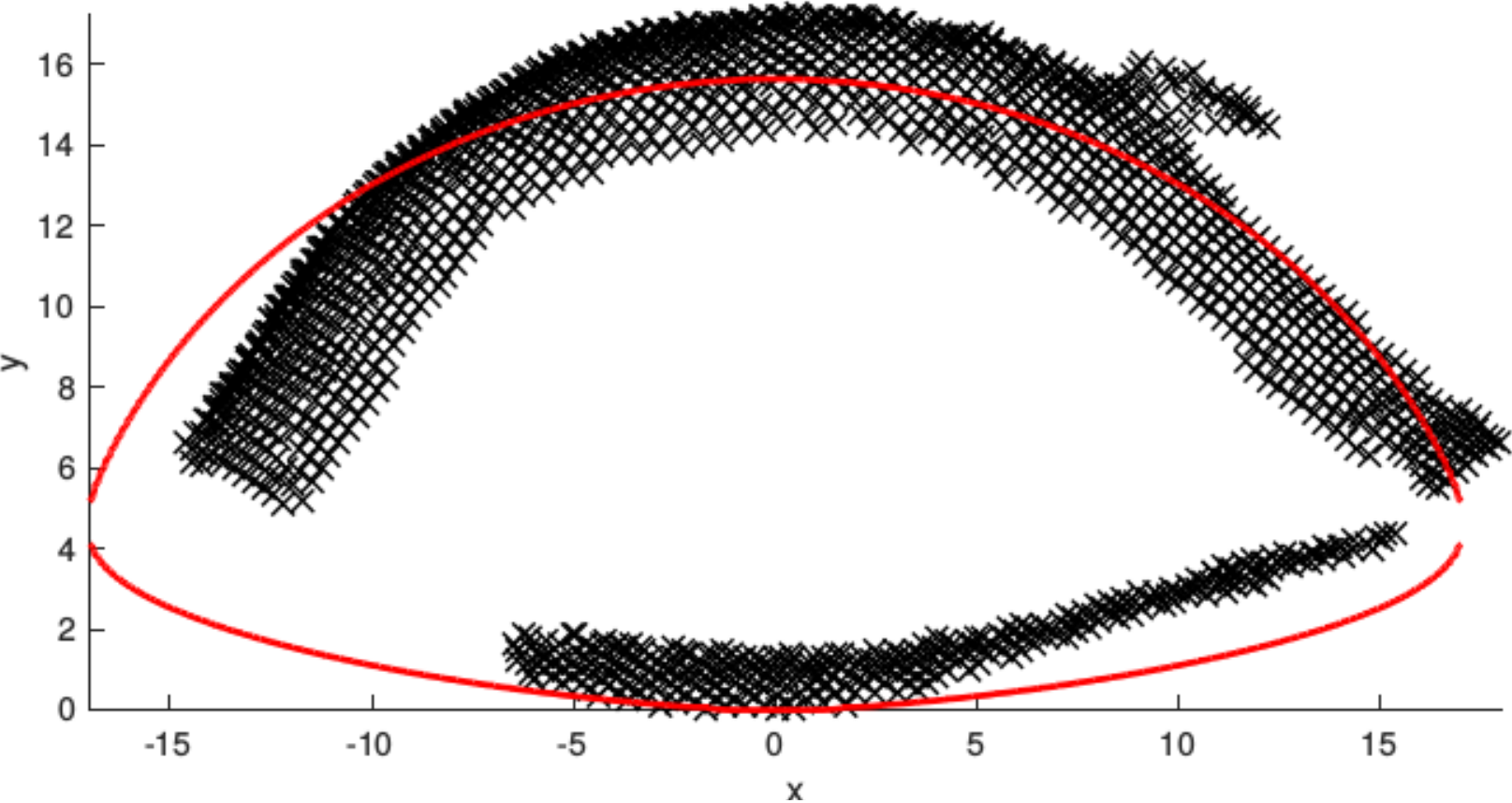} &
\includegraphics[height=3cm]{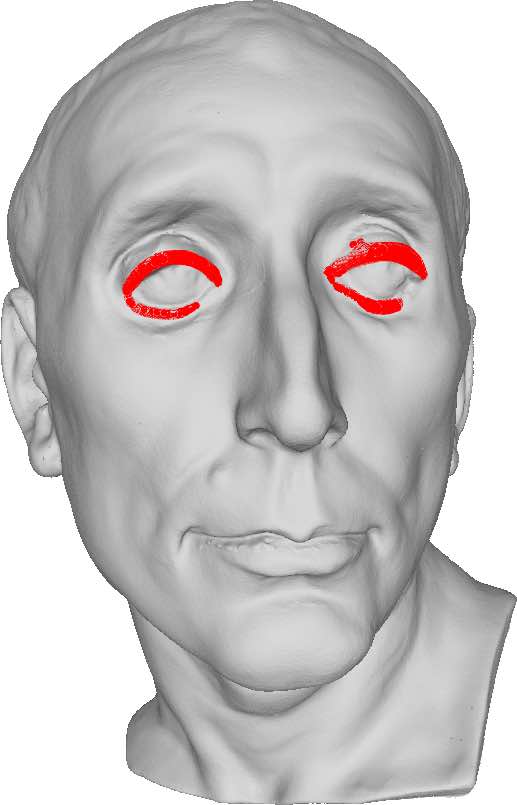} \\
& (a) & (b) & (c) & (d)\\
\\
III. &
\includegraphics[height=2.75cm]{fig_Nefertiti_soloModello2} &
\includegraphics[height=1.25cm]{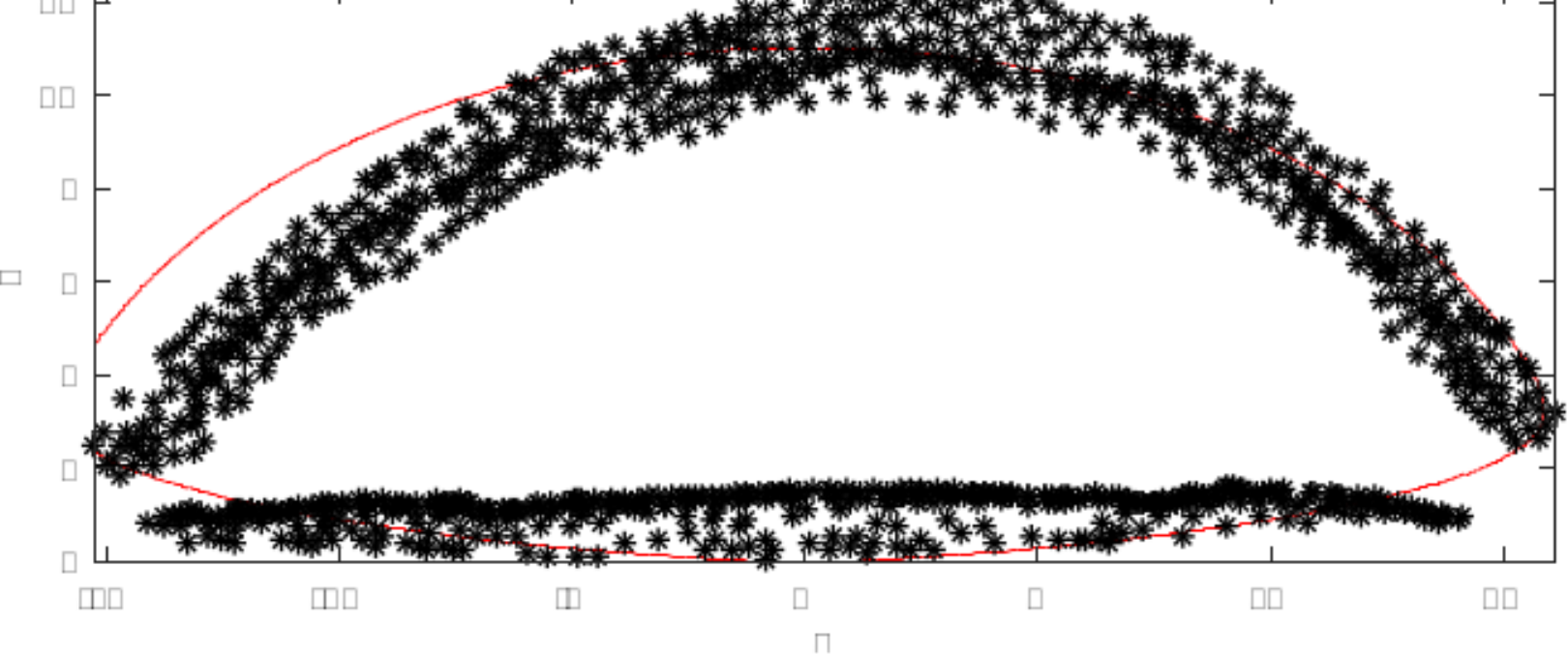} &
\includegraphics[height=1.25cm]{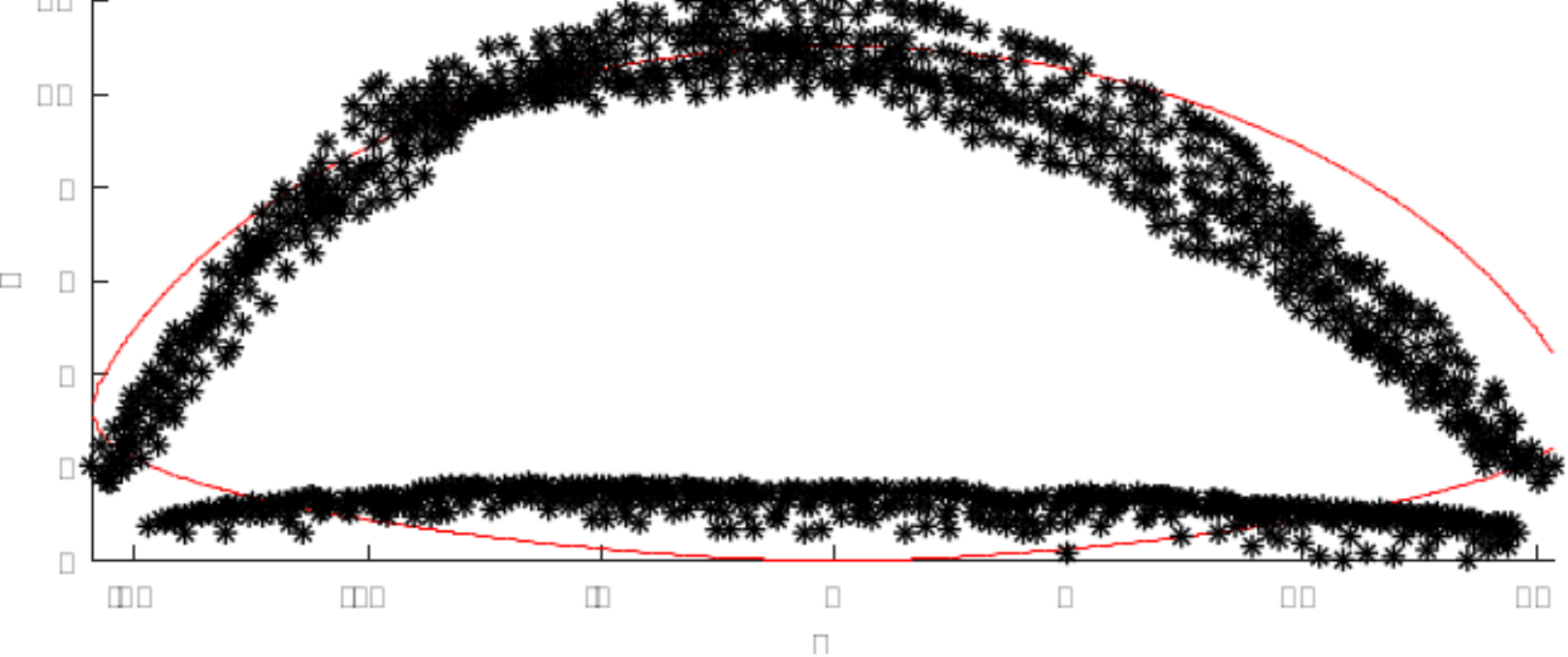} &
\includegraphics[height=2.75cm]{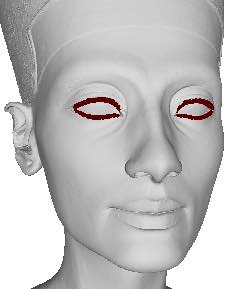} \\
& (a) & (b) & (c) & (d)\\
\\
IV. & 
\includegraphics[height=2.25cm]{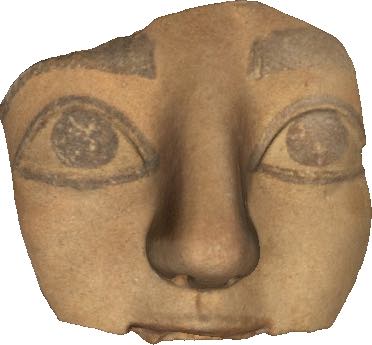} &
\includegraphics[height=1.75cm]{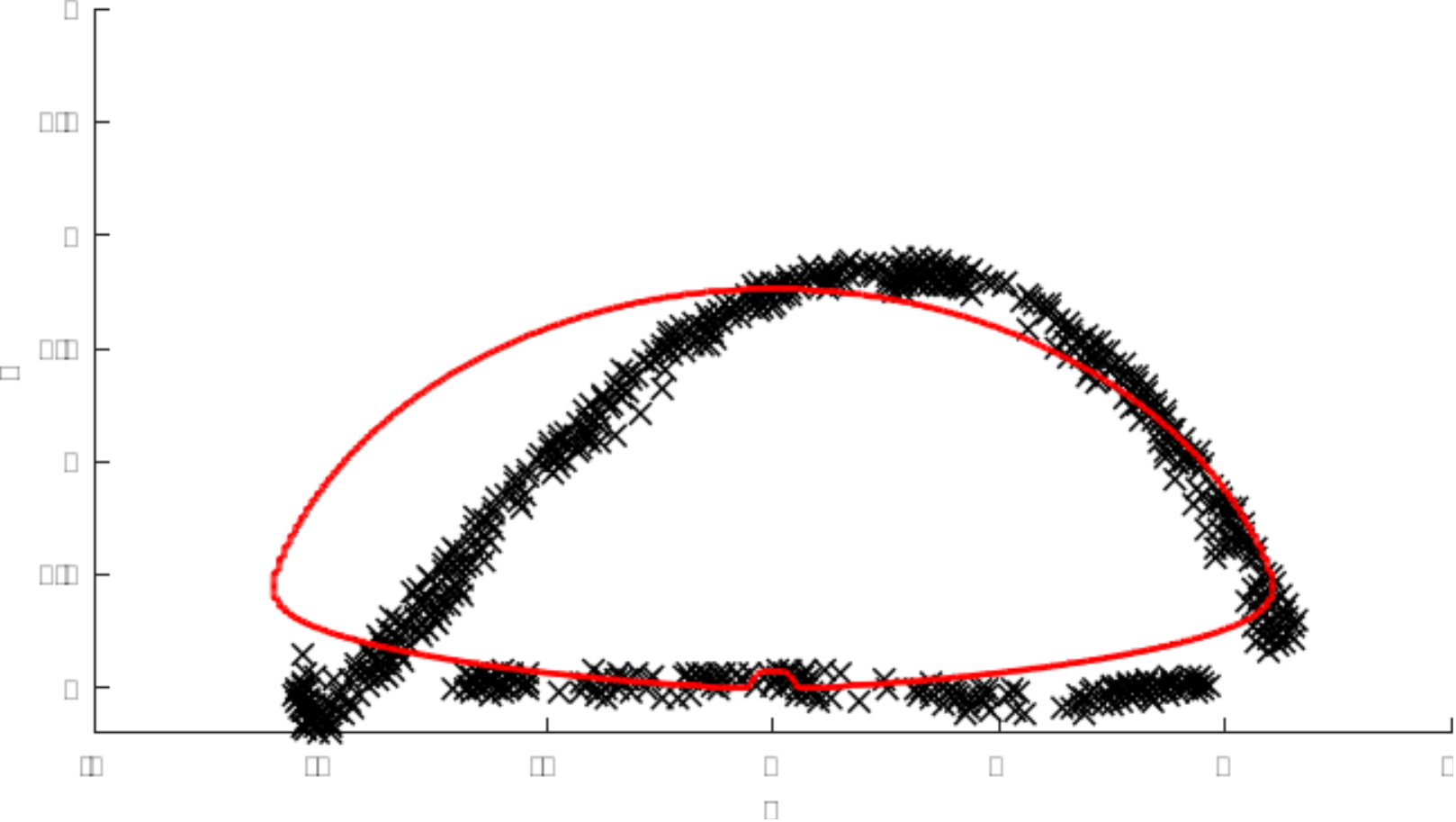} &
\includegraphics[height=1.75cm]{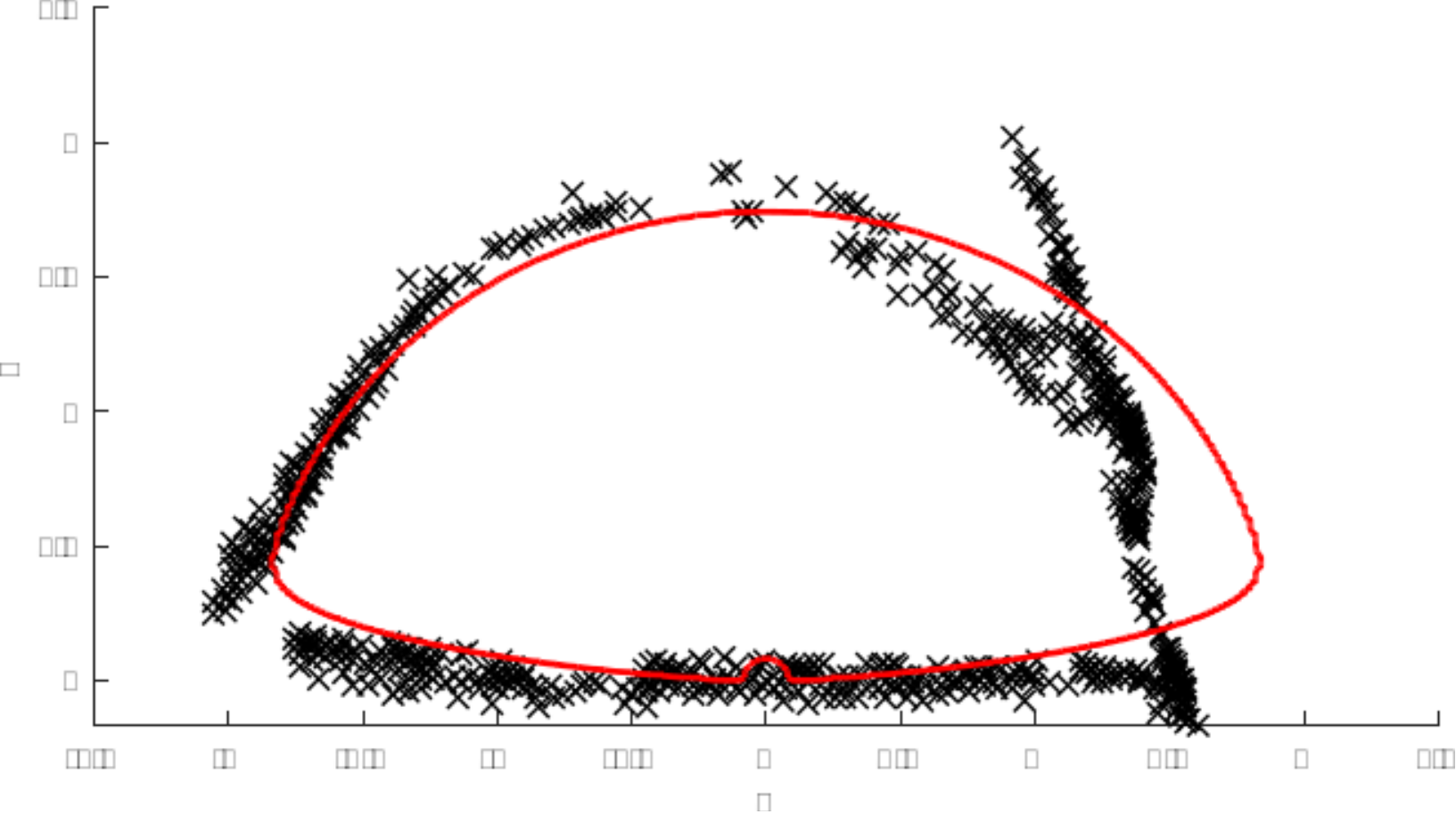} &
\includegraphics[height=2.25cm]{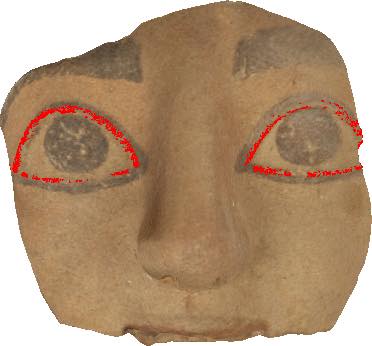} \\
& (a) & (b) & (c) & (d)\\
\end{tabular}
\caption{Detection of eye contours and pupils on collection of artefacts collected from  the AIM@SHAPE 
repository~\cite{VISIONAIR} and the STARC repository \cite{STARC}. }\label{figEyes}
\end{figure}

\begin{figure}[htb]
\centering
\begin{tabular}{ccccc}
I. & \includegraphics[height=2cm]{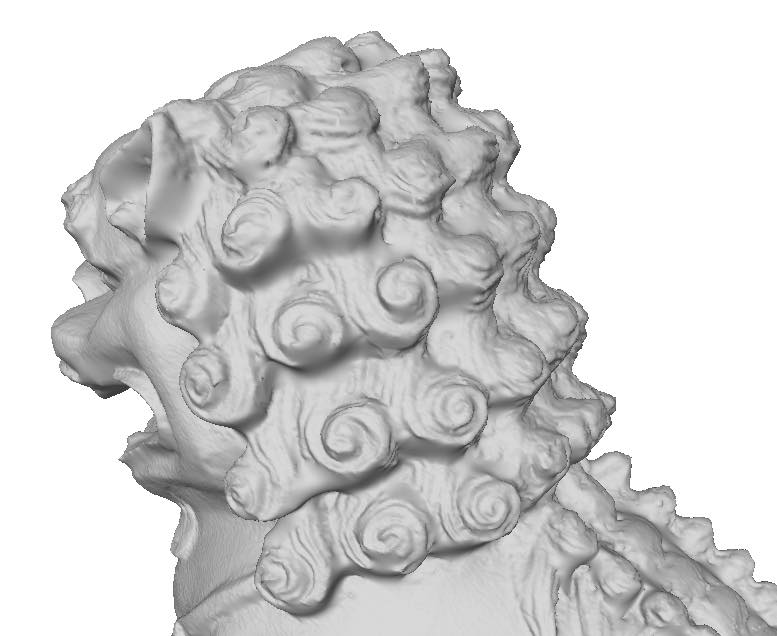} & 
 \includegraphics[height=2cm]{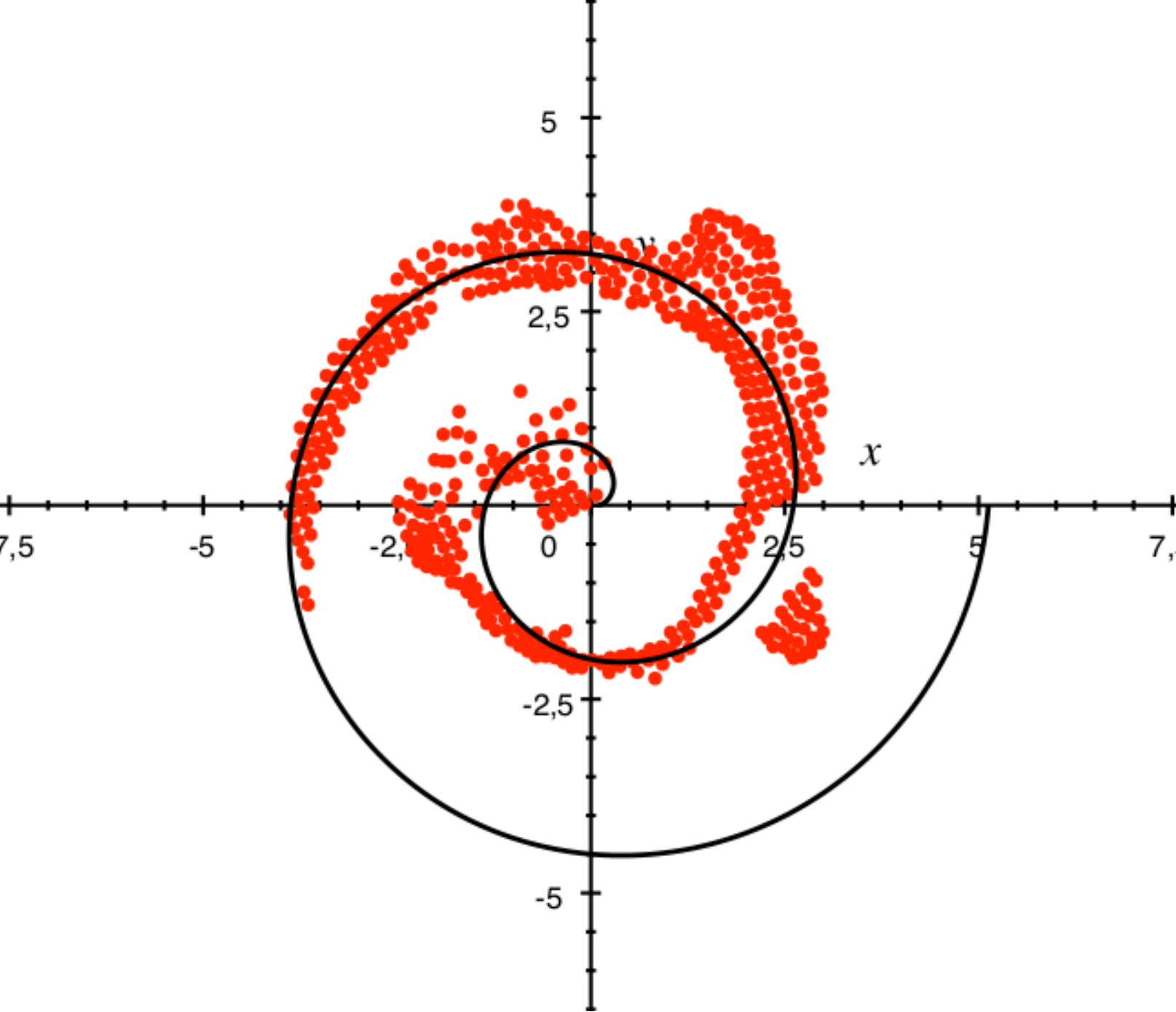} & 
  \includegraphics[height=2cm]{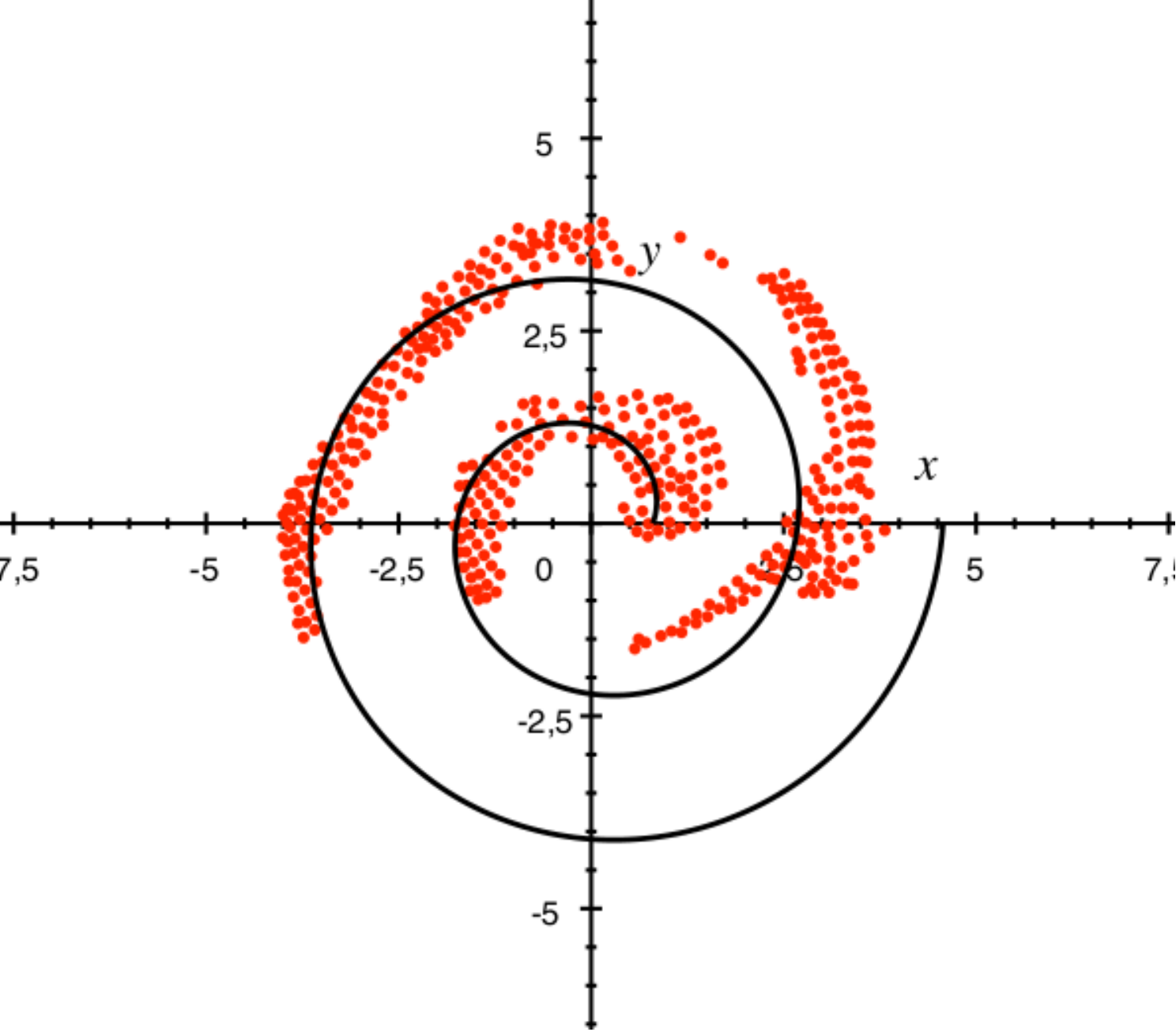} & 
 \includegraphics[height=2cm]{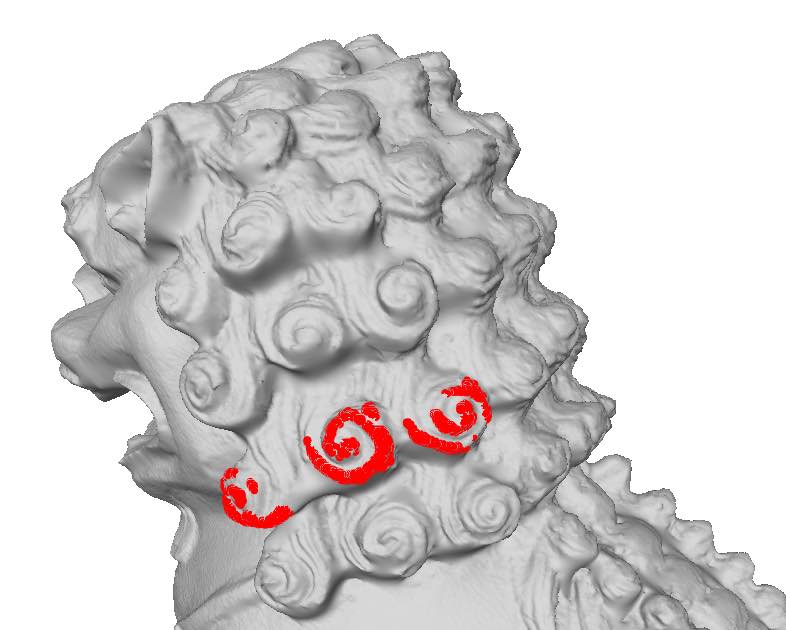}\\
& (a) & (b) & (c) & (d)\\
\\
II. & \includegraphics[height=2.5cm]{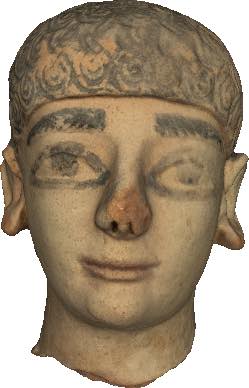} & 
\includegraphics[height=1.5cm]{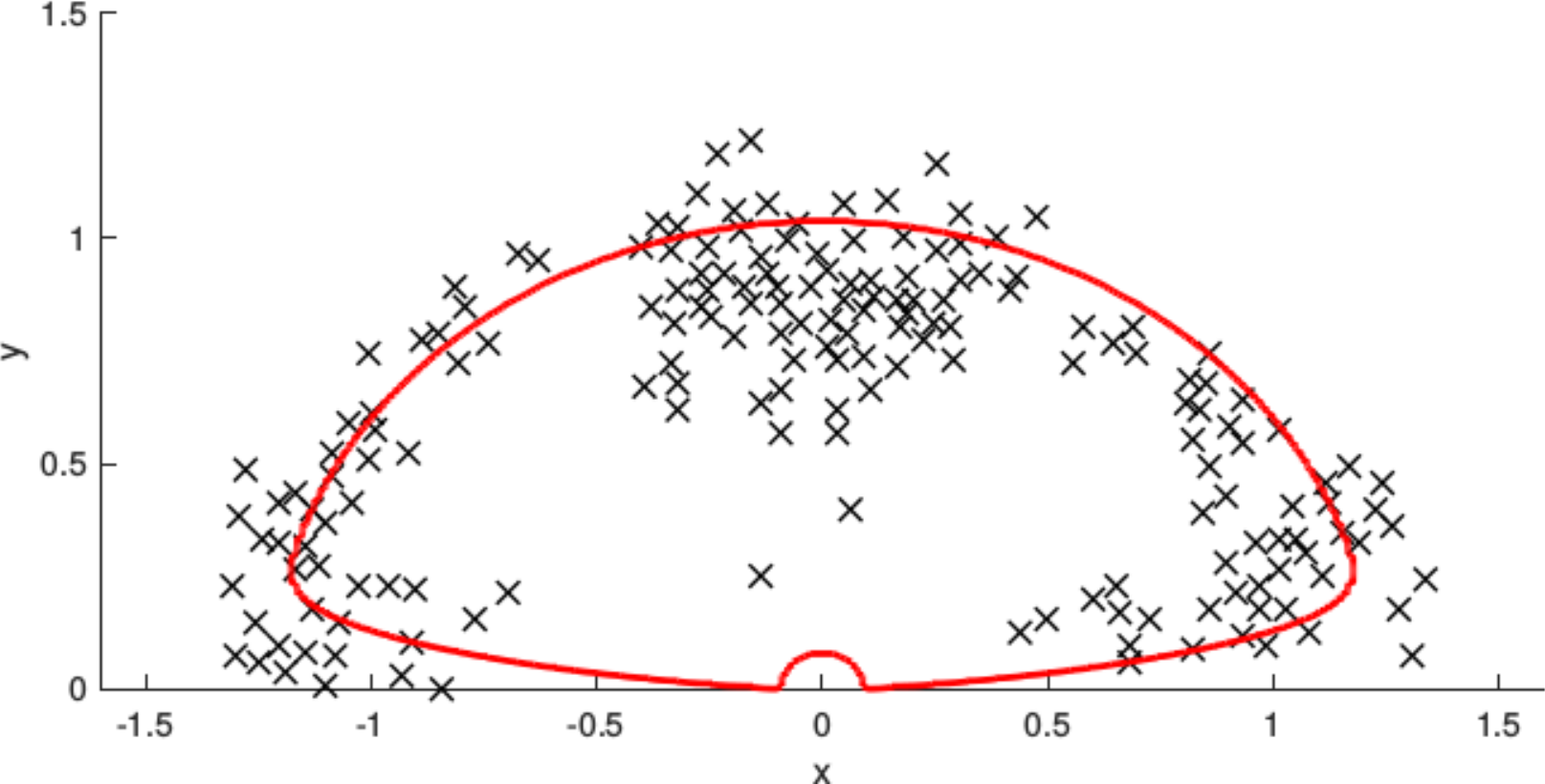} & 
\includegraphics[height=1.25cm]{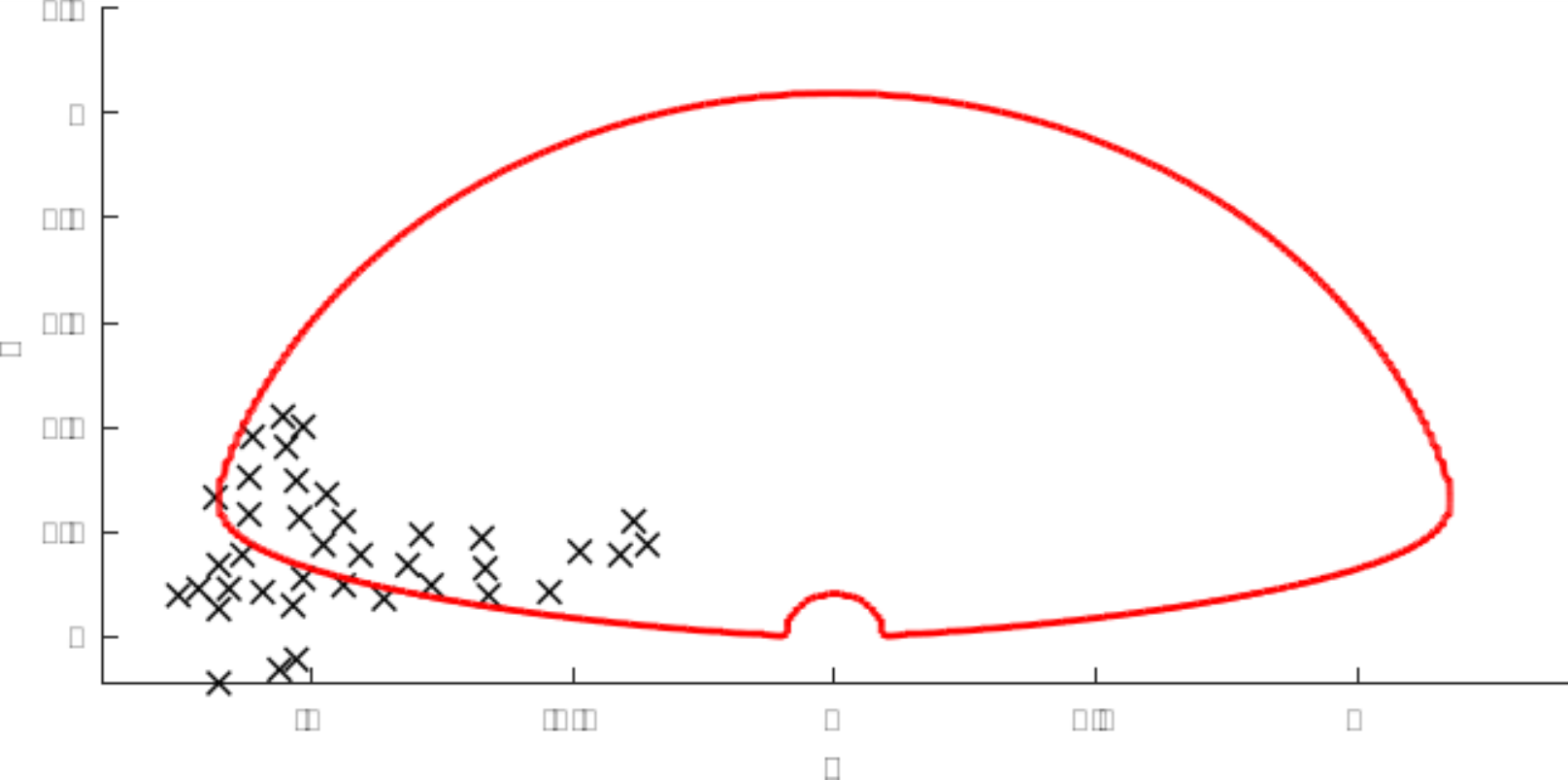} & 
\includegraphics[height=2.5cm]{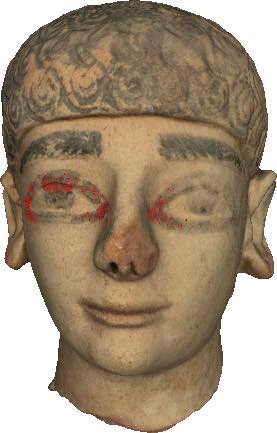}\\
& (a) & (b) & (c) & (d)\\
\end{tabular}
\caption{Feature recognition on models with degraded features.}\label{noise}
\end{figure}

\begin{figure}[hbt]
\centering
\begin{tabular}{cccccc}
I. & & \includegraphics[height=2.6cm]{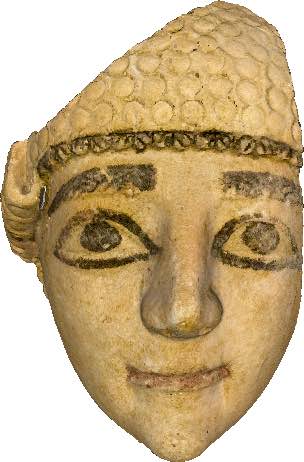} & 
\includegraphics[height=1.5cm]{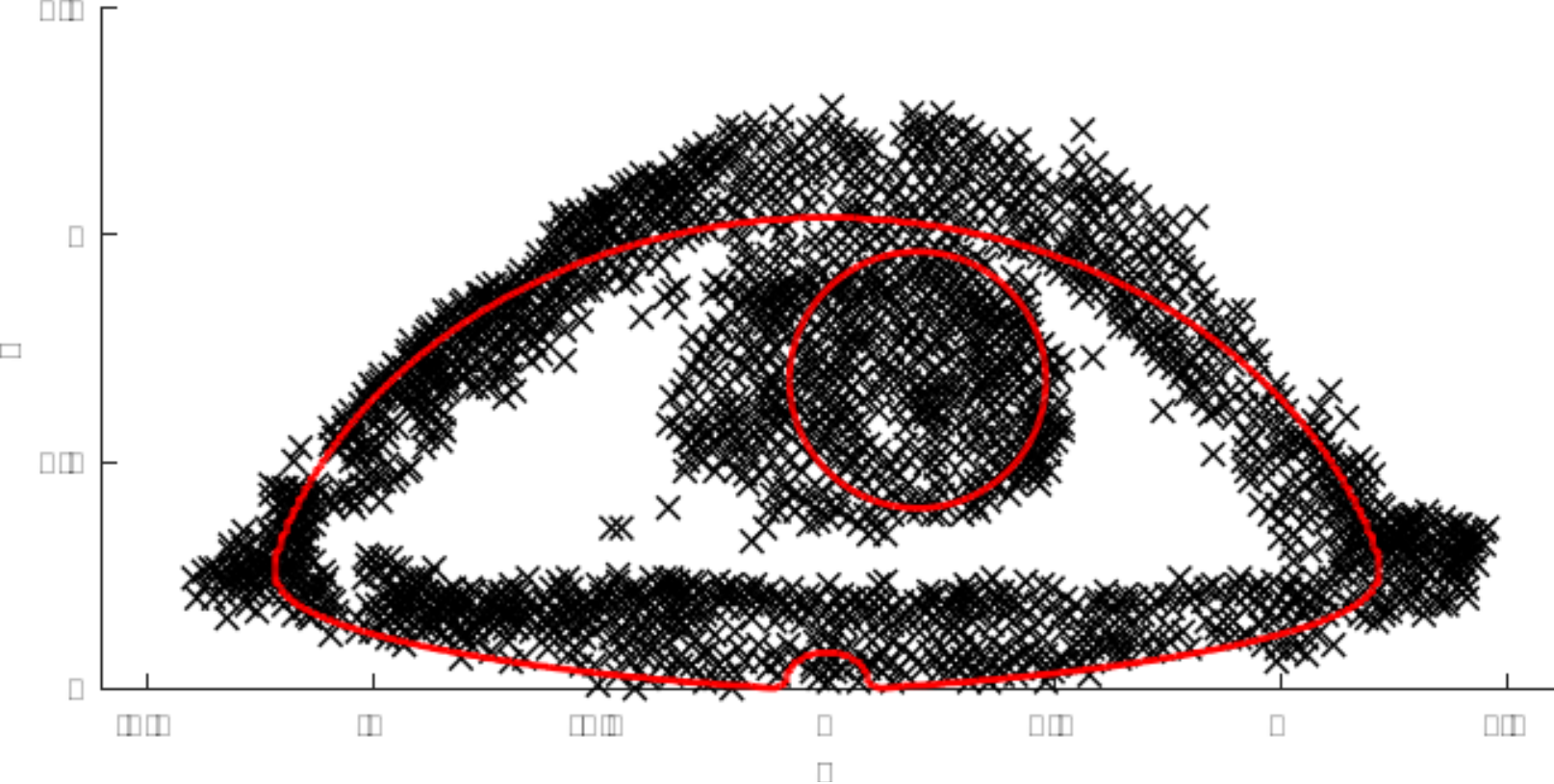} &
\includegraphics[height=1.5cm]{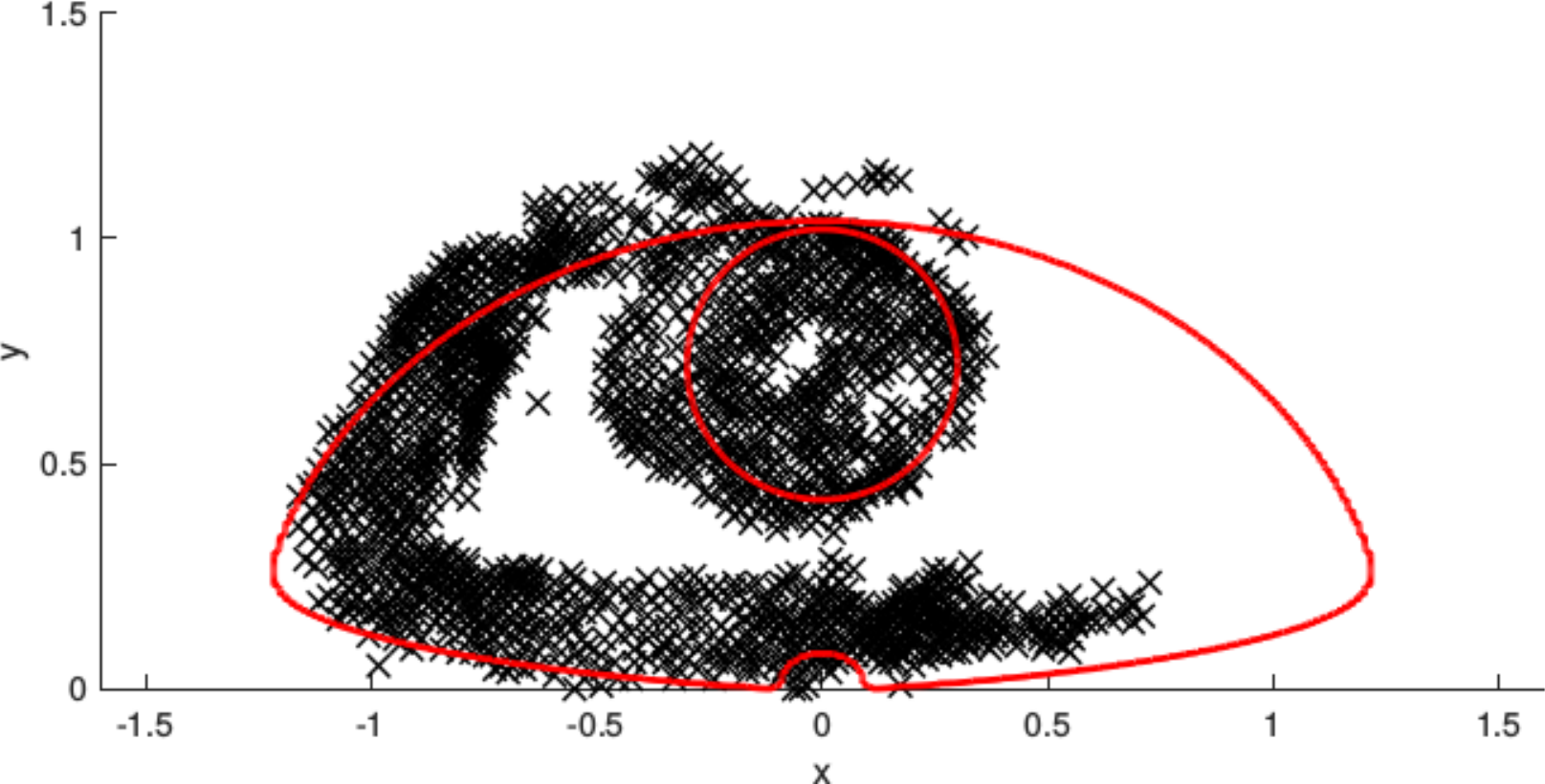} &
\includegraphics[height=2.6cm]{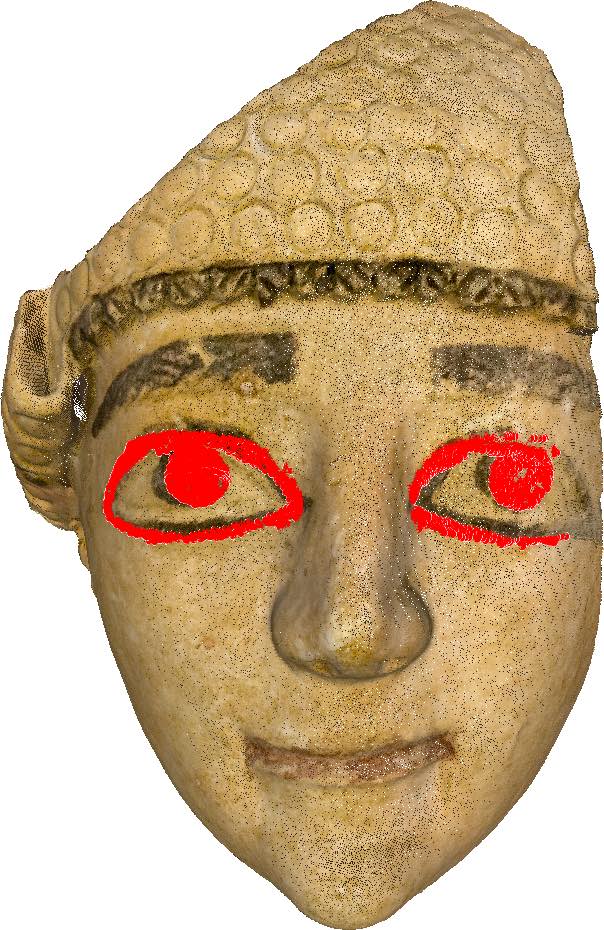}\\
& & (a) & (b) & (c) & (d)\\
II. & & \includegraphics[height=2.3cm]{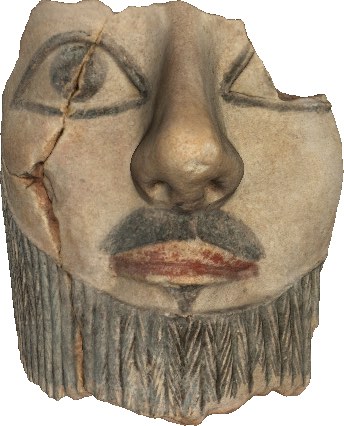} & 
\includegraphics[height=1.5cm]{fig_C_111_1935_eye_geomPetal_circumf} &
\includegraphics[height=1.5cm]{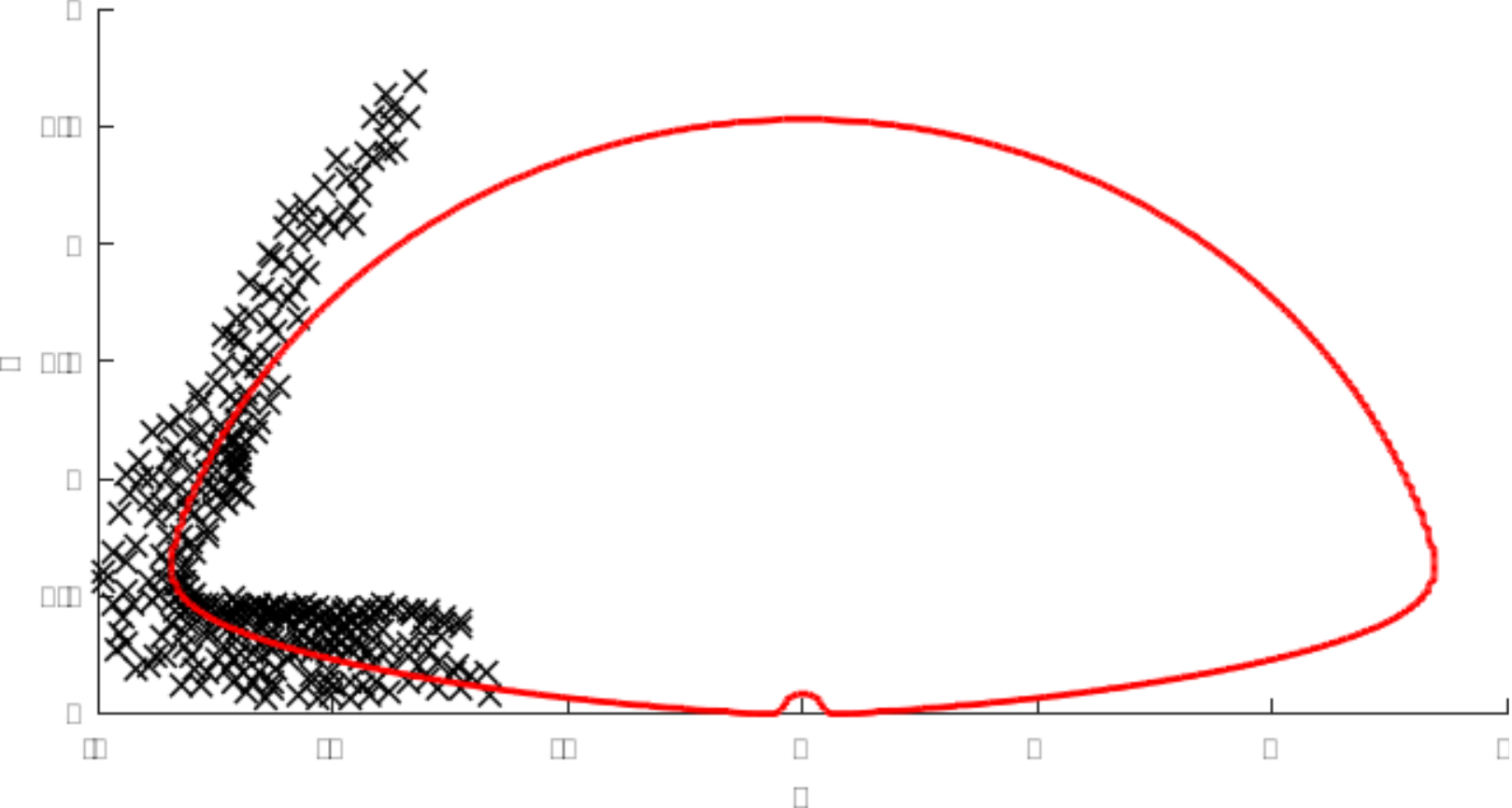} &
\includegraphics[height=2.3cm]{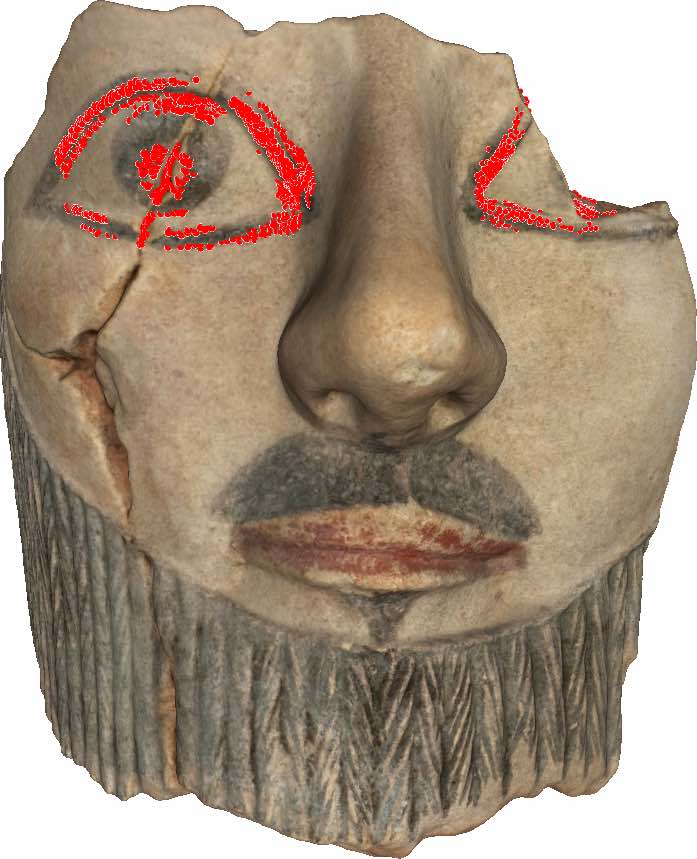}\\
& & (a) & (b) & (c) & (d)\\
\end{tabular}
\caption{Feature recognition on models (from the STARC repository \cite{STARC}) with partial features.}\label{partial}
\end{figure}

Most of the models are triangulated meshes; an exception is represented by the model in Figure \ref{partial}(I.a),
which is given as a point cloud. 
Further, in the examples in Figure \ref{figEyes}(IV.a) and \ref{partial}(I.a) the colorimetric information is available and combined with the curvature; in the models in Figure \ref{figMultipleII}(IV.a) and Figure \ref{noise}(II.a) features are only characterized by the dark colour of the decoration; the other models come without any colorimetric information and we adopt the maximum, minimum and mean curvature properties.
In all the examples, the model embedding in the $3$-dimensional space is completely random and the ``best" view shown in the pictures is artificially reported for the user's convenience. 
Furthermore, the features we identify are view-independent and represented by curves that can be locally projected onto a plane without any overlap.

For the feature identification, we assume to know in advance the class of features that are present in an object (for instance because there exists an archeological description of the artwork), thus converting the problem into a recognition of definite curves that identifies specific parts such as mouth, eyes, pupils, decorations, buttons, etc. on that model. Once aggregated and projected onto a plane, each single feature group of points is fitted with a potential feature curve and the value of the HT aggregation function is kept as the voting for that. After all curves are run, we keep the highest vote to select the curve that better fits with that feature. Sometimes more than one curve potentially fits the feature point set; in this case we have selected the curve with the highest value of the HT aggregation function.

\begin{figure}[htbp]
\centering
\begin{tabular}{ccccccc}
I. & & 
\includegraphics[height=2.7cm]{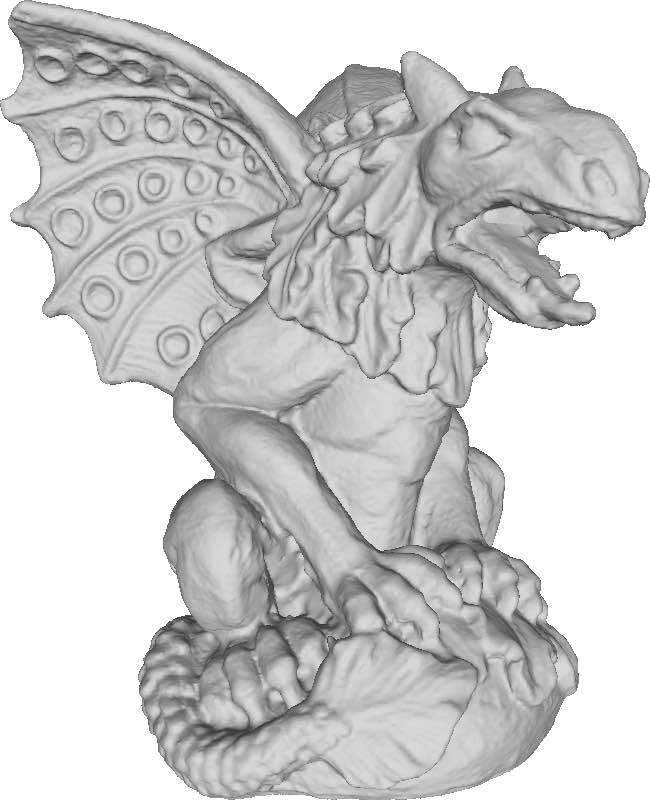} & &
\includegraphics[height=2.5cm]{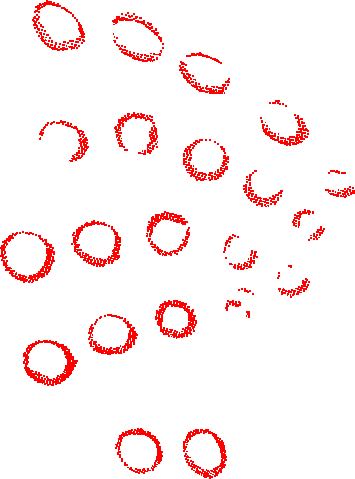} & &
\includegraphics[height=2.5cm]{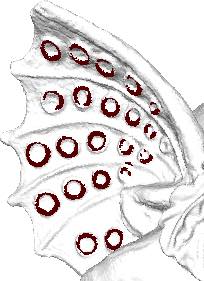}\\
&& (a) & & (b) & & (c)\\
\\
II. & &
\includegraphics[height=2.4cm]{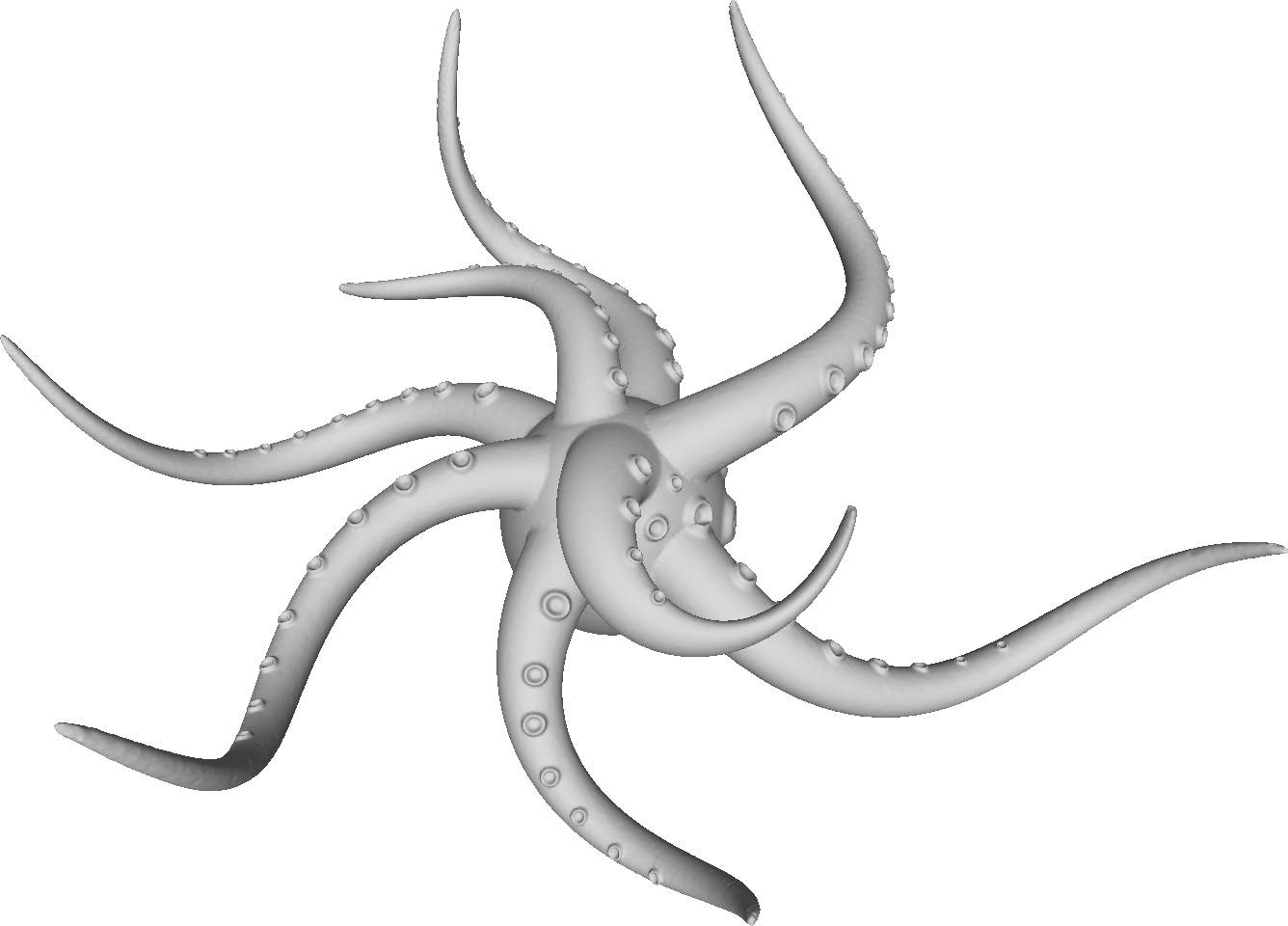} & &
\includegraphics[height=1.2cm]{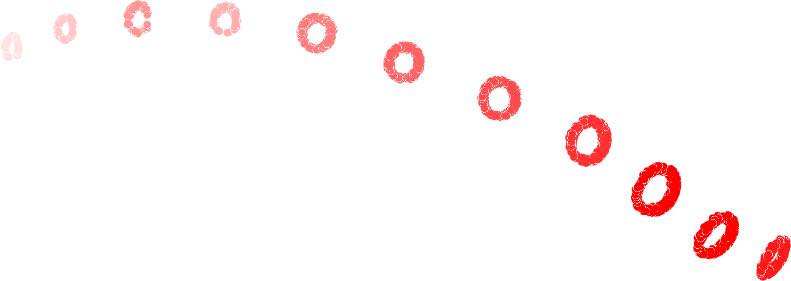} & &
\includegraphics[height=1.4cm]{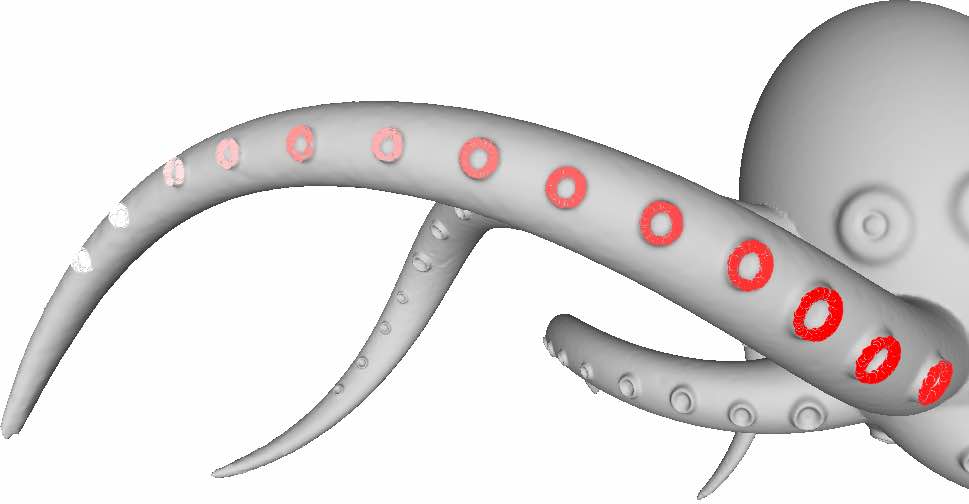} \\
&& (a) & & (b) & & (c)\\
\\
III. & &
\includegraphics[height=2.3cm]{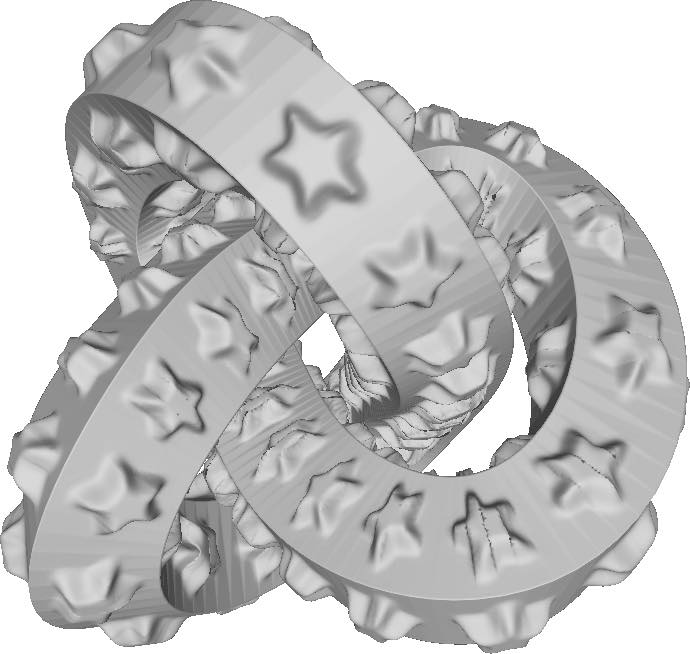} & &
\includegraphics[height=2cm]{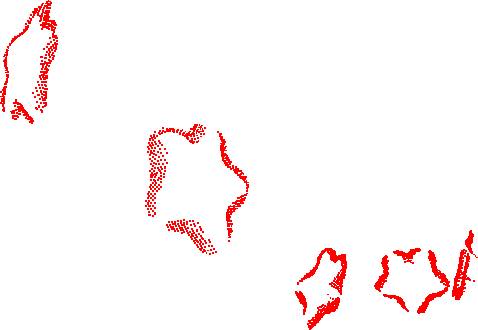} & &
\includegraphics[height=1.8cm]{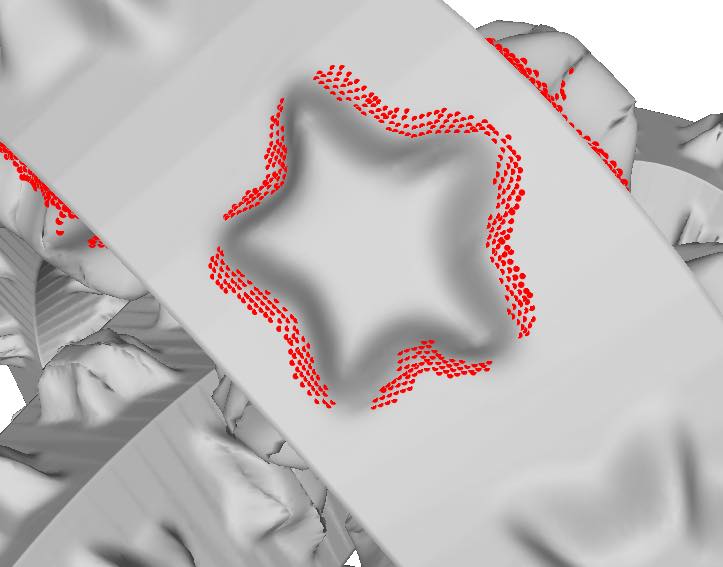}\\
&& (a) & & (b) & & (c)\\
\\
IV. & &
\includegraphics[height=2.5cm]{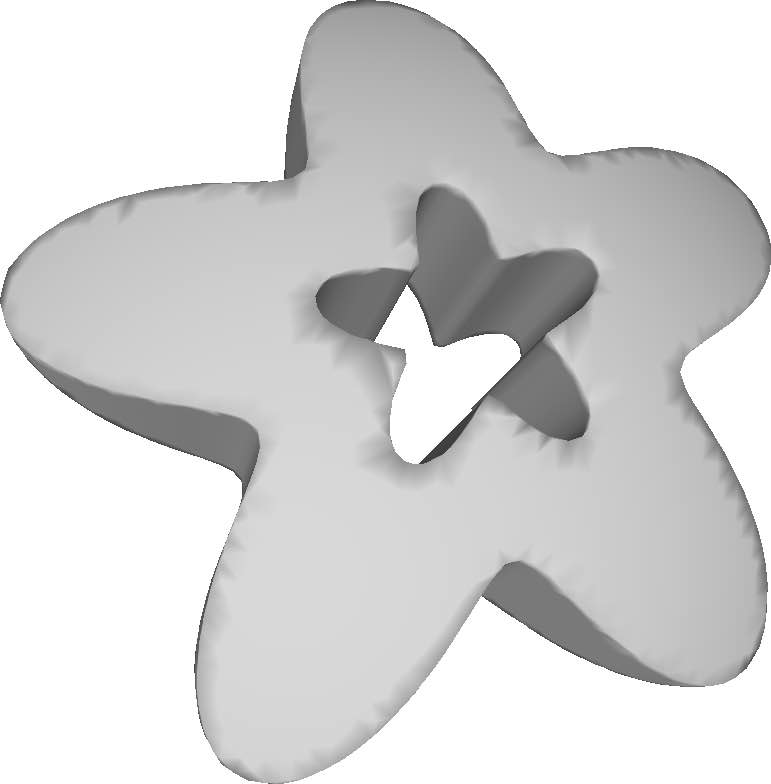} & &
\includegraphics[height=2.5cm]{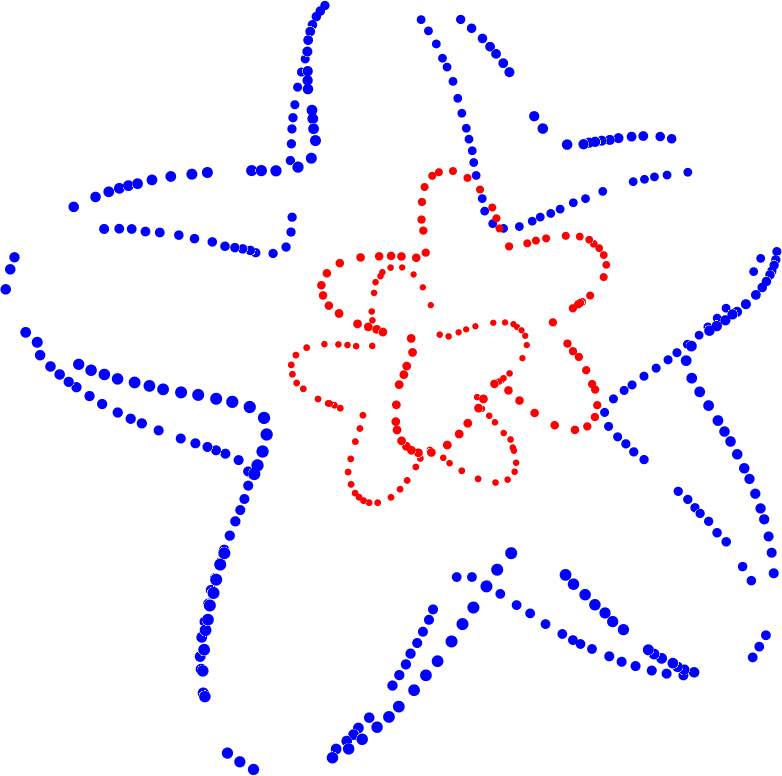} & &
\includegraphics[height=2.5cm]{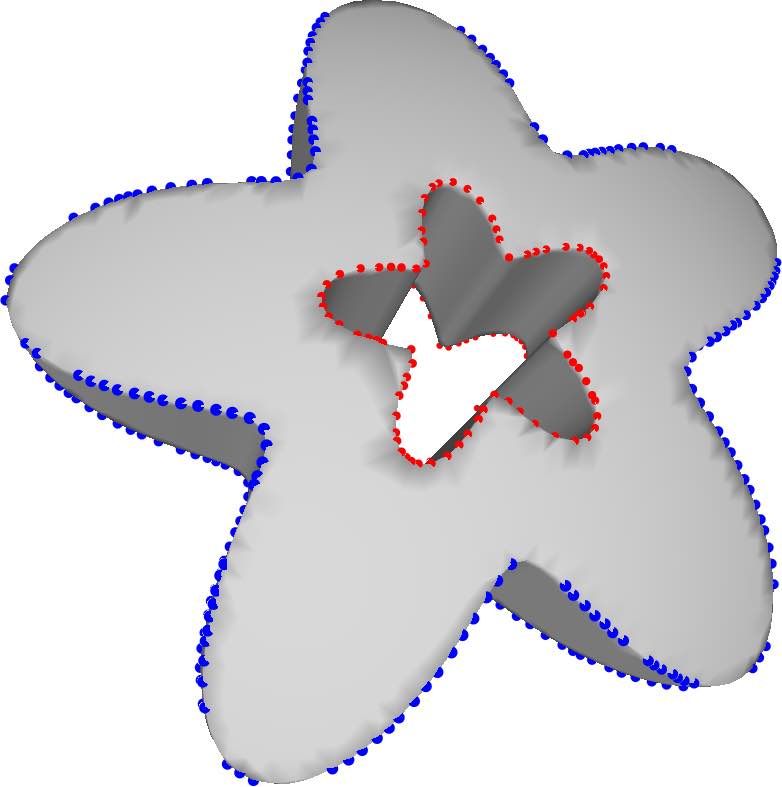}\\
&& (a) & & (b) & & (c)\\
\\
V. & &
\includegraphics[height=1.7cm]{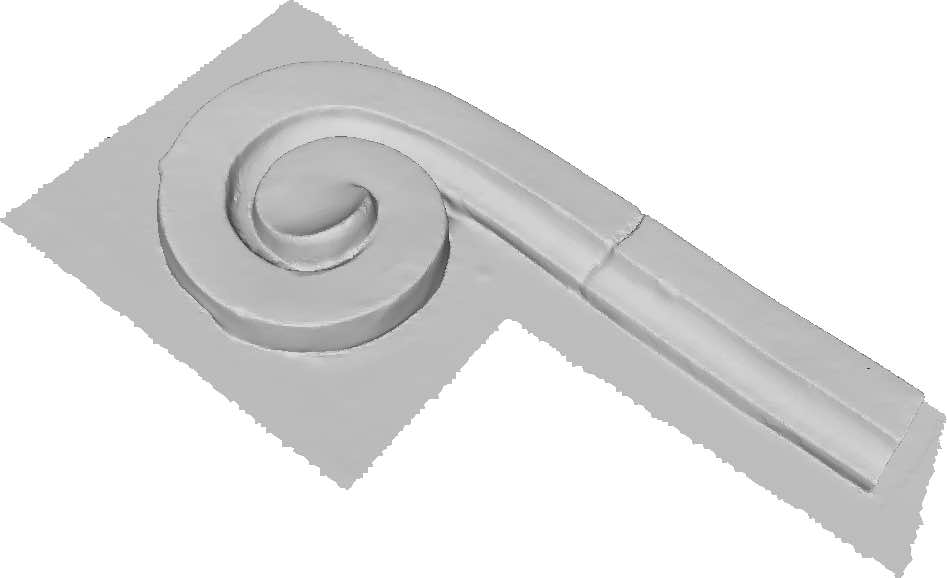} & &
\includegraphics[height=1.7cm]{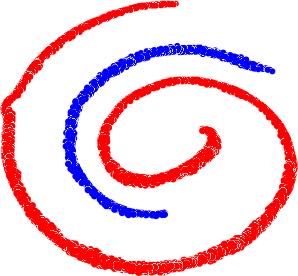} & &
\includegraphics[height=1.7cm]{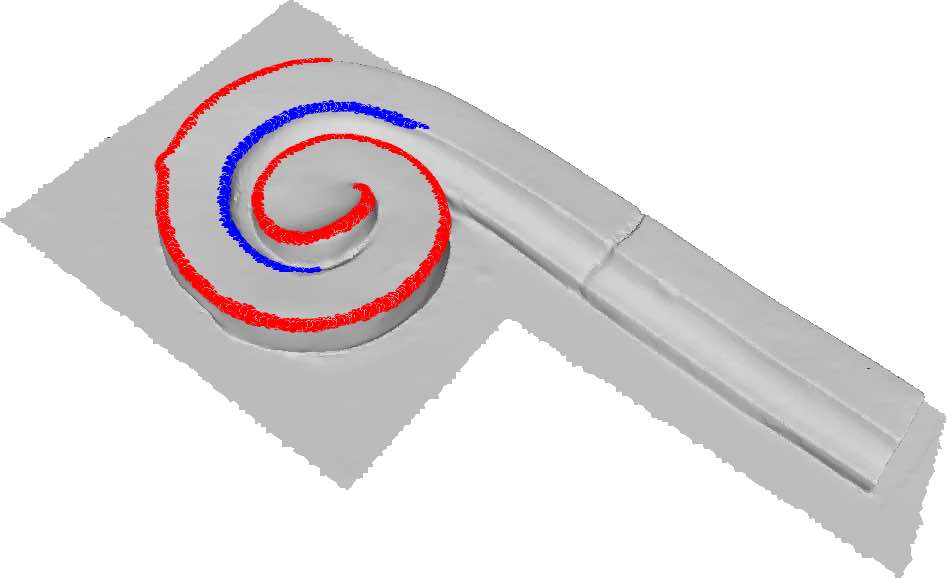}\\
&& (a) & & (b) & & (c)\\
\\
VI. & &
\includegraphics[height=1.3cm]{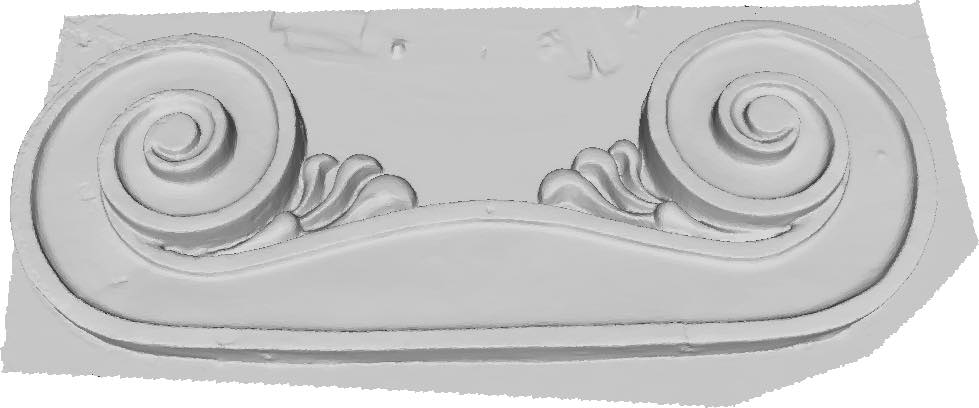} & &
\includegraphics[height=1.3cm]{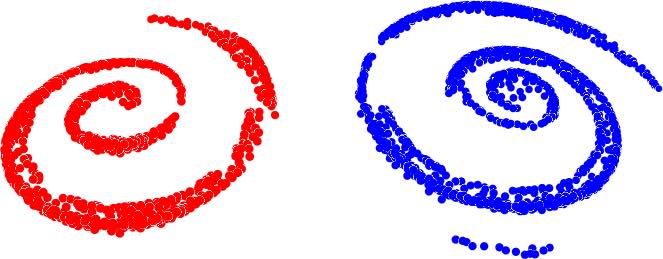} & &
\includegraphics[height=1.3cm]{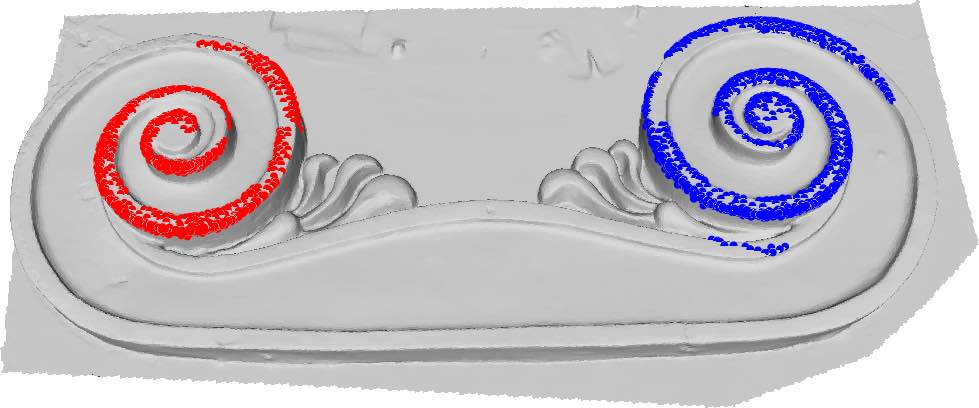}\\
&& (a) & & (b) & & (c)
\end{tabular}
\caption{Examples of recognition of various feature curves; 
I. circles on the Gargoyle model (the AIM@SHAPE repository \cite{VISIONAIR}); 
circles on a tentacle of an octopus model; 
III. stars (curves with $5$-convexities) on the knot model (the AIM@SHAPE repository \cite{VISIONAIR});
IV. stars (curves with $5$-convexities) of different sizes on the trim-star model (the AIM@SHAPE repository \cite{VISIONAIR});
V. and VI. spirals on architectural ornamental artefacts (from the 3D dataset of the EPSRC project \cite{projChevarria}).
}\label{figMultiple}
\end{figure}

Figure \ref{figMultiple} presents an overview of multiple feature curves obtained by our method. 
These examples are shown in a portion of the model but are valid for all the instances of the same features.
Multiple instances of the same family are shown in Figure \ref{figMultiple}(I) and Figure \ref{figMultiple}(II), 
in which the curves are circles with different radii, in Figure \ref{figMultiple}(III) and Figure \ref{figMultiple}(IV), 
where the detected curves belong to the family of curves with $5$-convexities, in Figure \ref{figMultiple}(V) and 
Figure \ref{figMultiple}(VI), where different Archimedean spirals have been used.

It is also possible to recognize different families of curves on the same surface, as it happens in the models in 
Figure \ref{figMultipleII}(I), where circles are combined with two curves of Lamet,
in Figure \ref{figMultipleII}(II), where a curve with $5$-convexities is used in combination with a circle, 
in Figure \ref{figMultipleII}(III), where a curve with $8$-convexities is contained in a region delimited by two 
concentric circles, and in Figure \ref{figMultipleII}(IV), where the decoration is a repeated pattern of leaves, 
each one identified with a specific group and approximated by a citrus or a geometric petal curve.

Figure \ref{figEyes} shows four examples of eye contours detection from 3D models. 
The eyes in the first row are better approximated by the citrus curve while in the other three cases 
the geometric petal is the best fitting curve.

The automatic detection of multiple instances of the same family of curves, possibly with parameter changes as 
in the case of circles of different radii (e.g. the suction cups of the octopus model, see again Figure \ref{figMultiple}.II), 
is automatically done in the space of the parameters by the HT procedure. As a side effect, this immediately yields the 
equation of the curve that represents these feature points.  

An important advantage of our approach is the flexible choice of the family of algebraic curves used to approximate 
the desired features, thus being adaptive to approximate various shapes.
Our set of primitives includes generic algebraic curves and it can be extended to all curves with an implicit representation, 
see Section \ref{Sec:curves}.

\begin{figure}[htb]
\begin{center}
\begin{tabular}{ccc}
\includegraphics[height=2.3cm]{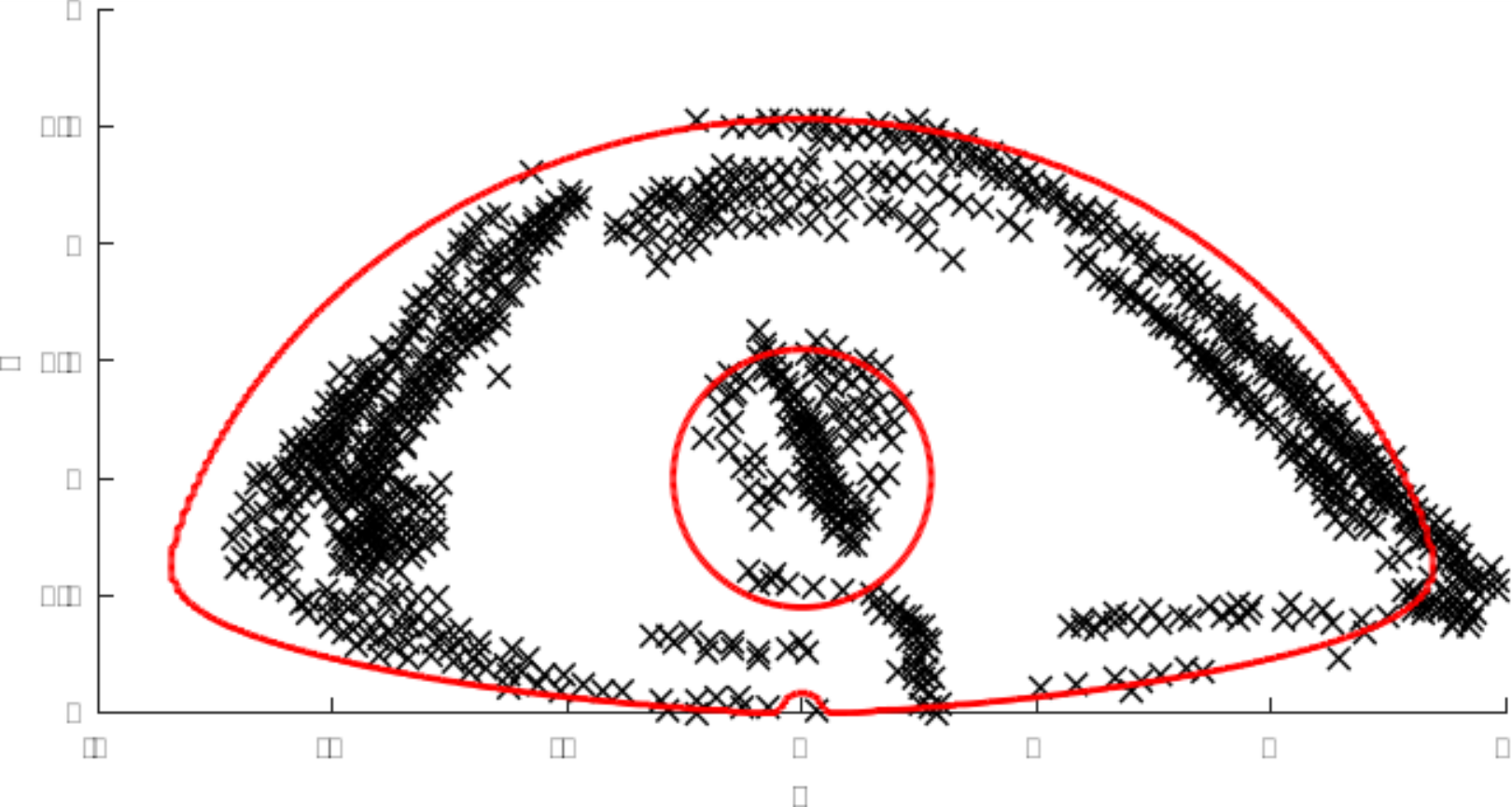} & &
\includegraphics[height=2.5cm]{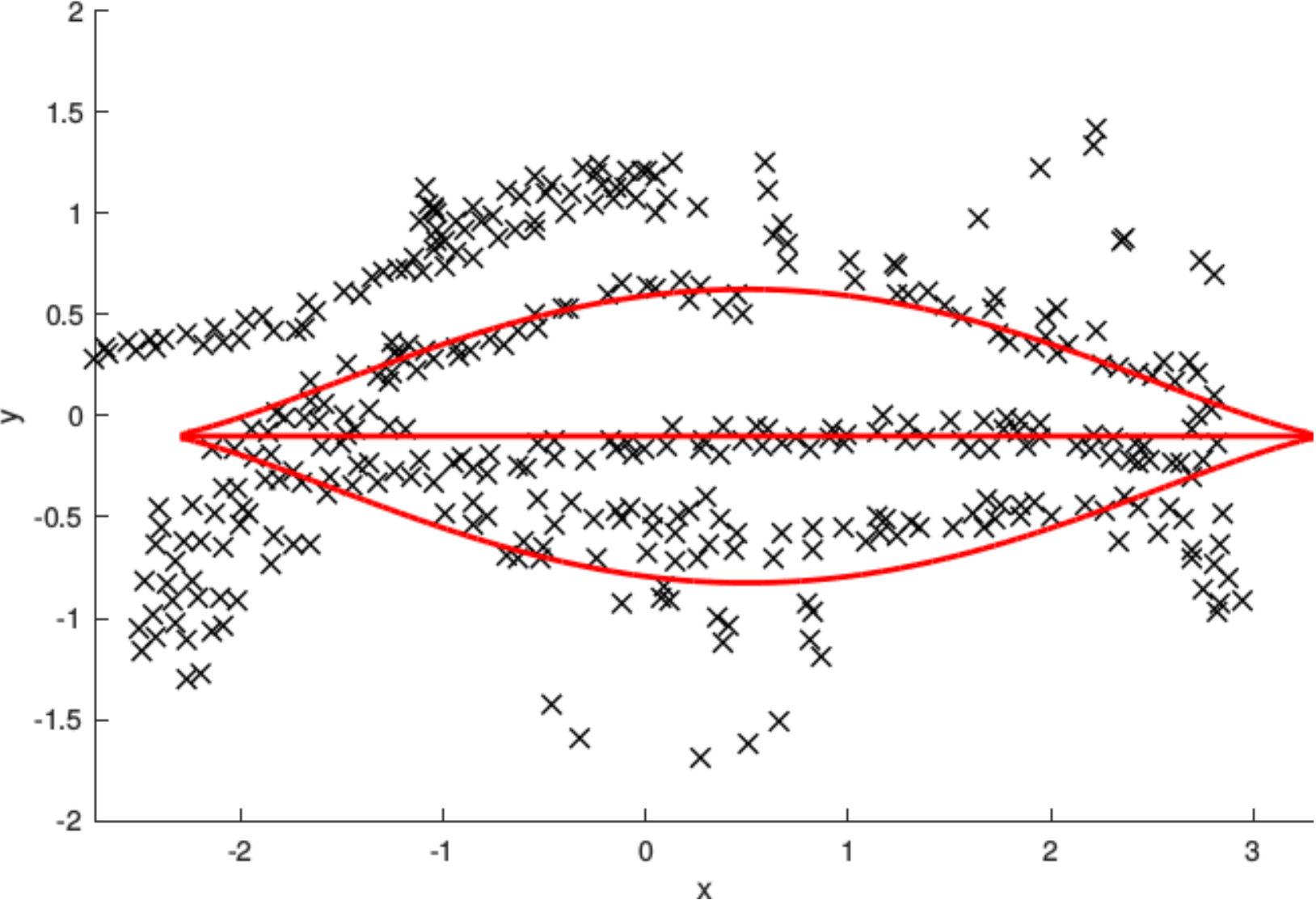}\\
(a) & & (b)\\
\end{tabular}
\caption{Combined anatomical shapes: (a) a circumference and a geometric petal curve detecting an eye;
(b) a citrus curve with a line detecting a mouth.}\label{combineFig}
\end{center}
\end{figure}

As discussed in Section \ref{sec:compound}, another interesting point is the possibility of recognizing compound features
as a whole, as for the eye contour and the pupil, see examples in Figures \ref{figEyes}.I and \ref{combineFig}(a). 
Figure \ref{combineFig}(b) detects a mouth contour combining a citrus curve with a line. Such an option opens the method 
to a wide range of curves that includes also repeated patterns like bundles of straight lines or quite complex decorations,
see examples in Figure \ref{figMultipleII}.

Another major benefit of using an HT-based method is the HT well known robustness to noise and outliers. Moreover, by selecting the curve with the highest score of the HT aggregation function, our method is able to keep a good recognition power also in the case of degraded (Figure \ref{noise}) and partial features (Figure \ref{partial}).
Moreover, we are able to work on both 3D meshes and point clouds, thus paving the road to the application of the method to object completion and model repairing.

In addition, it is important to point out that we can similarly parametrize features that are comparable. 
As we extract a feature curve and its parameters, we can use them to build a template useful for searching similar 
features in the artefacts, even if heavily incomplete, see Figure \ref{noise}(II.c-d) and Figure \ref{partial}(I.c-d) and Figure \ref{partial}(II.c-d). 

\begin{figure}[h]
\centering
\begin{tabular}{ccc}
\includegraphics[height=2.75cm]{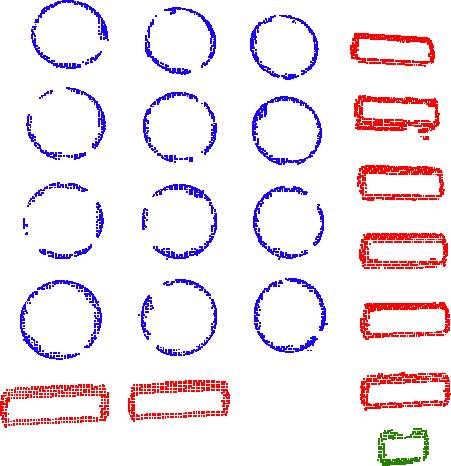} & &
\includegraphics[height=2cm]{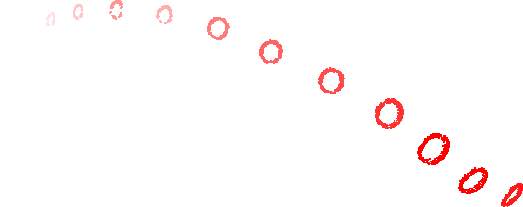}\\
(a) & & (b)\\
\end{tabular}
\caption{Colours represent different parameters of the HT.}\label{parameters}
\end{figure}

With reference to Figure \ref{parameters}, we use different colours to show the feature curves identified by different curves/parameters. In Figure \ref{parameters}(a), the blue lines represent the circles, all with the same radius, while the red and green lines depict the curves of Lamet. In particular, the red and the green lines 
are used for Lamet curves which differ for the values of the parameters~$a$ and~$b$. Similarly, in Figure \ref{parameters}(b) which represents a tentacle of an octopus model, the red graduate shadings 
are used to highlight the different radii of the detected circles.

\begin{figure}[hbt]
\centering
\begin{tabular}{ccccc}
\includegraphics[height=2cm]{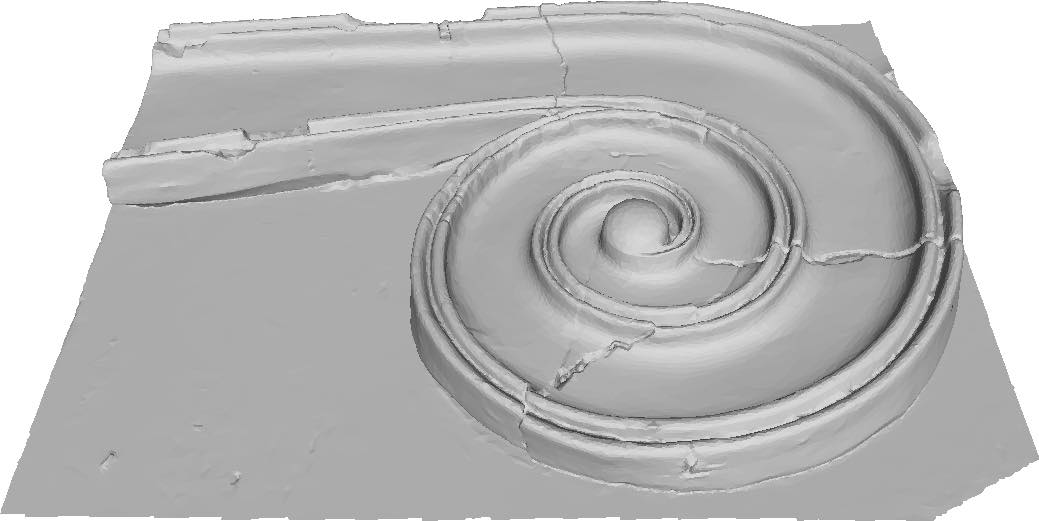} & &
\includegraphics[height=2cm]{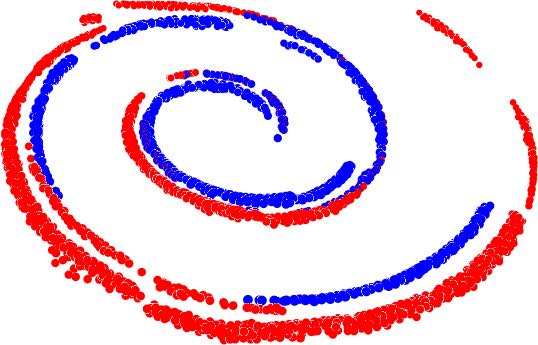} & &
\includegraphics[height=2cm]{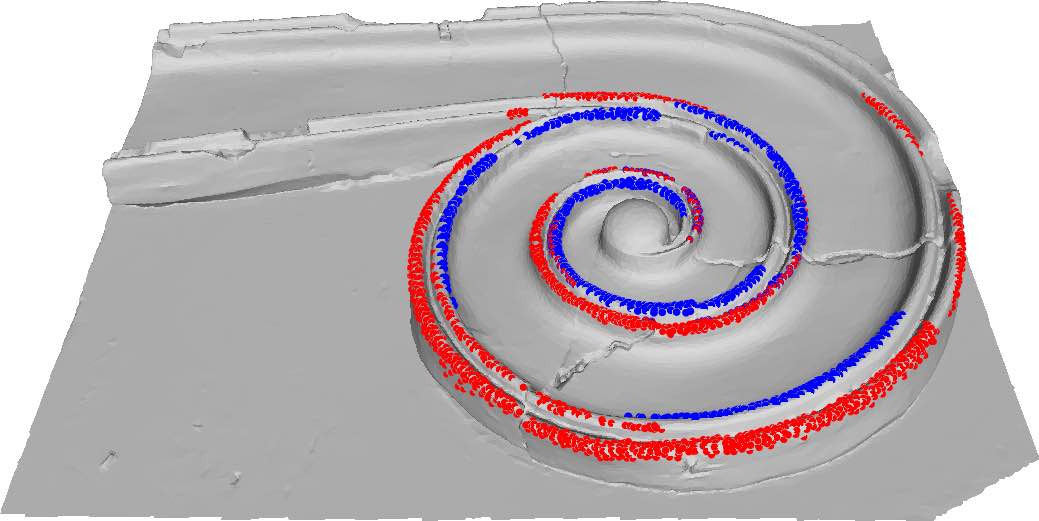}\\
(a) & & (b) && (c)\\
\\
\end{tabular}
\caption{Detection of two extremely nearby spirals on a degraded architectural ornament artefact 
(from the 3D dataset of the EPSRC project \cite{projChevarria}).}\label{closebySpirals}
\end{figure}

Different colours are also used in Figure \ref{closebySpirals} where two concentric spirals (red and blue) with different parameters, are detected. Because of the proximity of the two features 
curves and the presence of degradation and holes on the model, this example can be considered as a borderline case.
Nevertheless, our method results in a fairly good curve detection.

To give a concrete idea of the computational performances of our software prototype we report the running time of some of the examples discussed in the paper. 
The proposed method is relatively fast, a significant complexity being induced only by the Curve Detection Algorithm~\ref{algHough}. We implemented the whole pipeline in MATLAB 2016a, while for the HT detection routine we used the CoCoA library \cite{CoCoA-5}. All experiments were performed on an Intel Core i5 processor (at~2.7 GHz). Indeed, the Feature Points Recognition Algorithm~\ref{algFeaturePoints} generally takes an average of~$20$ seconds on a model of $150000$ vertices and a few seconds are taken by the Aggregation Algorithm~\ref{algClustering} and the Projection Algorithm~\ref{algProjection}. The complexity of the Curve Detection Algorithm~\ref{algHough} strongly depends on the family of curves, ranging from less than a minute for the curve with $5$-convexities
(example in Figure~\ref{figMultiple}-III), to a couple of minutes for the circle, the Archimedean spiral and the citrus curve 
(examples in Figures~\ref{figMultiple}-I, \ref{fig:spiral} and~\ref{figEyes}-I), to approximately three minutes for the Lamet curve
(example in Figure~\ref{figMultipleII}-I), to a dozen of minutes for the geometric petal (example in Figure~\ref{curvatures}).


The reasoning on parameters can be extended and possibly generalized. From our experiments, we noticed that eyes of the same collection, such as the ones in the STARC repository \cite{STARC}, share similar parameters. For instance, the eyes in Figure \ref{figEyes}(IV), \ref{noise}(II), \ref{partial}(I) and \ref{partial}(II) have nearly the same ratio among the parameters 
$a$ and $c$ (whose square roots represent the height and eccentricity of the curve) while in the case of the models in 
Figure \ref{figEyes}(II) and \ref{figEyes}(III) these values considerably differ. We plan to deepen these aspects on larger 
datasets aiming at automatically inferring a specific style from the models and supporting automatic model annotation.

As a minor drawback, we point out that the use of more complex algebraic curves may involve more than three parameters 
which has consequences in the definition and manipulation of the accumulator function, thus becoming computationally 
expensive (see Section \ref{computComplexity}), although ad-hoc methods have been introduced to solve it 
(see \cite{torrente2014almost}).

It is our opinion that the proposed method is particularly appropriate to drive the recognition of feature curves 
and the annotation of shape parts that are somehow expected to be in a model.
As an example, we refer to the case of archaeological artefacts that are always equipped with 
a textual description carrying information on the presence of features, e.g. eyes, mouth, decorations, etc. 
In our experience, this important extra information allowed us to recognize the eyes in the models
represented in Figure \ref{noise}(II), presenting a deep level of erosion, and \ref{partial}(II), where the 
feature is only partially discernible.
Analogously, we expect the method to be essential in situations where 
a taxonomic description of peculiar curve shapes exists (like architectural artefacts) 
or a standard reference shape is available, 
or when it is possible to infer a template of the feature curve to be identified.

\section*{Acknowledgements}
Work developed in the CNR research activity DIT.AD004.028.001, and partially supported by the GRAVI\-TATE European project, ``H2020 REFLECTIVE'', contract n. 665155, (2015-2018).
\bibliographystyle{plain}
\bibliography{references}

\begin{thebibliography}{10}

\bibitem{Imaginary}
{IMAGINARY} - open mathematics.
\newblock https://imaginary.org.

\bibitem{STARC}
{STARC} repository.
\newblock http://public.cyi.ac.cy/starcRepo/.

\bibitem{VISIONAIR}
{The Shape Repository}.
\newblock http://visionair.ge.imati.cnr.it/ontologies/shapes/, 2011--2015.

\bibitem{projChevarria}
{Automatic Semantic Analysis of 3D Content in Digital Repositories}.
\newblock http://www.ornament3d.org/, 2014--2016.

\bibitem{CoCoA-5}
J.~Abbott, A.~M. Bigatti, and G.~Lagorio.
\newblock {CoCoA-5}: a system for doing {C}omputations in {C}ommutative
  {A}lgebra.
\newblock Available at \texttt{http://cocoa.dima.unige.it}.

\bibitem{AKK00}
E.~Albuz, E.~D. Kocalar, and A.~A. Khokhar.
\newblock Quantized cielab* space and encoded spatial structure for scalable
  indexing of large color image archives.
\newblock In {\em Acoustics, Speech, and Signal Processing}, volume~6, pages
  1995--1998, 2000.

\bibitem{APM15}
A.~Andreadis, G.~Papaioannou, and P.~Mavridis.
\newblock Generalized digital reassembly using geometric registration.
\newblock In {\em Digital Heritage}, volume~2, pages 549--556, 2015.

\bibitem{ballard1981generalizing}
Dana~H Ballard.
\newblock Generalizing the {Hough} transform to detect arbitrary shapes.
\newblock {\em Pattern recognition}, 13(2):111--122, 1981.

\bibitem{beltrametti2012algebraic}
Mauro~C Beltrametti and Lorenzo Robbiano.
\newblock An algebraic approach to {Hough} transforms.
\newblock {\em J. of Algebra}, 37:669--681, 2012.

\bibitem{beltrametti2013hough}
MC~Beltrametti, AM~Massone, and M~Piana.
\newblock Hough transform of special classes of curves.
\newblock {\em SIAM J. Imaging Sci.}, 6(1):391--412, 2013.

\bibitem{Biasotti2016:CGF}
S.~Biasotti, A.~Cerri, A.~Bronstein, and M.~Bronstein.
\newblock Recent trends, applications, and perspectives in {3D} shape
  similarity assessment.
\newblock {\em Computer Graphics Forum}, 35(6):87--119, 2016.

\bibitem{CSUR2008}
S.~Biasotti, L.~{De~Floriani}, B.~Falcidieno, P.~Frosini, D.~Giorgi, C.~Landi,
  L.~Papaleo, and M.~Spagnuolo.
\newblock Describing shapes by geometrical-topological properties of real
  functions.
\newblock {\em ACM Computing Surveys}, 40(4):1--87, 2008.

\bibitem{BCF*15}
Silvia Biasotti, Andrea Cerri, Bianca Falcidieno, and Michela Spagnuolo.
\newblock {3D} artifacts similarity based on the concurrent evaluation of
  heterogeneous properties.
\newblock {\em J. Comput. Cult. Herit.}, 8(4):19:1--19:19, 2015.

\bibitem{Cao2015}
Yuanhao Cao, Dong-Ming Yan, and Peter Wonka.
\newblock Patch layout generation by detecting feature networks.
\newblock {\em Computers \& Graphics}, 46:275 -- 282, 2015.

\bibitem{Cohen03}
David Cohen-Steiner and Jean-Marie Morvan.
\newblock Restricted {Delaunay} triangulations and normal cycle.
\newblock In {\em Proc. of the $9^{th}$ Ann. Symp. on Computational Geometry},
  SCG '03, pages 312--321, New York, NY, USA, 2003. ACM.

\bibitem{Cole:2008}
Forrester Cole, Aleksey Golovinskiy, Alex Limpaecher, Heather~Stoddart Barros,
  Adam Finkelstein, Thomas Funkhouser, and Szymon Rusinkiewicz.
\newblock Where do people draw lines?
\newblock {\em ACM Trans. Graph.}, 27(3):1--11, August 2008.

\bibitem{DaTr05}
N.~Dalal and B.~Triggs.
\newblock Histograms of oriented gradients for human detection.
\newblock In {\em Computer Vision and Pattern Recognition (CVPR), 2005 IEEE
  Conference on}, volume~1, pages 886--893, 2005.

\bibitem{DOH08}
Joel Daniels~II, Tilo Ochotta, K.~Linh Ha, and T.~Cl{\'a}udio Silva.
\newblock Spline-based feature curves from point-sampled geometry.
\newblock {\em The Visual Computer}, 24(6):449--462, 2008.

\bibitem{deGoes2011}
Fernando de~Goes, Siome Goldenstein, Mathieu Desbrun, and Luiz Velho.
\newblock Exoskeleton: Curve network abstraction for 3d shapes.
\newblock {\em Computers \& Graphics}, 35(1):112 -- 121, 2011.

\bibitem{DeCarlo}
Doug DeCarlo and Szymon Rusinkiewicz.
\newblock Highlight lines for conveying shape.
\newblock In {\em International Symposium on Non-Photorealistic Animation and
  Rendering (NPAR)}, pages 63--70. ACM, August 2007.

\bibitem{DH72}
Richard~O. Duda and Peter~E. Hart.
\newblock Use of the {Hough} transformation to detect lines and curves in
  pictures.
\newblock {\em Commun. ACM}, 15(1):11--15, 1972.

\bibitem{EKS*96}
Martin Ester, Hans~P. Kriegel, Jorg Sander, and Xiaowei Xu.
\newblock A density-based algorithm for discovering clusters in large spatial
  databases with noise.
\newblock In {\em $2^{nd}$ Int. Conf. Knowledge Discovery and Data Mining},
  pages 226--231. AAAI Press, 1996.

\bibitem{Farin:1993}
Gerald Farin.
\newblock {\em Curves and Surfaces for Computer Aided Geometric Design (3rd
  Ed.): A Practical Guide}.
\newblock Academic Press Professional, Inc., San Diego, CA, USA, 1993.

\bibitem{FBF77}
Jerome~H. Friedman, Jon~Louis Bentley, and Raphael~Ari Finkel.
\newblock An algorithm for finding best matches in logarithmic expected time.
\newblock {\em ACM Trans. Math. Softw.}, 3(3):209--226, 1977.

\bibitem{Friedman:1977}
Jerome~H. Friedman, Jon~Louis Bentley, and Raphael~Ari Finkel.
\newblock An algorithm for finding best matches in logarithmic expected time.
\newblock {\em ACM Trans. Math. Softw.}, 3(3):209--226, September 1977.

\bibitem{gal2006salient}
Ran Gal and Daniel Cohen-Or.
\newblock Salient geometric features for partial shape matching and similarity.
\newblock {\em ACM Transactions on Graphics (TOG)}, 25(1):130--150, 2006.

\bibitem{Gehre16}
Anne Gehre, Isaak Lim, and Leif Kobbelt.
\newblock {Adapting Feature Curve Networks to a Prescribed Scale}.
\newblock {\em Computer Graphics Forum}, 35(2):319--330, 2016.

\bibitem{Gumhold01}
Stefan Gumhold, Xinlong Wang, and Rob Macleod.
\newblock Feature extraction from point clouds.
\newblock In {\em $10^{th}$ Int. Meshing Roundtable}, pages 293--305, 2001.

\bibitem{Harary2011}
Gur Harary and Ayellet Tal.
\newblock {The Natural 3D Spiral}.
\newblock {\em Computer Graphics Forum}, 30(2):237--246, 2011.

\bibitem{Harary2012}
Gur Harary and Ayellet Tal.
\newblock {3D Euler} spirals for {3D} curve completion.
\newblock {\em Computational Geometry}, 45(3):115 -- 126, 2012.

\bibitem{HPW05}
Klaus Hildebrandt, Konrad Polthier, and Max Wardetzky.
\newblock Smooth feature lines on surface meshes.
\newblock In {\em Eurographics Symposium on Geometry Processing}. The
  Eurographics Association, 2005.

\bibitem{c1962method}
Paul. V.~C Hough.
\newblock Method and means for recognizing complex patterns, 1962.
\newblock {US} Patent 3,069,654.

\bibitem{color2011}
R.~W.~G. Hunt and M.~R. Pointer.
\newblock {\em Measuring Colour, Fourth Edition}.
\newblock Wiley, 2011.

\bibitem{Itskovich2011}
Arik Itskovich and Ayellet Tal.
\newblock Surface partial matching and application to archaeology.
\newblock {\em Computers \& Graphics}, 35(2):334 -- 341, 2011.

\bibitem{Johnson:1999}
Andrew~E. Johnson and Martial Hebert.
\newblock Using spin images for efficient object recognition in cluttered 3d
  scenes.
\newblock {\em IEEE T. Pattern Anal.}, 21(5):433--449, 1999.

\bibitem{kalogerakis07}
Evangelos Kalogerakis, Patricio Simari, Derek Nowrouzezahrai, and Karan Singh.
\newblock Robust statistical estimation of curvature on discretized surfaces.
\newblock In {\em Proc. of the $5^{th}$ EG Symp. on Geometry Processing}, pages
  13--22. Eurographics Association, 2007.

\bibitem{Kanezaki:2010}
Asako Kanezaki, Tatsuya Harada, and Yasuo Kuniyoshi.
\newblock Partial matching of real textured 3d objects using color cubic
  higher-order local auto-correlation features.
\newblock {\em Visual Comput.}, 26(10):1269--1281, 2010.

\bibitem{Kassim1999}
A.A Kassim, T~Tan, and K.H Tan.
\newblock A comparative study of efficient generalised {Hough} transform
  techniques.
\newblock {\em Image and Vision Computing}, 17(10):737 -- 748, 1999.

\bibitem{Kerber2013}
J.~Kerber, M.~Bokeloh, M.~Wand, and H.-P. Seidel.
\newblock Scalable symmetry detection for urban scenes.
\newblock {\em Computer Graphics Forum}, 32(1):3--15, 2013.

\bibitem{koenderink1984}
Jan~J Koenderink.
\newblock What does the occluding contour tell us about solid shape?
\newblock {\em Perception}, 13(3):321--330, 1984.

\bibitem{KST11}
M.~Kolomenkin, I.~Ilan~Shimshoni, and A.~Tal.
\newblock Prominent field for shape processing of archaeological artifacts.
\newblock {\em Int. J. Comput. Vision}, 94(1):89--100, 2011.

\bibitem{KST08}
Michael Kolomenkin, Ilan Shimshoni, and Ayellet Tal.
\newblock Demarcating curves for shape illustration.
\newblock {\em ACM Trans. Graph.}, 27(5):157:1--157:9, 2008.

\bibitem{Lai2007}
Y.~K. Lai, Q.~Y. Zhou, S.~M. Hu, J.~Wallner, and H.~Pottmann.
\newblock Robust feature classification and editing.
\newblock {\em IEEE Transactions on Visualization and Computer Graphics},
  13(1):34--45, Jan 2007.

\bibitem{Lawonn2017}
K.~Lawonn, E.~Trostmann, B.~Preim, and K.~Hildebrandt.
\newblock Visualization and extraction of carvings for heritage conservation.
\newblock {\em IEEE Transactions on Visualization and Computer Graphics},
  23(1):801--810, Jan 2017.

\bibitem{Li2015}
C.~Li, M.~Wand, X.~Wu, and H.~P. Seidel.
\newblock Approximate 3d partial symmetry detection using co-occurrence
  analysis.
\newblock In {\em 2015 International Conference on 3D Vision}, pages 425--433,
  Oct 2015.

\bibitem{Liu:2012}
Yong-Jin Liu, Yi-Fu Zheng, Lu~Lv, Yu-Ming Xuan, and Xiao-Lan Fu.
\newblock {3D} model retrieval based on color + geometry signatures.
\newblock {\em Visual Comput.}, 28(1):75--86, 2012.

\bibitem{Lowe2004}
David~G. Lowe.
\newblock Distinctive image features from scale-invariant keypoints.
\newblock {\em Int. J. Comput. Vision}, 60(2):91--110, 2004.

\bibitem{Luo2010}
Tao Luo, Renju Li, and Hongbin Zha.
\newblock {3D} line drawing for archaeological illustration.
\newblock {\em Int. J. Comput. Vision}, 94(1):23--35, 2010.

\bibitem{Mukhopadhyay2015}
Priyanka Mukhopadhyay and Bidyut~B. Chaudhuri.
\newblock A survey of {Hough} transform.
\newblock {\em Pattern Recognition}, 48(3):993 -- 1010, 2015.

\bibitem{alliez}
Sven Oesau, Florent Lafarge, and Pierre Alliez.
\newblock {Indoor Scene Reconstruction using Feature Sensitive Primitive
  Extraction and Graph-cut}.
\newblock {\em {ISPRS Journal of Photogrammetry and Remote Sensing}},
  90:68--82, March 2014.

\bibitem{OBS04}
Yutaka Ohtake, Alexander Belyaev, and Hans-Peter Seidel.
\newblock Ridge-valley lines on meshes via implicit surface fitting.
\newblock {\em ACM Trans. Graph.}, 23(3):609--612, 2004.

\bibitem{ojala}
Timo Ojala, Matti Pietik{\"a}inen, and David Harwood.
\newblock A comparative study of texture measures with classification based on
  featured distributions.
\newblock {\em Pattern Recognition}, 29(1):51--59, 1996.

\bibitem{Pasqualotto2013}
Giuliano Pasqualotto, Pietro Zanuttigh, and Guido~M. Cortelazzo.
\newblock {Combining color and shape descriptors for 3D model retrieval}.
\newblock {\em Signal Process-Image}, 28(6):608 -- 623, 2013.

\bibitem{PKG03}
Mark Pauly, Richard Keiser, and Markus Gross.
\newblock Multi-scale feature extraction on point-sampled surfaces.
\newblock {\em Comput. Graph. Forum}, 22(3):281--289, 2003.

\bibitem{ToolboxGraph}
G.~Peyre.
\newblock Toolbox graph - {A} toolbox to process graph and triangulated meshes.
\newblock http://www.ceremade.dauphine.fr/~peyre/matlab/graph/content.html.

\bibitem{Piegl:1997}
Les Piegl and Wayne Tiller.
\newblock {\em The NURBS Book (2Nd Ed.)}.
\newblock Springer-Verlag New York, Inc., New York, NY, USA, 1997.

\bibitem{rudin}
W.~Rudin.
\newblock {\em Principles of Mathematical Analysis}.
\newblock McGraw-Hill,, Singapore:, 3rd edition edition, 1976.

\bibitem{Ruiz09}
C.R. Ruiz, R.~Cabredo, L.J. Monteverde, and Zhiyong Huang.
\newblock {Combining Shape and Color for Retrieval of 3D Models}.
\newblock In {\em INC, IMS and IDC (NCM'09). Fifth International Joint
  Conference on}, pages 1295--1300, 2009.

\bibitem{Shamir:2006:STAR}
Ariel Shamir.
\newblock {Segmentation and Shape Extraction of 3D Boundary Meshes}.
\newblock In Brian Wyvill and Alexander Wilkie, editors, {\em Eurographics 2006
  - State of the Art Reports}. The Eurographics Association, 2006.

\bibitem{shikin}
E.~V. Shikin and A.~I. Plis.
\newblock {\em Handbook on Splines for the User}.
\newblock CRC Press, Boca Raton, FL, 1995.

\bibitem{shikin1995handbook}
Eugene~V Shikin.
\newblock {\em Handbook and atlas of curves}.
\newblock CRC, 1995.

\bibitem{Starck07}
J.~Starck and A~Hilton.
\newblock Correspondence labelling for wide-timeframe free-form surface
  matching.
\newblock In {\em Computer Vision (ICCV), 2007 IEEE International Conference
  on}, pages 1--8, 2007.

\bibitem{SunkelEG2011}
Martin Sunkel, Silke Jansen, Michael Wand, Elmar Eisemann, and Hans-Peter
  Seidel.
\newblock Learning line features in {3D} geometry.
\newblock {\em Computer Graphics Forum (Proc. EUROGRAPHICS)}, 30(2), April
  2011.

\bibitem{Suzuki01}
M.T. Suzuki.
\newblock {A Web-based retrieval system for 3D polygonal models}.
\newblock In {\em IFSA World Congress and 20th NAFIPS International Conference.
  Joint 9th}, volume~4, pages 2271--2276, 2001.

\bibitem{TWW01}
James Tanaka, Daniel Weiskopf, and Pepper Williams.
\newblock The role of color in high-level vision.
\newblock {\em Trends in cognitive sciences}, 5(5):211--215, 2001.

\bibitem{TombariSS11}
F.~Tombari, S.~Salti, and L.~Di~Stefano.
\newblock A combined texture-shape descriptor for enhanced 3d feature matching.
\newblock In {\em Image Processing (ICIP), 2011 IEEE International Conference
  on}, pages 809--812, 2011.

\bibitem{torrente2014almost}
Maria-Laura Torrente and Mauro~C Beltrametti.
\newblock Almost vanishing polynomials and an application to the {Hough}
  transform.
\newblock {\em J. of Algebra and Its Applications}, 13(08):1450057, 2014.

\bibitem{TorrenteBeltramettiSendra}
Maria-Laura Torrente, Mauro~C. Beltrametti, and Juan~Rafael Sendra.
\newblock Perturbation of polynomials and applications to the hough transform.
\newblock {\em Journal of Algebra}, 486:328 -- 359, 2017.

\bibitem{gch.2016}
Maria-Laura Torrente, Silvia Biasotti, and Bianca Falcidieno.
\newblock {Feature Identification in Archaeological Fragments Using Families of
  Algebraic Curves}.
\newblock In Chiara~Eva Catalano and Livio~De Luca, editors, {\em Eurographics
  Workshop on Graphics and Cultural Heritage}. The Eurographics Association,
  2016.

\bibitem{vasa}
Libor V\'{a}\v{s}a, Petr Van\v{e}\v{c}ek, Martin Prantl, V\v{e}ra
  Skorkovsk\'{a}, Petr Mart\'{i}nek, and Ivana Kolingerov\'{a}.
\newblock {Mesh Statistics for Robust Curvature Estimation}.
\newblock {\em Computer Graphics Forum}, 35(5):271--280, 2016.

\bibitem{WuCLFP08}
Changchang Wu, B.~Clipp, Xiaowei Li, J.-M. Frahm, and M.~Pollefeys.
\newblock {3D model matching with Viewpoint-Invariant Patches ({VIP})}.
\newblock In {\em Computer Vision and Pattern Recognition (CVPR), 2008 IEEE
  Conference on}, pages 1--8, 2008.

\bibitem{YBY*08}
Shin Yoshizawa, Alexander Belyaev, Hideo Yokota, and Hans-Peter Seidel.
\newblock Fast, robust, and faithful methods for detecting crest lines on
  meshes.
\newblock {\em Comput. Aided Geom. Des.}, 25(8):545--560, 2008.

\bibitem{meshHOG}
Andrei Zaharescu, Edmond Boyer, and Radu Horaud.
\newblock Keypoints and local descriptors of scalar functions on {2D}
  manifolds.
\newblock {\em Int. J. Comput. Vision}, 100(1):78--98, 2012.

\bibitem{Zhang2016}
Yuhe Zhang, Guohua Geng, Xiaoran Wei, Shunli Zhang, and Shanshan Li.
\newblock A statistical approach for extraction of feature lines from point
  clouds.
\newblock {\em Computers \& Graphics}, 56:31 -- 45, 2016.

\end{thebibliography}

\newpage
\section*{Appendix}
We list the pseudocode of some procedures mentioned into Sections \ref{FeaturePointChar}-\ref{Hough}.

\begin{algorithm}[H]
\SetAlgorithmName{Filtering Algorithm}
\DontPrintSemicolon
\SetAlgoLined
\SetKwInOut{Input}{Input}
\SetKwInOut{Output}{Output}
\SetKwInOut{return}{return}
\Input{histogram $h$, cut percentage $p$}
\Output{cut value $v$}
   
 \Begin{  
Values = {\em get$\_$values}(h)\;
V=0; k=0\;
\While{$k\le size(h) \: \textrm{{\em and}} \; V < p \cdot size(h)$}{
   $k = k+1$;
   V = V+Values[k]\;
   }
m = {\em min$\_$bin$\_$value}(h)\;
width= {\em get$\_$bin$\_$width}(h)\;
\return{$v$ = m+k $\cdot$ width}
}
\caption{Computes the value $v$ of the histogram $h$ corresponding to the threshold~$p$.}\label{algFilter}
\end{algorithm}

\begin{algorithm}[H]
\SetAlgorithmName{Aggregation Algorithm}
\DontPrintSemicolon
\SetAlgoLined
\SetKwInOut{Input}{Input}
\SetKwInOut{Output}{Output}
\SetKwInOut{return}{return}
\Input{set of feature points $\mathbb X \subset \mathbb R^3$}
\Output{$\mathbb Y$, a set of groups $\mathbb Y_j$ of elements of $\mathbb X$}
   
\Begin{  
\tcc{searches the average distance from each point of $\mathbb X$ to its $K$-nearest neighbours;
the parameter $K$ is fixed and refers to how many neighbours we want to consider}
$K=50$\;
$[IDX, D] =$ {\em knnsearch}$(\mathbb X, \mathbb X, `k', K)$\;

\tcc{computes an estimate for the threshold $\varepsilon$}
$\varepsilon$ = {\em mean}(D(:, K))\;

\tcc{aggregates the points of $\mathbb X$ using the DBSCAN algorithm}
$MinPts=5$\;
IDX = DBSCAN($\mathbb X$, $\varepsilon$, MinPts)\;
\For{$j=1, \ldots, \textrm{{\em size}}(IDX)$}{
$\mathbb Y_j \gets $ the points of $\mathbb X$ of indices $IDX[j]$\;
add $\mathbb Y_j$ to the list $\mathbb Y$
}
\return{$\mathbb Y$}
}
 \caption{Groups the points of the set $\mathbb X$ into smaller dense subsets}\label{algClustering}
 \end{algorithm}

\begin{algorithm}[H]
\SetAlgorithmName{Projection Algorithm}
\DontPrintSemicolon
\SetAlgoLined
\SetKwInOut{Input}{Input}
\SetKwInOut{Output}{Output}
\SetKwInOut{return}{return}
\Input{a set of points $\mathbb Y$}
\Output{the projection $\mathbb Z$ of $\mathbb Y$ onto a best fitting plane}
   
\Begin{  
\tcc{shifts the points of $\mathbb Y$ to move the centroid onto  $(0,0,0)$}
$\mathbb Y \gets \mathbb Y - \textrm{\em mean}(\mathbb Y)$\;

\tcc{computes the best fitting plane $\Pi$ of $\mathbb Y$ with linear regression}
$XY \gets \mathbb{Y}[1,2]$; $Z \gets\mathbb{Y}[3]$;
$\Pi \gets \textrm{\em{regress}}(Z, XY)$\;

\tcc{defines the orthogonal transformation which moves $\Pi$ to the plane $z=0$, and apply it to the points of $\mathbb Y$}
$n \gets [\Pi(1), \Pi(2), -1]$;
$v_1 \gets [1, 0, \Pi(1)]$;
$v_2 \gets \textrm{\em CrossProduct}(v_1, n)$\;
$R \gets [v_1/\textrm{\em norm}(v_1), v_2/\textrm{\em norm}(v_2), n/\textrm{\em norm}(n)]$\;
$\mathbb Z \gets R(\mathbb Y)$\;
\return{$\mathbb Z[1,2]$} 
}
\caption{Projects a set of feature points $\mathbb Y$ onto a best fitting plane}\label{algProjection}
\end{algorithm}

\bigskip
\bigskip

\noindent
Maria-Laura Torrente,
Istituto di Matematica Applicata e Tecnologie Informatiche ``E. Magenes" CNR, Genova, Italy. 
e-mail {\tt laura.torrente@ge.imati.cnr.it}

\smallskip

\noindent 
Silvia Biasotti,
Istituto di Matematica Applicata e Tecnologie Informatiche ``E. Magenes" CNR, Genova, Italy. 
e-mail {\tt silvia.biasotti@ge.imati.cnr.it}

\smallskip

\noindent 
Bianca Falcidieno,
Istituto di Matematica Applicata e Tecnologie Informatiche ``E. Magenes" CNR, Genova, Italy. 
e-mail {\tt bianca.falcidieno@ge.imati.cnr.it}

\end{document}